\RequirePackage{fix-cm}
\documentclass[twocolumn, final]{svjour3}          
\usepackage{amsfonts}
\usepackage{times}		
\usepackage{amssymb}
\usepackage{mathrsfs}
\usepackage{amsmath}
\usepackage[linesnumbered,ruled]{algorithm2e}
\usepackage{graphicx}
\usepackage{subfig}
\usepackage{colortbl,booktabs}
\usepackage{threeparttable}
\usepackage{subeqnarray}
\usepackage{cases}
\usepackage{multirow}
\usepackage{array}
\usepackage[table]{xcolor}
\usepackage{verbatim}
\usepackage{url}
\usepackage{float}

\DeclareMathOperator*{\argmin}{argmin}

\let\oldnl\nl
\newcommand{\nonl}{\renewcommand{\nl}{\let\nl\oldnl}}

\smartqed  
\begin{document}

\title{On Unifying Multi-View Self-Representations for Clustering by Tensor Multi-Rank Minimization
}


\author{Yuan Xie         \and
        Dacheng Tao      \and
        Wensheng Zhang   \and
        Yan Liu          \and
        Lei Zhang        \and
        Yanyun Qu
}


\institute{Y. Xie \at
              the Research Center of Precision Sensing and Control, Institute of Automation, Chinese Academy of Sciences, Beijing, 100190, China \\
              Tel.: +86-13716206758\\
              \email{yuan.xie@ia.ac.cn}           
           \and
           D. Tao \at
              the Center for Quantum Computation \& Intelligent Systems and the Faculty of Engineering \& Information Technology, University of Technology, Sydney, Australia \\
              \email{dacheng.tao@uts.edu.au}
           \and
           W. Zhang \at
              the Research Center of Precision Sensing and Control, Institute of Automation, Chinese Academy of Sciences, Beijing, 100190, China \\
              \email{wensheng.zhang@ia.ac.cn}           
           \and
           L. Zhang \at
              the Department of Computing, The Hong Kong Polytechnic University, Hong Kong, China \\
              \email{cslzhang@comp.polyu.edu.hk}
           \and
           Y. Liu \at
              the Department of Computing, The Hong Kong Polytechnic University, Hong Kong, China \\
              \email{csyliu@comp.polyu.edu.hk}
           \and
           Y. Qu \at
              the School of Information Science and Technology, Xiamen University, Fujian, China \\
              \email{yyqu@xmu.edu.cn}
}

\date{Received: date / Accepted: date}

\maketitle

\begin{abstract}
In this paper, we address the multi-view subspace clustering problem. Our method utilizes the circulant algebra for tensor, which is constructed by stacking the subspace representation matrices of different views and then rotating, to capture the low rank tensor subspace so that the refinement of the view-specific subspaces can be achieved, as well as the high order correlations underlying multi-view data can be explored. By introducing a recently proposed tensor factorization, namely tensor-Singular Value Decomposition (t-SVD) \cite{kilmer13}, {\color{red}we can impose a new type of low-rank tensor constraint on the rotated tensor to ensure the consensus among multiple views.} Different from traditional unfolding based tensor norm, this low-rank tensor constraint has optimality properties similar to that of matrix rank derived from SVD, {\color{red}so the complementary information can be explored and propagated among all the views more thoroughly and effectively.} The established model, called t-SVD based Multi-view Subspace Clustering (t-SVD-MSC), falls into the applicable scope of augmented Lagrangian method, and its minimization problem can be efficiently solved with theoretical convergence guarantee and relatively low computational complexity. Extensive experimental testing on eight challenging image dataset shows that the proposed method has achieved highly competent objective performance compared to several state-of-the-art multi-view clustering methods.
\keywords{t-SVD \and Tensor Multi-Rank \and Multi-View Features \and Subspace Clustering}
\end{abstract}

\section{Introduction}\label{intro}
Many scientific data have heterogeneous features, which are collected from diverse domains or generated from various feature extractors. For example, in real-world applications, datasets are naturally comprised of multiple views: a) web-pages can be represented by using both page-text and hyperlinks pointing to them; b) images can be described by different kinds of features, such as color, edge and texture. Each type of feature is referred to as a particular view, and combining multiple views of dataset for data analysis has been a popular practice for improving performance. {\color{red}Commonly, the success of the multi-view learning stems from the following two \textbf{principles}: (1) Consensus principle, which aims to maximize the agreement on multiple distinct views; (2) Complementary principle, which means that each view of the data may contain some knowledge that other views do not have; therefore, multiple views can be employed to comprehensively and accurately describe the data.} For a comprehensive review of multi-view learning, please refer to \cite{multiview-survey}.

In this work, we mainly focus on multi-view clustering, where the absence of a groundtruth to guide the learning process makes the underlining task much harder. {\color{red}\textbf{Basic assumptions of the multi-view clustering}: (1) The feature in each individual view are sufficient to discover most of the clustering information; (2) The feature in each individual view might be corrupted by noise, {\it i.e.,} these noise might result in a small portion of samples being assigned to wrong clusters.} As different views are different representations of the same set of instances, we aim to capture the relationship among multiple views to improve the clustering results generated by the limited information from a single view.

\begin{figure*}[htb]
\setlength{\abovecaptionskip}{0pt}
\setlength{\belowcaptionskip}{0pt}
\renewcommand{\figurename}{Figure}
\centering
\includegraphics[width=0.85\textwidth]{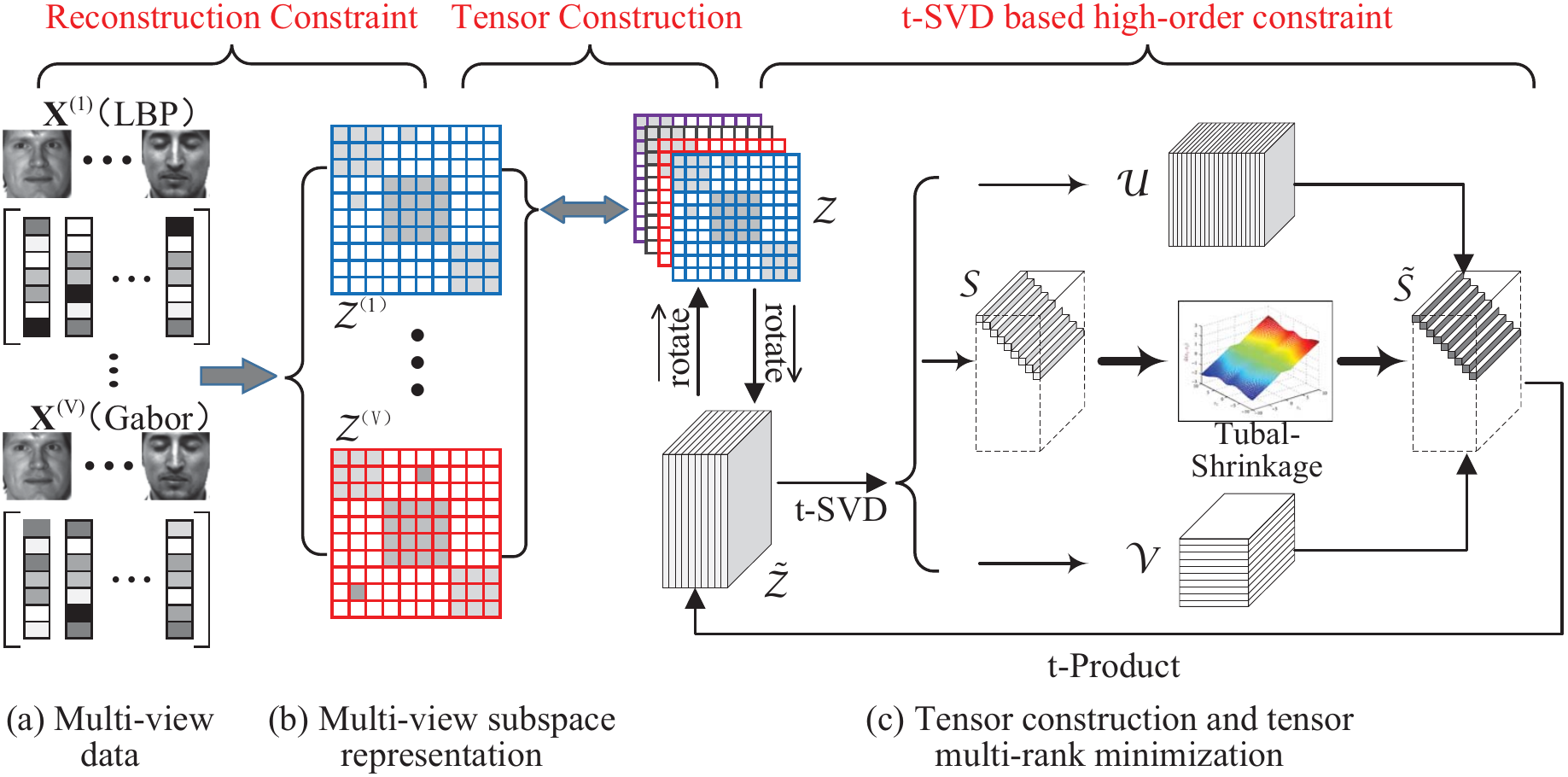}
\caption{The Flowchart of t-SVD-MSC. Given a collection of data points with multi-view representation (a), $\mathbf{X}^{(1)},\ldots, \mathbf{X}^{(V)}$, t-SVD-MSC stacks all the subspace representations (b), $\mathbf{Z}^{(1)},\ldots, \mathbf{Z}^{(V)}$, into a tensor $\boldsymbol{\mathcal{Z}}$, and then rotates to $\boldsymbol{\tilde{\mathcal{Z}}}$; the $\boldsymbol{\tilde{\mathcal{Z}}}$ will be updated by using t-SVD based tensor multi-rank minimization (c).}
\label{fig:overview}
\end{figure*}



Our work is motivated by self-representation based subspace clustering ({\it i.e.,} low-rank representation (LRR) \cite{LRR}) and a new type of factorization for tensor and its approximation problem proposed in \cite{t-SVD-theorem}. While having promising clustering performance, the method \cite{LRR} only {\color{red}considers} the single view feature. Then, the method \cite{LRTM}, which is most relevant to our work, extends the LRR to the multi-view setting by imposing unfolding based low rank \cite{unfolding-lr} (defined in Eqn. (\ref{g-TNN})) on tensor that stacked by the subspace representation matrices from all the views. While easy to implement, different from matrix scenarios, such a simple rank-sum tensor norm is short of a clear physical meaning for general tensors. Furthermore, it tries to model the tensor low rank in the matrix SVD-based vector space, resulting in the loss of optimality in the representation.

By contrast, the high order constraint used in our approach is based on recently proposed tensor-Singular Value Decomposition (t-SVD) and its derived tensor nuclear norm (t-TNN) \cite{kilmer13}. t-SVD has a similar structure to the matrix SVD, and model a tensor in the matrix space through a well-defined t-product operation \cite{t-SVD-theorem}, which can be shown in the theoretical analysis in motivation subsection \ref{motivation} {\color{red} (due to the need for some key notations and preliminaries, we postpone the detailed motivation until section \ref{motivation})}. By applying this well-defined tensor constraint to our multi-view model, a natural physical meaning for low-rank structure underneath tensor can be achieved. More importantly, in our approach, each subspace representation matrix can be considered as a view-specific distance metric learning among different samples but with measurement noisy. {\color{red}The proposed method can filter out the noisy to ensure the {\it \textbf{consensus principle}} implicitly by using t-SVD based tensor multi-rank minimization (see Fig. \ref{fig:overview} (c)).} {\color{red}In summary, for the first time to our knowledge, we introduce a circulant algebra based low-rank tensor constraint to achieve consensus among views and explore complementary principle efficiently and thoroughly, which can be confirmed by our excellent clustering performance presented in Section \ref{sec:experiment}.} 

In this paper, we propose a new multi-view clustering method, namely t-SVD based Multi-view Subspace Clustering (t-SVD-MSC). Fig. \ref{fig:overview} illustrates the flowchart of our method. Given a collection of data points with multiple views $\mathbf{X}^{(1)},\ldots, \mathbf{X}^{(V)}$, t-SVD-MSC can obtain the subspace representation matrices $\mathbf{Z}^{(1)},\ldots, \mathbf{Z}^{(V)}$, and then merge them to construct a $3$-order tensor $\boldsymbol{\mathcal{Z}}$. This tensor needs to be rotated so as to keep self-representation coefficient in Fourier domain, and the detailed merits can be found in Section \ref{motivation}. Subsequently, the rotated tensor $\boldsymbol{\mathcal{\tilde{Z}}}$ is efficiently updated by t-SVD based tensor nuclear norm minimization, such that the high order information hidden among multi-view representation can be captured. After that, each $\mathbf{Z}^{(v)}$ ( $v=1,\ldots, V$) will be updated under the self-reconstruction constraint. This process runs iteratively until convergence is arrived. We need to emphasize here that our contributions are not meant as a simple replacement for the unfolding based tensor norm presented in \cite{LRTM}. The proposed t-SVD-MSC carefully consider the complicated structure of the subspace representation matrices from all the views, so that the subspace coefficients are transformed into Fourier domain; meanwhile the information among different samples and views can be explored by comparing every row (sample-specific) and every column (view-specific) of frontal slices over the third dimension (coefficient-specific), which is the intrinsic property of tensor low rank built upon t-SVD.

The main contributions of this paper are summarized as follows:
\begin{itemize}
\item {\color{red}We propose a new multi-view subspace clustering model, {\it i.e.}, t-SVD-MSC, to effectively ensure the consensus among different views by utilizing a well-founded tensor norm in a unified tensor space, so that the complementary information can be captured and propagated among all the views.}
\item {\color{red}To accommodate the circulant algebra, we design a rotated tensor structure to preserve the self-representation coefficient in Fourier domain, as well as explore the high order correlations by comparing every row (sample-specific) and every column (view-specific) of frontal tensor slices.}
\item We present an efficient optimization algorithm to solve the t-SVD-MSC optimization problem with relatively low computational complexity and theoretical convergence guarantee.
\item We conduct the extensive evaluation of our method on several challenge datasets, where a significant improvement over state-of-the-art MSC approaches is achieved. By incorporating CNN feature as a view, the proposed model has achieved highly competent (even better) performance compared to recent proposed CNN based clustering method on some large-scale datasets.

\end{itemize}

The rest of this paper is organized as follows. Section \ref{sec:related_works} introduces related works. Section \ref{sec:notations} gives the preliminaries on tensors and the notations that will be used throughout the paper. In Section \ref{LRC-MSC}, we motivate the proposed model in detail, give an optimization algorithm to solve it, analyze its computational complexity and convergence, {\color{red}and provide some discussions}. Experimental analysis and completion results are shown in Section \ref{sec:experiment} to verify our method. Finally, we conclude the proposed method in Section \ref{sec:conclusion}.

\section{Related Work}\label{sec:related_works}

Multi-view clustering methods have been extensively studied in recent years, we roughly divide them into three categories in accordance with \cite{multiview-survey}: 1) graph-based approaches, 2) co-training or co-regularized approaches, 3) subspace learning algorithms.

The first stream is the graph-based approaches \cite{graph-based1,graph-based2,graph-based3,graph-based4,graph-based5} which exploit the relationship among different views by using multiple graph fusion strategy. \cite{graph-based1} constructed a bipartite graph underlying the minimizing-disagreement criterion to connect the two-view feature, and then solved standard spectral clustering problem on the bipartite graph. The method \cite{graph-based4} proposed to learn a latent graph transition probability matrix via low-rank and sparse decomposition to handle the noise from different views. Given graphs constructed separately from single view data, \cite{graph-based5} built cross-view tensor product graphs to explore higher order information. Moreover, graph based algorithms is closely related to Multiple Kernel Learning (MKL) technique, in which views are considered as given kernel matrices. The aim is to learn the weighted combination of these kernel and the partitioning simultaneously \cite{MKL1}.

Co-training and co-regularized style methods often construct separate learners on distinct views, then utilize the information in each learner to constrain other views. \cite{co-training1} provided a clustering method by interchanging the partition information among different views. \cite{co-training2} proposed to utilize the spectral embedding from one view to constrain the adjacent matrices in other views. By co-regularizing the clustering hypotheses across views, \cite{co-training3} designed novel spectral clustering objective functions that implicitly combine graphs from multiple views of the data to achieve a better result. In \cite{DiMSC}, authors extended the recent subspace clustering to multi-view domain, and utilized the Hilbert Schmidt Independence Criterion (HSIC) as a co-regularized term to explore the complementarity between views. To cluster the video face by multiple intrinsic cues, \cite{CMV-VFC} considered both the video face pairwise constraints as well as the multi-view consistence, which is a co-regularization term that penalizes the disagreement among different graphs of multiple views, leading to a state-of-the-art performance on several real-world video datasets.

Subspace learning approaches are built on the assumption that all the views are generated from a latent subspace. Its goal is to capture shared latent subspace first and then conduct clustering. The representative methods in this stream are proposed in \cite{subspace1,subspace2}, which applied canonical correlation analysis (CCA) and kernel CCA to project the multi-view high-dimensional data onto a low-dimensional subspace, respectively. By including robust losses to replace the squared loss used in CCA, \cite{subspace5-CCCA} provided a convex reformulation of multi-view subspace learning that enforces conditional independence between views. Inspired by deep representation, \cite{subspace4-deepCCA} proposed a DNN-based model combining CCA and autoencoder-based terms to exploit the deep information from two views. Since those CCA based methods are limited by capability of only handling two-view features, tensor CCA \cite{subspace3-TCCA} generalized CCA to handle the data of an arbitrary number of views by analyzing the covariance tensor of different views.

Besides CCA, the recent proposed subspace clustering methods \cite{MVSC,LRTM} resorted to explore the relationship between samples with self-representation ({\it e.g.,} sparse subspace clustering (SSC) \cite{SSC} and low-rank representation (LRR) \cite{LRR}) in multi-view setting. Our approach is closely related to \cite{LRTM}, which extended the LRR based subspace clustering to multi-view by employing the rank-sum of different mode unfoldings to constrain the subspace coefficient tensor. However, such a kind of tensor constraint lacks a clear physical meaning for general tensor, so that it can not thoroughly explore the complementary information among different views. On the contrary, the high order constraint within our model is built upon a new tensor decomposition scheme \cite{kilmer13,t-SVD-theorem}, which is referred to as t-SVD and has been applied to various tasks, such as image reconstruction and tensor completion \cite{semerci14,zhang14,cyilu-tensor}. Therefore, the proposed model possesses good theoretical properties and clear physical meaning for handling the subspace representation tensor. The detailed motivation will be presented in Section \ref{motivation}.

\section{Notations and Preliminaries}\label{sec:notations}
In this section, we will introduce the notations and give the basic definitions used throughout the paper. We use bold calligraphy letters for tensors, {\it e.g.}, $\boldsymbol{\mathcal{X}}$, bold upper case letters for matrices, {\it e.g.}, $\mathbf{X}$, bold lower case letters for vectors, {\it e.g.}, $\mathbf{x}$, and lower case letters for the entries, {\it e.g.}, $x_{ij}$. The Frobenius norm of a matrix $\mathbf{X}$ is defined as $||\mathbf{X}||_{F} := (\sum_{i,j}|x_{ij}|^{2})^{\frac{1}{2}}$. Let $\mathbf{X} = \mathbf{U} \mathbf{\Sigma} \mathbf{V}^{\mathrm{T}}$ be the SVD of $\mathbf{X}$ and $\sigma_{i}(\mathbf{X})$ the $i$th largest singular value, then the matrix nuclear norm of $X$ is $||\mathbf{X}||_{\ast} := \sum_{i}\sigma_{i}(\mathbf{X})$. The corresponding singular-value thresholding (SVT) operation with threshold $\tau$ is $\mathbf{\mathcal{D}}_{\tau}(\mathbf{X})=\mathbf{U} \mathbf{\Sigma}_{\tau}\mathbf{V}^{\mathrm{T}}$, where $\Sigma_{\tau} = \mathrm{diag}\left\{\left(\sigma_{i}(\mathbf{X})-\tau \right)_{+}\right\}$ and $t_{+}$ is the positive part of $t$.

An $N$-way (or $N$-mode) tensor is a multi-linear structure in $\mathbb{R}^{n_{1} \times n_{2} \times \ldots \times n_{N}}$. A \textbf{slice} of an tensor is a 2D section defined by fixing all but two indices, and a \textbf{fiber} is a 1D section defined by fixing all indices but one \cite{kolda09}. For a 3-way tensor $\boldsymbol{\mathcal{X}}$, we use the Matlab notation $\boldsymbol{\mathcal{X}}(k, :, :)$, $\boldsymbol{\mathcal{X}}(:, k, :)$ and $\boldsymbol{\mathcal{X}}(:, :, k)$ to denote the $k$th horizontal, lateral and frontal slices, respectively; $\boldsymbol{\mathcal{X}}(:, i, j)$, $\boldsymbol{\mathcal{X}}(i, :, j)$ and $\boldsymbol{\mathcal{X}}(i, j, :)$ to denote the mode-1, mode-2 and mode-3 fibers, and $\boldsymbol{\mathcal{X}}_{f} = \mathrm{fft}(\boldsymbol{\mathcal{X}},[~],3)$ to denote the Fourier transform along the third dimension. In particular, $\boldsymbol{\mathcal{X}}^{(k)}$ is used to represent ${\boldsymbol{\mathcal{X}}}(:, :, k)$. Unfolding the tensor $\boldsymbol{\mathcal{X}}$ along the $l$th mode defined as $\mathrm{unfold}_{l}(\boldsymbol{\mathcal{X}}) = \mathbf{X}_{(l)} \in \mathbb{R}^{n_{l} \times \prod_{l'\neq l} n_{l'}}$, which is a matrix whose columns are mode-$l$ fibers \cite{kolda09}. The opposite operation ``fold'' of the unfolding is defined as $\mathrm{fold}_{l}({\mathbf{X}}_{(l)}) = {\boldsymbol{\mathcal{X}}}$. The Frobenius norm of ${\boldsymbol{\mathcal{X}}}$ is $||{\boldsymbol{\mathcal{X}}}||_{F} :=  (\sum_{i,j,k}|x_{ijk}|^{2})^{\frac{1}{2}}$, and the $l_{1}$ norm of ${\boldsymbol{\mathcal{X}}}$ is $||{\boldsymbol{\mathcal{X}}}||_{1} := \sum_{i,j,k}|x_{ijk}|$.

Before introducing the t-SVD and its derived tensor nuclear norm, it is necessary to define five block-based operators, i.e., $\mathrm{bcirc}$, $\mathrm{bvec}$, $\mathrm{bvfold}$, $\mathrm{bdiag}$ and $\mathrm{bdfold}$ \cite{kilmer13}. For $\boldsymbol{{\mathcal{X}}} \in \mathbb{R}^{n_{1} \times n_{2} \times n_{3}}$ specially, the $\boldsymbol{{\mathcal{X}}}^{(k)}$s can be used to form the block circulant matrix:
\begin{equation}\label{fml:bcm}
\mathrm{bcirc}(\boldsymbol{{\mathcal{X}}}) :=
\left[
\begin{matrix}
 \boldsymbol{{\mathcal{X}}}^{(1)}    &\boldsymbol{{\mathcal{X}}}^{(n_{3})}  & \cdots         & \boldsymbol{{\mathcal{X}}}^{(2)} \\
 \boldsymbol{{\mathcal{X}}}^{(2)}    &\boldsymbol{{\mathcal{X}}}^{(1)}      &  \cdots        & \boldsymbol{{\mathcal{X}}}^{(3)} \\
 \vdots            &\ddots              & \ddots         & \vdots         \\
 \boldsymbol{{\mathcal{X}}}^{(n_{3})}&\boldsymbol{{\mathcal{X}}}^{(n_{3}-1)}& \cdots         & \boldsymbol{{\mathcal{X}}}^{(1)}
\end{matrix}
\right],
\end{equation}
the block vectorizing and its opposite operation
\begin{equation}\label{fml:bvec_bvfold}
\mathrm{bvec}({\boldsymbol{{\mathcal{X}}}}) :=
\left[
\begin{matrix}
 \boldsymbol{{\mathcal{X}}}^{(1)} \\
 \boldsymbol{{\mathcal{X}}}^{(2)} \\
 \vdots         \\
 \boldsymbol{{\mathcal{X}}}^{(n_{3})}
\end{matrix}
\right],~~~
\mathrm{bvfold}(\mathrm{bvec}(\boldsymbol{{\mathcal{X}}})) = \boldsymbol{{\mathcal{X}}},
\end{equation}
and the block diag matrix and its opposite operation
\begin{equation}
\label{fml:bdiag}
\mathrm{bdiag}({\mathcal{X}}) :=
\left[
\begin{matrix}
\boldsymbol{{\mathcal{X}}}^{(1)} & & \\
 &\ddots & \\
 && \boldsymbol{{\mathcal{X}}}^{(n_{3})}
\end{matrix}
\right],~~~
\mathrm{bdfold}(\mathrm{bdiag}(\boldsymbol{{\mathcal{X}}})) = \boldsymbol{{\mathcal{X}}}.
\end{equation}

\subsection{Tensor Singular Value Decomposition (t-SVD)}
To help understand the t-SVD, the following related notions, which are defined in \cite{kilmer13}, need to be introduced. The t-product between two $3$-mode tensors is defined as follows:
\begin{definition}[\textbf{t-product}]\label{def:t-prod}
Let $\boldsymbol{{\mathcal{X}}}$ be $n_{1} \times n_{2} \times n_{3}$, and $\boldsymbol{{\mathcal{Y}}}$ be $n_{2} \times n_{4} \times n_{3}$. The t-product $\boldsymbol{{\mathcal{X}}}*\boldsymbol{{\mathcal{Y}}}$ is an $n_{1} \times n_{4} \times n_{3}$ tensor
\begin{equation}
\label{fml:t-prod}
\boldsymbol{{\mathcal{M}}} = \boldsymbol{{\mathcal{X}}}*\boldsymbol{{\mathcal{Y}}} = : \mathrm{bvfold}\{\mathrm{bcirc}(\boldsymbol{{\mathcal{X}}})\mathrm{bvec}(\boldsymbol{{\mathcal{Y}}})\}.
\end{equation}
\end{definition}
The t-product is analogous to the matrix multiplication except that the {\it circular convolution} replaces the multiplication operation between the elements, which are now mode-3 fibers \cite{zhang14}, as follows:
\begin{equation}\label{fml:t-prod2}
\boldsymbol{{\mathcal{M}}}(i,j,:) = \sum_{k=1}^{n_{2}}{\boldsymbol{{\mathcal{X}}}}(i,k,:) \circ {\boldsymbol{{\mathcal{Y}}}}(k,j,:),
\end{equation}
where $\circ$ denotes the circular convolution between two tubes. The t-product in the original domain corresponds to the matrix multiplication of the frontal slices in the Fourier domain, as follows :
\begin{equation}
\label{fml:t-prod3}
\boldsymbol{{\mathcal{M}}}_{f}^{(k)} = \boldsymbol{{\mathcal{X}}}_{f}^{(k)} \boldsymbol{{\mathcal{Y}}}_{f}^{(k)},~ k = 1,\ldots,n_{3},
\end{equation}

\begin{definition}[\textbf{Tensor Transpose}]
\label{def:t-trans}
Let $\boldsymbol{{\mathcal{X}}} \in \mathbb{R}^{ n_{1} \times n_{2} \times n_{3} }$, the transpose tensor $\boldsymbol{{\mathcal{X}}}^{\mathrm{T}}$ is an $n_{2} \times n_{1} \times n_{3}$ tensor obtained by transposing each frontal slice of $\boldsymbol{{\mathcal{X}}}$ and then reversing the order of the transposed frontal slices 2 through $n_3$.
\end{definition}

\begin{definition}[\textbf{Identity Tensor}]
\label{def:t-I}
The identity tensor $\boldsymbol{{\mathcal{I}}} \in \mathbb{R}^{n_{1} \times n_{1} \times n_{3}}$ is a tensor whose first frontal slice is the $n_{1} \times n_{1}$ identity matrix and all other frontal slices are zero.
\end{definition}

\begin{definition}[\textbf{Orthogonal Tensor}]
\label{def:orth-tensor}
A tensor $\boldsymbol{{\mathcal{Q}}} \in \mathbb{R}^{n_{1} \times n_{1} \times n_{3}}$ is orthogonal if
\begin{equation}
\boldsymbol{{\mathcal{Q}}}^{\mathrm{T}}*\boldsymbol{{\mathcal{Q}}} = \boldsymbol{{\mathcal{Q}}}*\boldsymbol{{\mathcal{Q}}}^{\mathrm{T}} = \boldsymbol{{\mathcal{I}}},
\end{equation}
where $*$ is the t-product.
\end{definition}

\begin{figure}[htb]
\setlength{\abovecaptionskip}{0pt}  
\setlength{\belowcaptionskip}{0pt} 
\renewcommand{\figurename}{Figure}
\centering
\includegraphics[width=0.5\textwidth]{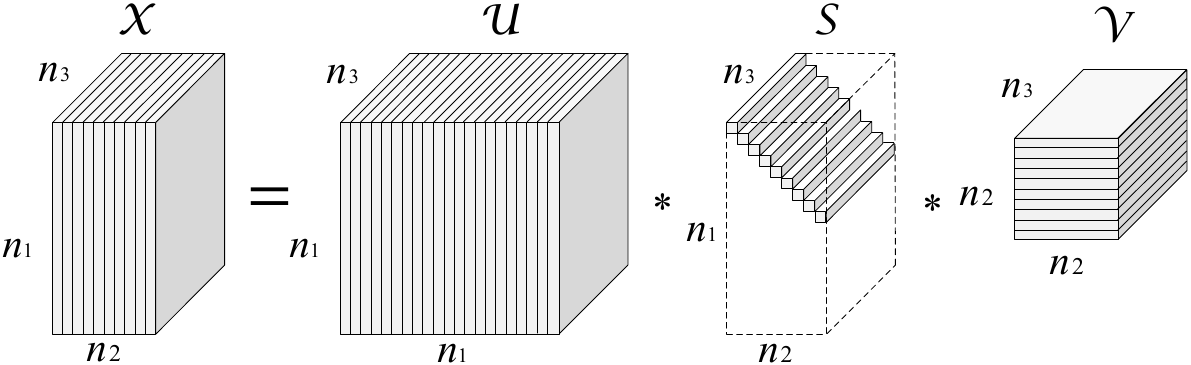}
\caption{The t-SVD of an $n_{1} \times n_{2} \times n_{3}$ tensor.}
\label{fig:tSVD}
\end{figure}

\begin{definition}[\textbf{f-diagonal Tensor}]
\label{def:f-diag}
A tensor is called f-diagonal if each of its frontal slices is diagonal matrix. The t-production of two f-diagonal tensors with the same size $n_{1} \times n_{2} \times n_{3}$, i.e., $\boldsymbol{{\mathcal{M}}}=\boldsymbol{{\mathcal{X}}}*\boldsymbol{{\mathcal{Y}}}$, is also an $n_{1} \times n_{2} \times n_{3}$ f-diagonal tensor, and its diagonal tube fibers are
\begin{equation}
\label{fml:tprod_fdiag}
\boldsymbol{{\mathcal{M}}}(i,i,:) = \boldsymbol{{\mathcal{X}}}(i,i,:) \circ \boldsymbol{{\mathcal{Y}}}(i,i,:),~ i = 1, \ldots, \mathrm{min}(n_{1}, n_{2}).
\end{equation}
\end{definition}
Given the aforementioned definitions, the tensor Singular Value Decomposition (t-SVD) of $\boldsymbol{{\mathcal{X}}} $ is given by
\begin{equation}\label{fml:t-svd}
\boldsymbol{{\mathcal{X}}}=\boldsymbol{{\mathcal{U}}}*\boldsymbol{{\mathcal{S}}}*\boldsymbol{{\mathcal{V}}}^{\mathrm{T}},
\end{equation}
where $\boldsymbol{{\mathcal{U}}}$ and $\boldsymbol{{\mathcal{V}}}$ are orthogonal tensors of size $n_{1} \times n_{1} \times n_{3}$ and $n_{2} \times n_{2} \times n_{3}$ respectively. $\boldsymbol{{\mathcal{S}}}$ is an f-diagonal tensor of size $n_{1} \times n_{2} \times n_{3}$, and $*$ denotes the t-product. Fig. \ref{fig:tSVD} illustrates the decomposition. As demonstrated in Eq. (\ref{fml:t-prod3}), the t-production can be computed efficiently in the Fourier domain, which leads to Algorithm \ref{alg:tSVD}.

\begin{algorithm}
\SetAlgoLined
\caption{t-SVD \cite{kilmer13}}\label{alg:tSVD}
\KwIn{~~~$\boldsymbol{\mathcal{X}} \in \mathbb{R}^{n_{1} \times n_{2} \times n_{3}}$;\\}
\KwOut{~$\boldsymbol{\mathcal{U}}$,~$\boldsymbol{\mathcal{S}}$,~$\boldsymbol{\mathcal{V}}$;\\}
\BlankLine
$\boldsymbol{\mathcal{X}}_{f} = \mathrm{fft}(\boldsymbol{\mathcal{X}},[~],3)$;\\
\For{$k = 1 : n_{3}$}
{
$[\mathbf{U},\mathbf{\Sigma},\mathbf{V}] = \mathrm{SVD}(\boldsymbol{\mathcal{X}}_{f}^{(k)})$;\\
$\boldsymbol{\mathcal{U}}_{f}^{(k)} = \mathbf{U}$,~$\boldsymbol{\mathcal{S}}_{f}^{(k)} = \mathbf{\Sigma}$,~$\boldsymbol{\mathcal{V}}_{f}^{(k)} = \mathbf{V}$;\\
}
$\boldsymbol{\mathcal{U}} = \mathrm{ifft}({\cal U}_{f},[~],3)$,~$\boldsymbol{\mathcal{S}} = \mathrm{ifft}({\cal S}_{f},[~],3)$,~ $\boldsymbol{\mathcal{V}} = \mathrm{ifft}({\cal V}_{f},[~],3)$;\\
\BlankLine
\textbf{Return} $\boldsymbol{\mathcal{U}}$, $\boldsymbol{\mathcal{S}}$, $\boldsymbol{\mathcal{V}}$.
\end{algorithm}

\subsection{Tensor Nuclear Norm via t-SVD}\label{t-TNN}
The t-SVD allows the tensor $\boldsymbol{\mathcal{X}}$ to be written as a finite sum of outer product of matrices \cite{t-SVD-theorem}:
\begin{equation}\label{outer-product}
    \boldsymbol{\mathcal{X}} = \sum_{i=1}^{\min(n_1, n_2)} \boldsymbol{\mathcal{U}}(:,i,:)\ast \boldsymbol{\mathcal{S}}(i,i,:)\ast \boldsymbol{\mathcal{V}}(:,i,:)^{\mathrm{T}},
\end{equation}
which is equivalent to the following equation in the Fourier domain \cite{t-SVD-theorem}:
\begin{equation}
\begin{aligned}
\label{fml:bdiag}
\left[
\begin{matrix}
{\boldsymbol{\mathcal{X}}}^{(1)}_f & & \\
 &\ddots & \\
 && {\boldsymbol{\mathcal{X}}}^{(n_{3})}_f
\end{matrix}
\right] =
\left[
\begin{matrix}
{\boldsymbol{\mathcal{U}}}^{(1)}_f & & \\
 &\ddots & \\
 && {\boldsymbol{\mathcal{U}}}^{(n_{3})}_f
\end{matrix}
\right] \cdot \\
\left[
\begin{matrix}
{\boldsymbol{\mathcal{S}}}^{(1)}_f & & \\
 &\ddots & \\
 && {\boldsymbol{\mathcal{S}}}^{(n_{3})}_f
\end{matrix}
\right]\cdot
\left[
\begin{matrix}
{\boldsymbol{\mathcal{V}}}^{(1)}_f & & \\
 &\ddots & \\
 && {\boldsymbol{\mathcal{V}}}^{(n_{3})}_f
\end{matrix}
\right]^{\mathrm{T}}.
\end{aligned}
\end{equation}
where $\cdot$ denotes common matrix product, and we have $n_3$ blocks matrix SVD: $\boldsymbol{\mathcal{X}}^{(i)}_f = \boldsymbol{\mathcal{U}}^{(i)}_f \boldsymbol{\mathcal{S}}^{(i)}_f (\boldsymbol{\mathcal{V}}^{(i)}_f)^{\mathrm{T}}, i=1,\ldots,n_3$. Now, we can define the tensor multi-rank as follows \cite{kilmer13,zhang14,semerci14} :
\begin{definition}[\textbf{Tensor multi-rank}]
\label{def:multi-rank}
The multi-rank of ${\boldsymbol{\mathcal{X}}} \in \mathbb{R}^{n_{1} \times n_{2} \times n_{3}}$ is a vector $\mathbf{r} \in \mathbb{R}^{n_{3} \times 1}$ with the $i$-th element equal to the rank of the $i$-th frontal slice of ${\boldsymbol{\mathcal{X}}}_{f}$.
\end{definition}

Then the t-SVD based tensor nuclear norm (t-TNN) is given as
\begin{equation}
\label{fml:gtnn}
||\boldsymbol{\mathcal{X}}||_{\circledast} :=\sum_{i=1}^{\mathrm{min}(n_{1},n_{2})}\sum_{k=1}^{n_{3}}|{ \boldsymbol{\mathcal{S}}}_{f}(i,i,k)|,
\end{equation}
which is proven to be a valid norm and the tightest convex relaxation to $\ell_{1}$ norm of the tensor multi-rank in  \cite{semerci14,zhang14}. Due to the unitary invariance of matrix nuclear norm, we have
\begin{equation}
\label{fml:mf_nuclear_norm}
||\mathrm{bdiag}(\boldsymbol{\mathcal{X}}_{f})||_{*} = ||\mathrm{bdiag}(\boldsymbol{\mathcal{S}}_{f})||_{*} = ||{\boldsymbol{\mathcal{X}}}||_{\circledast},
\end{equation}
and since block circulant matrixes can be block diagonalized by using the Fourier transform, there is
\begin{equation}
\label{fml:mf-bcirc}
\begin{aligned}
||\mathrm{bdiag}(\boldsymbol{\mathcal{X}}_{f})||_{*} & = ||(\mathbf{F}_{n_{3}}\otimes \mathbf{I}_{n_{1}}) \mathrm{bcirc}(\boldsymbol{\mathcal{X}}) (\mathbf{F}_{n_{3}}^{*}\otimes \mathbf{I}_{n_{2}})||_{*}\\
                                         &=|| \mathrm{bcirc}(\boldsymbol{\mathcal{X}})||_{*}.
\end{aligned}
\end{equation}
where, $\otimes$ denotes the Kronecker product, $\mathbf{F}_{n}$ is the $n\times n$ Discrete Fourier Transform (DFT) matrix, and $\mathbf{I}_{n}$ is an $n\times n$ identity matrix. Finally, we obtain
\begin{equation}
\label{fml:gtnn_birc}
||\boldsymbol{\mathcal{X}}||_{\circledast} = || \mathrm{bcirc}(\boldsymbol{\mathcal{X}})||_{*}.
\end{equation}
The equivalence in Eq. (\ref{fml:gtnn_birc}) endows the t-TNN with interpretability in the original domain, {\it i.e.,} $|| \mathrm{bcirc}(\boldsymbol{\mathcal{X}})||_{*}$ measures the rank of $\mathrm{bcirc}(\boldsymbol{\mathcal{X}})$ by comparing every row and every column of frontal slices over the third dimension, which exploits structural information of a tensor deeper than the monotonous matrix nuclear norm of certain unfolding.

\section{The Proposed Approach}\label{LRC-MSC}
Subspace clustering is a technology for clustering data according to the underlying subspaces. In the paper, we consider the self-representation based subspace clustering method, specifically the LRR approach \cite{LRR}, which constructs affinity matrix through reconstruction coefficients, as well as explores low-dimensional subspace structures embedded in data. Suppose $\mathbf{X} = [\mathbf{x}_1, \mathbf{x}_2, \ldots, \mathbf{x}_N] \in \mathbb{R}^{d\times N}$ is the matrix of data vectors, whose column is a sample vector, and $d$ is the dimensionality of the feature space. Formally, LRR solves the following optimization problem:
\begin{equation}\label{reduced-LRR}
    \begin{aligned}
    &\min_{\mathbf{Z}, \mathbf{E}} \lambda ||\mathbf{E}||_{2,1} + ||\mathbf{Z}||_{\ast},\\
    &\text{s.t.} \quad \mathbf{X} = \mathbf{X}\mathbf{Z} + \mathbf{E},
    \end{aligned}
\end{equation}
where $\mathbf{Z} = [\mathbf{z}_1, \mathbf{z}_2, \ldots, \mathbf{z}_N] \in \mathbb{R}^{N\times N}$ is the coefficient matrix with each $\mathbf{z}_i$ being the new representation of sample $\mathbf{x}_i$, and $||\cdot||_{\ast}$ is the nuclear norm, $||\cdot||_{2,1}$ denotes the $\ell_{2,1}$-norm of a matrix. After achieving the self-representation matrix $\mathbf{Z}$, the affinity matrix $\mathbf{A}$ is usually constructed as
\begin{equation}\label{affinity-matrix}
    \mathbf{A} = \frac{1}{2}\left(|\mathbf{Z}| + |\mathbf{Z}^{T}|\right),
\end{equation}
where $|\cdot|$ represents the absolute operator. Then, the obtained affinity matrix $\mathbf{A}$ will be sent to a spectral clustering algorithm \cite{spectral-clustering} to produce the final clustering result.

Intuitively, the above single view subspace clustering method can be extended to the multi-view setting in a simple and direct way. We use $\mathbf{X}^{(v)}$ to denote the feature matrix corresponding to the $v$-th view, and use $\mathbf{Z}^{(v)}$ to represent the $v$-th view's learned subspace representation. Hence, the objective function of the LRR based naive multi-view subspace clustering turns out to be:
\begin{equation}\label{naive-msc-lrr}
    \begin{aligned}
    &\min_{\mathbf{Z}^{(v)}, \mathbf{E}^{(v)}} \sum_{v}^{V} \big(\lambda_v ||\mathbf{E}^{(v)}||_{2,1} + ||\mathbf{Z}^{(v)}||_{\ast}\big),\\
    &\text{s.t.} \quad \mathbf{X}^{(v)} = \mathbf{X}^{(v)}\mathbf{Z}^{(v)} + \mathbf{E}^{(v)}, v=1,2,\ldots,V,
    \end{aligned}
\end{equation}
where $V$ denotes the number of all views. After obtaining $\left\{\mathbf{Z}^{(v)}\right\}_{v=1}^{V}$, the final affinity matrix is calculated by combining all subspace representation of each view: $\mathbf{A} = \frac{1}{V}\sum_{v=1}^{V}(|\mathbf{Z}^{(v)}| + |\mathbf{Z}^{(v)^{\mathbf{T}}}|)/2$. However, this formulation treats each subspace representation independently, {\color{red}ignoring the relationship among different views}. To overcome this drawback, we propose to utilize t-SVD based tensor nuclear norm to capture the high order correlations among different views.

\subsection{Motivation}\label{motivation}

{\color{red}
\subsubsection{Ensuring Consensus Principle among Views}

Recall the LRR \cite{LRR}, it can achieve the self-representation coefficient matrix $\mathbf{Z}\in \mathbb{R}^{N\times N}$ by representing the data samples as linear combinations of the bases in a given dictionary (usually, the whole dataset itself). In other words, LRR leads to dense representation coefficients within the same subspace. When we employ multiple features to describe the data, we will have multiple self-representation coefficient matrix $\{\mathbf{Z}^{(v)}\}_{v=1}^{V}$ correspondingly. Not only should we keep the low rank constraint for each $\mathbf{Z}^{(v)}$, but also need to ensure the consensus principle by imposing low rank across all views. The proposed approach is capable of modeling those two level low rank constraints in a unified tensor space by imposing the t-TNN. Consequently, after the optimization, all the $\{\mathbf{Z}^{(v)}\}_{v=1}^{V}$ are much more close to well structure, which means that the fused $\mathbf{Z} = \frac{1}{V}\sum_{v=1}^{V}(|\mathbf{Z}^{(v)}| + |\mathbf{Z}^{(v)^{\mathbf{T}}}|)/2$ can be easily segmented by common spectral clustering method.}

{\color{red}
\subsubsection{Requiring a Well-Founded Low Rank Constraint in Tensor Space}}

To extend the self-representation based subspace clustering to multi-view setting, \cite{LRTM} introduced a low-rank tensor constraint \cite{unfolding-lr}, which directly extended the matrix nuclear norm to higher-order case:
\begin{equation}\label{g-TNN}
    ||\boldsymbol{\mathcal{Z}}||_{\ast} = \sum_{m=1}^{3} \xi_{m}||\mathbf{Z}_{(m)}||_{\ast},
\end{equation}
where the weight $\xi_{m}$ needs to satisfy $\xi_m > 0$ and $\sum_{m=1}^{3} \xi_m = 1$ ($\xi_1 = \xi_2 = \xi_3$ is used in \cite{LRTM}), $\boldsymbol{\mathcal{Z}}$ is a 3-order tensor constructed by merging different $\mathbf{Z}^{(v)}$ along the third dimension, and $\mathbf{Z}_{(m)}$ is the unfolding matrix along the $m$-th mode. We refer to it as the {\it generalized tensor nuclear norm} (g-TNN). Albeit easy to implement, different from matrix scenarios, such a simple rank-sum term is short of a clear physical meaning for general tensors. Besides, the strategy of using the same weights to penalize all dimensionality ranks of a tensor is not always rational. However, incorporating the t-TNN in the proposed model possesses obvious and clear physical meaning. {\color{red}We introduce the following theorem to theoretically explain why the t-TNN is adopted in the proposed model.}

\begin{theorem}\label{theory:truncated-tsvd}
{\it \cite{t-SVD-theorem}} Let the t-SVD of $\boldsymbol{\mathcal{A}}\in \mathbb{R}^{n_1 \times n_2 \times n_3}$ be given by $\boldsymbol{\mathcal{A}} = {\boldsymbol{\mathcal{U}}}*{\boldsymbol{\mathcal{S}}}*{\boldsymbol{\mathcal{V}}}^{\mathrm{T}}$, and for $k<\min(n_1, n_2)$ define $\boldsymbol{\mathcal{A}}_k = \sum_{i=1}^{k} \boldsymbol{\mathcal{U}}(:,i,:)*{\boldsymbol{\mathcal{S}}(i,i,:)}*{\boldsymbol{\mathcal{V}}(:,i,:)}^{\mathrm{T}}$, then \begin{equation}\label{}
    \boldsymbol{\mathcal{A}}_k = \argmin_{\boldsymbol{\mathcal{\widetilde{A}}}\in \mathbb{M}} ||\boldsymbol{\mathcal{A}} - \boldsymbol{\mathcal{\widetilde{A}}}||_{F}
\end{equation}
where $\mathbb{M} = \{\boldsymbol{\mathcal{C}} = \boldsymbol{\mathcal{X}} \ast \boldsymbol{\mathcal{Y}}| \boldsymbol{\mathcal{X}} \in \mathbb{R}^{n_1 \times k \times n_3}, \boldsymbol{\mathcal{Y}} \in \mathbb{R}^{k \times n_2 \times n_3}\}$.
\end{theorem}
Theorem \ref{theory:truncated-tsvd} indicates that a truncated t-SVD representation could provide an optimal approximation in the same way as the truncated matrix SVD, which gives a best low rank approximation to the matrix in terms of the Frobenius norm under rank $k$ constraint. Moreover, matrix nuclear norm is the tightest convex relaxation of the original rank minimization, while t-TNN also has been proven to be the tightest convex relaxation to $\ell_1$ norm of the tensor multi-rank (see Section \ref{t-TNN}) \cite{zhang14}. Theoretically, the t-TNN is {\it \textbf{more analogous}} to the matrix nuclear norm than the g-TNN defined in (\ref{g-TNN}).

\begin{figure}[htb]
\setlength{\abovecaptionskip}{0pt}  
\setlength{\belowcaptionskip}{0pt} 
\renewcommand{\figurename}{Figure}
\centering
\includegraphics[width=0.5\textwidth]{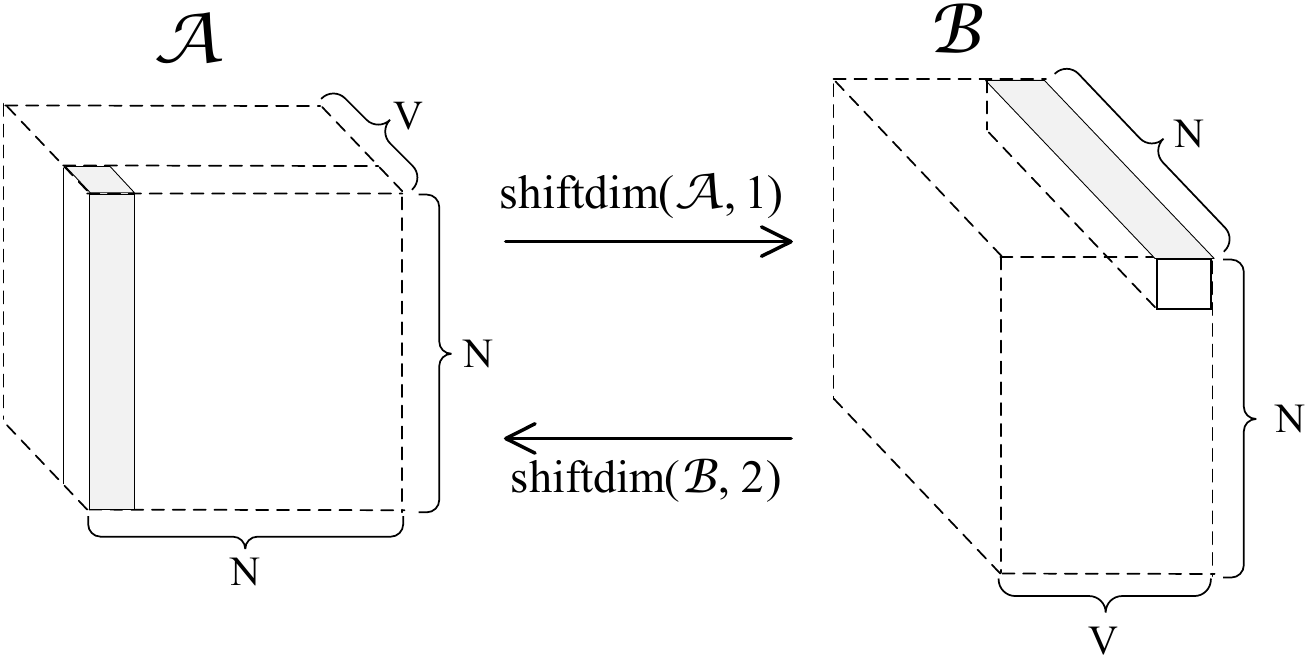}
\caption{The rotated coefficient tensor in our approach.}
\label{fig:shift-tSVD}
\end{figure}

{\color{red}
\subsubsection{Constructing a Structure for Tensor Circulant Algebra}

\noindent Directly utilizing the t-TNN to model the low rank constraint is still far from effectiveness, which can be evidenced by observing the performance of Ut-SVD-MSC method in experimental section. To accommodate the intrinsic circulant algebra underlying t-TNN, we choose to transform the self-represented coefficient (mode-1 fiber) into the mode-3 fiber by using \emph{the rotation operation}\footnote{The tensor rotation in Matlab can be achieved by using the command ``shiftdim''.} of the coefficient tensor, as illustrated in Fig. \ref{fig:shift-tSVD}, where the marked fiber denotes a self-represented feature coefficient of a certain sample belonging to a certain view.}

While relatively simple, the proposed model will benefit from the rotate operation in three aspects. First of all, through tensor rotation, the self-representation coefficient can be preserved in Fourier domain, since the Fourier transform along the third dimension. Secondly, each frontal slice in Fourier domain considers the information among different samples and different views. By measuring every row and every column of frontal slices over the third dimension, the t-TNN provides a deeper insight into multi-view feature tensor than g-TNN. Another advantage of this rotate operation is the significant reduction of computational complexity, which will be analyzed in Section \ref{converge-complexity} and Section \ref{convergence-and-complexity}. {\color{red}To sum up, the aforementioned satisfaction of principle, good theoretical properties, and well-designed tensor structure motivate us to design the proposed t-SVD-MSC model.}

\subsection{Problem Formulation}

The objective function of the proposed method is:
\begin{equation}\label{original-problem}
    \begin{aligned}
    &\min_{\mathbf{Z}^{(v)}, \mathbf{E}^{(v)}} \lambda ||\mathbf{E}||_{2,1} + ||\boldsymbol{ \mathcal{Z}}||_{\circledast},\\
    &\text{s.t.} \quad \mathbf{X}^{(v)} = \mathbf{X}^{(v)}\mathbf{Z}^{(v)} + \mathbf{E}^{(v)}, v=1,\ldots, V,\\
    &\qquad \boldsymbol{\mathcal{Z}} = \Phi(\mathbf{Z}^{(1)}, \mathbf{Z}^{(2)}, \ldots, \mathbf{Z}^{(V)}),\\
    &\qquad \mathbf{E} = [\mathbf{E}^{(1)}; \mathbf{E}^{(2)}; \ldots, \mathbf{E}^{(V)}],
    \end{aligned}
\end{equation}
where the function $\Phi(\cdot)$ constructs the tensor $\boldsymbol{\mathcal{Z}}$ by merging different representation $\mathbf{Z}^{(v)}$ to a $3$-mode tensor, and then rotate its dimensionality to $N\times V \times N$, as shown in Fig. \ref{fig:shift-tSVD}. Also, we can easily get the following relationship:
\begin{equation}\label{tensor-Z2z}
    \Phi_{(v)}^{-1}(\boldsymbol{\mathcal{Z}}) = \mathbf{Z}^{(v)},
\end{equation}
where $\Phi^{-1}(\cdot)$ denotes the inverse function of $\Phi(\cdot)$, and its subscript $(v)$ means to extract the $v$-th frontal slice. As suggested in \cite{LRR}, the vertical concatenation along the column of error matrix, {\it i.e.,} $\mathbf{E} = [\mathbf{E}^{(1)}; \mathbf{E}^{(2)}; \ldots, \mathbf{E}^{(V)}]$, can enforce the column of $\mathbf{E}^{(v)}$ in each view to have jointly consistent magnitude values. Consequently, the objective function in Eq. (\ref{original-problem}) aims to find the optimal self-representations through capturing the informational and structural complexity of multi-view features.

The above optimization problem can be solved by using the Augmented Lagrange Multiplier (ALM) \cite{alm-lin}. To adopt alternating direction minimizing strategy to problem (\ref{original-problem}), we need to make the objective function seperable. By introducing the auxiliary tensor variable $\boldsymbol{\mathcal{G}}$, the optimization problem can be transferred to minimize the following unconstrained problem:
\begin{equation}\label{t-SVD-MSC}
    \begin{aligned}
    &\boldsymbol{\mathcal{L}} (\mathbf{Z}^{(v)}, \ldots, \mathbf{Z}^{(V)}; \mathbf{E}^{(1)}, \ldots, \mathbf{E}^{(V)};\boldsymbol{\mathcal{G}} ) \\
    &= \lambda ||\mathbf{E}||_{2,1} + ||\boldsymbol{ \mathcal{G}}||_{\circledast} + \sum_{v=1}^{V} \bigg(\langle \mathbf{Y}_{v}, \mathbf{X}^{(v)} - \mathbf{X}^{(v)}\mathbf{Z}^{(v)} - \mathbf{E}^{(v)} \rangle \\
    &+ \frac{\mu}{2} ||\mathbf{X}^{(v)} - \mathbf{X}^{(v)}\mathbf{Z}^{(v)} - \mathbf{E}^{(v)}||_{F}^{2} \bigg)+ \langle \boldsymbol{\mathcal{W}}, \boldsymbol{\mathcal{Z} - \boldsymbol{\mathcal{G}}} \rangle\\
    &+ \frac{\rho}{2} ||\boldsymbol{\mathcal{Z} - \boldsymbol{\mathcal{G}}}||_{F}^{2}.
    \end{aligned}
\end{equation}
where the matrix $\mathbf{Y}_{v}$ and the tensor $\boldsymbol{\mathcal{W}}$ represent two Lagrange multipliers, $\mu$ and $\rho$ are actually the penalty parameters, which are adjusted by using adaptive updating strategy as suggested in \cite{ladmap}. The optimization problem (\ref{t-SVD-MSC}) seems challenging to solve, not only because of the t-TNN on $\boldsymbol{\mathcal{G}}$, but also since the tensor $\boldsymbol{\mathcal{Z}}$ depends on the subspace representation of all views.

\subsection{Optimization Procedure}
The alternative minimization scheme is adopted for updating $\mathbf{Z}^{(v)}$, $\mathbf{E}^{(v)}$, and $\boldsymbol{\mathcal{G}}$, respectively. The detailed procedure can be partitioned into three steps alternatingly.

\textbf{$\mathbf{Z}^{(v)}$-subproblem:} When $\mathbf{E}$ and $\boldsymbol{\mathcal{G}}$ are fixed, since $\Phi_{(v)}^{-1}(\boldsymbol{\mathcal{W}}) = \mathbf{W}^{(v)}$ and $\Phi_{(v)}^{-1}(\boldsymbol{\mathcal{G}}) = \mathbf{G}^{(v)}$, we will solve the following subproblem for updating the subspace representation $\mathbf{Z}^{(v)}$:
\begin{equation}\label{z-subproblem}
    \begin{aligned}
    \min_{\mathbf{Z}^{(v)}} &\quad \langle \mathbf{Y}_{v}, \mathbf{X}^{(v)} - \mathbf{X}^{(v)}\mathbf{Z}^{(v)} - \mathbf{E}^{(v)} \rangle + \frac{\mu}{2} ||\mathbf{X}^{(v)} - \mathbf{X}^{(v)}\mathbf{Z}^{(v)}\\
    &- \mathbf{E}^{(v)}||_{F}^{2} + \langle \mathbf{W}^{(v)}, \mathbf{Z}^{(v)} - \mathbf{G}^{(v)} \rangle + \frac{\rho}{2} ||\mathbf{Z}^{(v)} - \mathbf{G}^{(v)}||_{F}^{2}.
    \end{aligned}
\end{equation}
By setting the derivative of (\ref{z-subproblem}) to zero, the closed-form of $\mathbf{Z}^{(v)}$ can be obtained by
\begin{equation}\label{z-subproblem-solution}
    \begin{aligned}
    \mathbf{Z}^{(v)^{\ast}} = & (\mathbf{I} + \frac{\mu}{\rho} \mathbf{X}^{(v)^{T}}\mathbf{X}^{(v)})^{-1}\bigg((\mathbf{X}^{(v)^{T}}\mathbf{Y}_{v} + \mu\mathbf{X}^{(v)^{T}}\mathbf{X}^{(v)}\\
    &- \mu\mathbf{X}^{(v)^{T}}\mathbf{E}^{(v)} - \mathbf{W}^{(v)})/\rho +
    \mathbf{G}^{(v)}\bigg).
    \end{aligned}
\end{equation}

\textbf{$\mathbf{E}^{(v)}$-subproblem:}
\begin{equation}\label{E-subproblem}
    \begin{aligned}
    \mathbf{E}^{\ast} &= \argmin_{\mathbf{E}}\lambda ||\mathbf{E}||_{2,1} + \sum_{v=1}^{V} \bigg(\langle \mathbf{Y}_{v}, \mathbf{X}^{(v)} - \mathbf{X}^{(v)}\mathbf{Z}^{(v)} - \mathbf{E}^{(v)} \rangle \\
    & +\frac{\mu}{2} ||\mathbf{X}^{(v)} - \mathbf{X}^{(v)}\mathbf{Z}^{(v)} - \mathbf{E}^{(v)}||_{F}^{2} \bigg)\\
    &= \argmin_{\mathbf{E}} \frac{\lambda}{\mu} ||\mathbf{E}||_{2,1} + \frac{1}{2} ||\mathbf{E} - \mathbf{D}||_{F}^{2},
    \end{aligned}
\end{equation}
where $\mathbf{D}$ is constructed by vertically concatenating the matrices $\mathbf{X}^{(v)} - \mathbf{X}^{(v)}\mathbf{Z}^{(v)} + (1/\mu) \mathbf{Y}_{v}$ together along column. According to the Lemma 4.1 in \cite{LRR}, this subproblem has the following solution,
\begin{equation}\label{E-subproblem-solution}
    \mathbf{E}_{:,i}^{\ast} =
    \left\{
        \begin{aligned}
        &\frac{||\mathbf{D}_{:,i}||_2 - \frac{\lambda}{\mu}}{||\mathbf{D}_{:,i}||_2}\mathbf{D}_{:,i}, \quad \quad ||\mathbf{D}_{:,i}||_2 >  \frac{\lambda}{\mu}\\
        &\mathbf{0} \quad \text{otherwise.} \\
        \end{aligned}
    \right.
\end{equation}
where $\mathbf{D}_{:,i}$ represents the $i$-th column of the matrix $\mathbf{D}$.

\begin{algorithm}[]
\SetAlgoLined
\caption{t-SVD based Tensor Multi-Rank Minimization}
\label{TNN}
\KwIn{Observed tensor $\boldsymbol{\mathcal{F}}\in \mathbb{R}^{n_1 \times n_2 \times n_3}$, scalar $\tau > 0$}
\KwOut{tensor $\boldsymbol{\mathcal{G}}$}
\BlankLine
$\boldsymbol{\mathcal{F}}_{f} = \mathrm{fft}(\boldsymbol{\mathcal{F}},[~], 3)$, $\tau' = n_3 \tau$\;
\For{$j=1:n_3$}
{
    $[\boldsymbol{\mathcal{U}}_{f}^{(j)}, \boldsymbol{\mathcal{S}}_{f}^{(j)}, \boldsymbol{\mathcal{V}}_{f}^{(j)}] = \mathrm{SVD}(\boldsymbol{\mathcal{F}}_{f}^{(j)})$\;

    $\boldsymbol{\mathcal{J}}_{f}^{(j)} = \mathrm{diag}\{(1 - \frac{\tau'}{\boldsymbol{\mathcal{S}}_{f}^{(j)}(i,i)})_{+}\}, \quad i=1, \ldots, \min(n_1, n_2)$\;

    $\boldsymbol{\mathcal{S}}_{f,\tau'}^{(j)} = \boldsymbol{\mathcal{S}}_{f}^{(j)} \boldsymbol{\mathcal{J}}_{J}^{(j)}$\;

    $\boldsymbol{\mathcal{G}}_{f}^{(j)} = \boldsymbol{\mathcal{U}}_{f}^{(j)} \boldsymbol{\mathcal{S}}_{f,\tau'}^{(j)} \boldsymbol{\mathcal{V}}_{f}^{(j)^{\mathrm{T}}}$\;
}
$\boldsymbol{\mathcal{G}} = \mathrm{ifft}(\boldsymbol{\mathcal{G}}_{f},[~], 3)$\;
\textbf{Return} tensor $\boldsymbol{\mathcal{G}}$.
\end{algorithm}

\textbf{$\boldsymbol{\mathcal{G}}$-subproblem:} When $\mathbf{Z}^{(v)}, (v=1, 2, \ldots, V)$ are fixed, for updating the tensor $\boldsymbol{\mathcal{G}}$, we solve the following subproblem:
\begin{equation}\label{G-subproblem}
    \boldsymbol{\mathcal{G}}^{\ast} = \argmin_{\boldsymbol{\mathcal{G}}} ||\boldsymbol{\mathcal{G}}||_{\circledast} + \frac{\rho}{2} ||\boldsymbol{\mathcal{G}} - (\boldsymbol{\mathcal{Z}} + \frac{1}{\rho}\boldsymbol{\mathcal{W}})||_{F}^{2}.
\end{equation}
which is referred to as the {\it tensor multi-rank minimization} in this paper. The solution of the optimization problem (\ref{G-subproblem}) can be achieved through the following theorem\footnote{A similar discussion about the optimization of the TNN regularized low-rank tensor completion problem can be found in \cite{zhang14}.}:

\begin{theorem}
\label{theory:tsvc}
For $\tau > 0$ and $\boldsymbol{\mathcal{G}}, \boldsymbol{\mathcal{F}} \in \mathbb{R}^{n_{1} \times n_{2} \times n_{3}}$, the globally optimal solution to the following problem
\begin{equation}\label{fml:ntsvd}
\min_{\boldsymbol{\mathcal{G}}} ~ \tau||\boldsymbol{\mathcal{G}}||_{\circledast} + \frac{1}{2}||\boldsymbol{\mathcal{G}} - \boldsymbol{\mathcal{F}}||_{F}^{2}
\end{equation}
is given by the tensor tubal-shrinkage operator
\begin{equation}\label{fml:tsvc}
\boldsymbol{\mathcal{G}} = \mathcal{C}_{n_{3}\tau}(\boldsymbol{\mathcal{F}})=\boldsymbol{\mathcal{U}}*\mathcal{C}_{n_{3}\tau}(\boldsymbol{\mathcal{S}}) *\boldsymbol{\mathcal{V}}^{\mathrm{T}},
\end{equation}
where $\boldsymbol{\mathcal{F}} = \boldsymbol{\mathcal{U}}*\boldsymbol{\mathcal{S}}*\boldsymbol{\mathcal{V}}^{\mathrm{T}}$ and $\mathcal{C}_{n_{3}\tau}(\boldsymbol{\mathcal{S}}) = \boldsymbol{\mathcal{S}}*\boldsymbol{\mathcal{J}}$, herein, $\boldsymbol{\mathcal{J}}$ is an $n_{1} \times n_{2} \times n_{3}$ f-diagonal tensor whose diagonal element in the Fourier domain is $\boldsymbol{\mathcal{J}}_{f}(i,i,j) = (1 - \frac{n_{3}\tau}{{ \boldsymbol{\mathcal{S}}}_{f}^{(j)}(i,i)})_{+}$.
\end{theorem}

The proof can be found in Appendix. We summarize the t-SVD based tensor multi-rank minimization in Algorithm \ref{TNN}. Additionally, the Lagrange multipliers $\mathbf{Y}_{v}$ and $\boldsymbol{\mathcal{W}}$ need to be updated as follows:
\begin{align}
    \mathbf{Y}_{v}^{\ast} &= \mathbf{Y}_{v} + \mu(\mathbf{X}^{(v)} - \mathbf{X}^{(v)}\mathbf{Z}^{(v)} - \mathbf{E}^{(v)}), \label{update-multipliers-Y}\\
    \boldsymbol{\mathcal{W}}^{\ast} &= \boldsymbol{\mathcal{W}} + \rho(\boldsymbol{\mathcal{Z}} - \boldsymbol{\mathcal{G}}). \label{update-multipliers-W}
\end{align}
Finally, the optimization procedure of the proposed multi-view subspace clustering method is described in Algorithm \ref{proposed-MSC}.

\begin{algorithm}[]
\SetAlgoLined
\caption{MSC via Tensor Multi-Rank Minimization}
\label{proposed-MSC}
\KwIn{Multi-view feature matrices: $\mathbf{X}^{(1)}, \mathbf{X}^{(2)}, \ldots, \mathbf{X}^{(V)}$, $\lambda$, and cluster number $K$}
\KwOut{Clustering results $\mathcal{C}$}
\BlankLine
Initialized $\mathbf{Z}^{(v)} = \mathbf{0}$, $\mathbf{E}^{(v)} = \mathbf{0}$, $\mathbf{Y}_{v} = \mathbf{0}$, $i = 1,\ldots, V$; $\boldsymbol{\mathcal{G}} = \boldsymbol{\mathcal{W}} = \mathbf{0}$;\\
\nonl $\mu = 10^{-5}$, $\rho = 10^{-4}$, $\eta = 2$, $\mu_{\max} = \rho_{\max} = 10^{10}$, $\varepsilon = 10^{-7}$;\\
\While{not converge}
{
   \tcp*[h]{Solving $\mathbf{Z}$}\\
    \For{$v=1:V$}
    {
        Update $\mathbf{Z}^{(v)}$ by using (\ref{z-subproblem-solution});\\
    }
    \tcp*[h]{Solving $\mathbf{E}$}\\
    Update $\mathbf{E}$ by solving (\ref{E-subproblem})\;

    \For{$v=1:V$}
    {
        Update $\mathbf{Y}_{v}$ by using (\ref{update-multipliers-Y})\;
    }
    Obtain $\boldsymbol{\mathcal{Z}} = \Phi(\mathbf{Z}^{(1)}, \mathbf{Z}^{(2)}, \ldots, \mathbf{Z}^{(V)})$\;
    \tcp*[h]{Solving $\boldsymbol{\mathcal{G}}$}\\
    Update $\boldsymbol{\mathcal{G}}$ via subproblem (\ref{G-subproblem}) by using Algorithm \ref{TNN}\;
    Update $\boldsymbol{\mathcal{W}}$ by using (\ref{update-multipliers-W})\;
    Update parameters $\mu$ and $\rho$: $\mu = \min(\eta \mu, \mu_{\max})$, $\rho = \min(\eta \rho, \rho_{\max})$\;

    $(\mathbf{G}^{(1)}, \ldots, \mathbf{G}^{(V)}) = \Phi^{-1}(\boldsymbol{\mathcal{G}})$\;
    Check the convergence conditions:\\
    \nonl \qquad $||\mathbf{X}^{(v)} - \mathbf{X}^{(v)}\mathbf{Z}^{(v)} - \mathbf{E}^{(v)}||_{\infty} < \varepsilon$ and \\
    \nonl \qquad $||\mathbf{Z}^{(v)} - \mathbf{G}^{(v)}||_{\infty} < \varepsilon$\;
}

Obtain the affinity matrix by\\
\nonl \qquad $\mathbf{A} = \frac{1}{V}\sum_{v=1}^{V} |\mathbf{Z}^{(v)}| + |\mathbf{Z}^{(v)^{\mathrm{T}}}|$\;
Apply the spectral clustering method with the affinity matrix $\mathbf{A}$\;
\textbf{Return} Clustering result $\mathcal{C}$.
\end{algorithm}

\subsection{Convergence Properties and Computational Complexity}\label{converge-complexity}
The convergence properties of the inexact ALM have been well established when the number of blocks is at most two \cite{alm-lin}. Despite of its success in practice, its convergence properties for minimizing the objective function with $N~(N\geq 3)$ blocks variables linked by linear constraints, have remained unclear. Since there are several blocks (including $\{\mathbf{Z}^{(v)}\}_{v=1}^{V}, \mathbf{E}$, and $\boldsymbol{\mathcal{G}}$) in Algorithm \ref{proposed-MSC}, and the objective function of (\ref{original-problem}) is not smooth, it would be not easy to prove the convergence in theory. Fortunately, as suggested in \cite{LRR}, two conditions are sufficient (but may not necessary) for Algorithm \ref{proposed-MSC} to converge: (1) each feature matrix $\mathbf{X}^{(v)}$ is of {\it \textbf{full column rank}}; (2) the optimality gap produced in each iteration step is {\it \textbf{monotonically decreasing}}. The first condition can be met by factorizing $\mathbf{Z}^{(v)}$ into $\mathbf{P}^{(v)}\mathbf{\hat{Z}}^{(v)}$, where $\mathbf{P}^{(v)}$ can be computed in advance by orthogonalizing the columns of $\mathbf{X}^{(v)^{\mathrm{T}}}$. Moreover, due to the convexity of the Lagrangian function (\ref{t-SVD-MSC}), the monotonically decreasing condition can be guaranteed to some extent according to \cite{optimal-gap}. Therefore, the proposed MSC algorithm ensures good convergence properties. Furthermore, the proposed method performs well and indeed converges fast in reality, which will be illustrated in Section \ref{convergence-and-complexity}.

Since inverse matrix can be calculated in advance in Eq. (\ref{z-subproblem-solution}) for solving $\mathbf{Z}^{(v)}$, the computational bottleneck of the proposed Algorithm \ref{proposed-MSC} only lies in solving the subproblems for $\mathbf{E}$ and $\boldsymbol{\mathcal{G}}$. As for the $\mathbf{E}$ subproblem, it takes $\mathcal{O}(VN^2)$ in each iteration. As for the $\boldsymbol{\mathcal{G}}$ subproblem, calculating the $3D$ FFT and $3D$ inverse FFT of an $N \times V \times N$ tensor and $N$ SVDs of $N \times V$ matrices in the Fourier domain, actually dominate the main computation. Since in multi-view setting we have $N \gg V$ and $\log(N) > V$, the computation at each iteration will take $\mathcal{O}(2N^2 V \log(N) + N^{2}V^{2}) \approx \mathcal{O}(2N^2 V \log(N))$. By considering the cost of spectral clustering (usually $\mathcal{O}(N^3)$) and the number of iterations needed to converge, the complexity of Algorithm \ref{proposed-MSC} is:
\begin{equation}\label{BigO}
    \mathcal{O}(N^3) + \mathcal{O}(K(2N^2 V \log(N))),
\end{equation}
where $K$ denotes the number of iterations. The iteration number $K$ depends on the choice of $\eta$: larger $\eta$ leads to a smaller $K$, and vice versa. In our experiments, we fix the $\eta$ to $2$, such that the iteration number $K$ commonly locates within the range of $30\thicksim50$.

{\color{red}
\subsection{Discussion}\label{contribution-of-multi-view}

Furthermore, we can analyze the contribution of each view to final clustering from the perspective of feature's characteristic, {\it i.e.,} discriminative power, both theoretically and experimentally (Section \ref{view-contribution}). Recent studies on sparse subspace clustering \cite{SSC} have proved that a sample can be represented by its corresponding dictionary if the signals satisfy certain incoherence condition. In other words, low rank representation of a data point ideally corresponds to a combination of all the point from its own subspace, leading to a block-diagonal connectivity in affinity matrix. As it is proved in \cite{LRR}, for a certain feature, the closer the affinity matrix $\mathbf{A}^{(v)} = \frac{1}{2}(|\mathbf{Z}^{(v)}| + |\mathbf{Z}^{(v)^{\mathbf{T}}}|)$ is to block-diagonal structure, the better the clustering result is. It is worth noting that the block-diagonal property does not require the data samples to have been grouped together according to their subspace memberships, because the solution produced by low rank representation is globally optimal and does not depend on the arrangements of the data samples \cite{LRR}.

However, in real application, different features have different capabilities of discriminative power. The feature with less discriminative power incurs more non-zero responses in the atoms belonging to different subspaces ({\it e.g.,} $\mathbf{Z}_{ij} \neq 0$, where samples $i$ and $j$ belong to different subspaces), while strongly discriminant feature will force the linear representation coefficients on different subspaces tend to zero ({\it e.g.,} $\mathbf{Z}_{ij} = 0$, where samples $i$ and $j$ belong to different subspaces). Discriminant feature will make the edges between points in different subspaces weak, such that spectral clustering can find the correct segmentation. Therefore, theoretically, discriminant feature will provide greater contribution to the final clustering results.}

\section{Experimental Results and Analysis}\label{sec:experiment}
In this section, we perform experiments on several challenging image clustering datasets to present a comprehensive evaluation of the proposed method. We test our method on three applications: face clustering, scene clustering, and generic object clustering. The statistics of all the datasets are summarized in Table \ref{dataset}.

\textbf{Competitors:} We compare the proposed method with seven representative clustering algorithms: the standard spectral clustering algorithm with the most informative view ($\text{SPC}_{\text{best}}$), LRR algorithm with the most informative view ($\text{LRR}_{\text{best}}$), robust multi-view spectral clustering via low-rank and sparse decomposition (RMSC) \cite{graph-based4}, diversity-induced multi-view subspace clustering (DiMSC) \cite{DiMSC}, low-rank tensor constrained multi-view subspace clustering (LTMSC) \cite{LRTM}, the proposed method with unrotated coefficient tensor (Ut-SVD-MSC), learning and transferring deep ConvNet (convolutional neural network) representations with group-sparse factorization (GSNMF-CNN) \cite{GSNMF-CNN}. The first two methods are the single view baselines. The RMSC, DiMSC, and LTMSC represent the state-of-the-art methods in multi-view clustering. Comparing with the method Ut-SVD-MSC is used to illustrate the advantage of the tensor rotation. The last approach, {\it i.e.}, GSNMF-CNN, does not belong to multi-view method but with the claim that it achieves state-of-the-art image clustering performance by using deep ConvNet.

\textbf{Evaluation Methodology:} Different experimental {\it \textbf{strategies}} are adopted for different applications. As for face clustering, we use relatively simple image features ({\it e.g.}, intensity, LBP, Gabor) to test the performance of different multi-view clustering methods. As for scene clustering, some sophisticated features (such as PHOW \cite{phow}, CENTRIST \cite{centrist}, etc.) are considered as views to handle scene clustering. Besides traditional handcrafted features, we utilize the CNN feature trained on large-scale annotated dataset (ImageNet) to handle two challenging datasets, {\it i.e.,} MITIndoor-67 and Caltech-101 for scene clustering and generic object clustering, respectively. Since CNN feature is adopted in our experiments, it is necessary to compare with a state-of-the-art CNN based image clustering method, termed the GSNMF-CNN \cite{GSNMF-CNN}, which shows the transferability of the deep ConvNet trained on ImageNet to be used for enhancing image clustering. Due to the different scales, features, and challenges of different datasets, we leave the description of the detailed experimental setup to the corresponding section of each application.

\begin{table}[!ht]
\footnotesize
\renewcommand{\arraystretch}{1.3}
\begin{centering}
\begin{threeparttable}[]
\caption{\color{red}{Statistics of different test datasets}}\label{dataset}
\tabcolsep=4pt
\begin{minipage}{12cm}
\begin{tabular}{@{}l|ccc}
\toprule[2pt] 
Dataset         &Images &Objective &Clusters  \\ \hline
Yale               &165  &Face &15 \\
Extended YaleB     &640  &Face &10 \\
ORL                &400  &Face &{\color{red}40} \\
Notting-Hill        &4660 &Face &5 \\ \hline
Scene-15           &4485 &Scene &15 \\
MITIndoor-67       &5360 &Scene &67 \\ \hline
COIL-20            &1440 &Generic Object &20 \\
Caltech-101        &8677 &Generic Object &101 \\\hline \bottomrule[1pt]
\end{tabular}
\end{minipage}
\end{threeparttable}
\end{centering}
\end{table}

\textbf{Evaluation Measures.} The evaluation of clustering results is a challenging problem. {\color{red}Two types of criteria are generally used for measuring cluster quality \cite{evaluation-metric1}: external and internal criteria. External criteria measures the agreement between the clustering result and an external input (usually the groundtruth of the dataset). Internal criteria, on the other hand, measures quality based on characteristic of the data and the partitioning result ({\it e.g.,} between-cluster and within-cluster scatter). However, good scores on an internal criterion do not necessarily translate into good effectiveness in an application \cite{evaluation-metric1}. Moreover, under subspace clustering setting, since data in a subspace are often distributed arbitrarily and not around a centroid \cite{SSC}, standard internal criteria measurement that take advantage of the spatial proximity of the data can not be applicable. So, external criteria has been widely used for evaluating clustering performance \cite{graph-based4,LRTM,DiMSC}.

In our experiments, six popular external metrics are used to evaluate the performances \cite{evaluation-metric1,evaluation-metric2}: Normalized Mutual Information (NMI), Accuracy (ACC), Adjusted Rank index (AR), F-score, Precision and Recall.

NMI can be information-theoretically interpreted. Suppose that $C$ and $C'$ represent the predicted partition and the groundtruth partition respectively, the NMI metric is calculated as:
\begin{equation}\label{NMI-clustering}
    NMI(C, C') = \frac{\sum_{i=1}^{K} \sum_{j=1}^{S} |C_i\cap C'_j| log\frac{N|C_i\cap C'_j|}{|C_i||C'_j|}}{\sqrt{(\sum_{i=1}^{K}|C_i|log\frac{C_i}{N}) (\sum_{j=1}^{S}|C'_j|log\frac{C'_j}{N})}},
\end{equation}

For the definition of accuracy, suppose the clustering algorithm is tested on $N$ samples. For a sample $\mathbf{x}_i$, the cluster label is denoted as $r_i$, and groundtruth is $t_i$. The accuracy is defined as follows:
\begin{equation}\label{ACC}
    ACC = \frac{\sum_{i=1}^{N} \delta(t_i, map(r_i))}{N},
\end{equation}
where
\begin{equation}\label{delta}
    \begin{aligned}
        \delta(a, b) =
        \left\{
            \begin{aligned}
                &1 \quad \text{if } a = b\\
                &0 \quad \text{otherwise},\\
            \end{aligned}
        \right.
    \end{aligned}
\end{equation}
Function $map(x)$ denotes the best permutation mapping function gained by Hungarian algorithm \cite{hungarian}, which maps cluster to the corresponding groundtruth label. So the more labels of samples are predicted correctly, the greater the accuracy is.

As for F-score, Precision, Recall, and AR, these four metrics view the clustering as a series of decisions, one for each the $N(N-1)/2$ pairs of samples on the dataset. The goal is to assign two samples to the same cluster if and only if they are similar. For more details about their definitions, please refer to \cite{evaluation-metric1}.}

For each of the metrics, the higher it is, the better the performance is. Those metrics favor different properties in the clustering such that a comprehensive evaluation can be achieved. {\color{red}Note that in all dataset, the reported final results on those metrics are measured by the average and standard derivation of $20$ runs.} We highlight the best values in bold font in each table.

Only one parameter $\lambda$ needs to be tuned, and we found its empirical value is within the range $[0.1, 2]$. More details about the parameter will be discussed in Section \ref{model-analysis}. The parameters in other competitors are set within ranges suggested by original papers, and we tune those parameters so as to show the best results. All experiments are implemented in Matlab on a workstation with $4.0$GHz CPU, $32$GB RAM, and TITANX GPU (12GB caches). To promote the culture of reproducible research, source codes and experimental results accompanying this paper can be achieved at https://www.researchgate.net/profile/Yuan\_Xie4.

\subsection{Experiments on Face Clustering}\label{face-clustering}

The {\it Yale}\footnote{https://cvc.yale.edu/projects/yalefaces/yalefaces.html} face dataset contains $165$ grayscale images of $15$ individuals. There are $11$ images per subject, one per different facial expression or configuration.

The {\it Extended YaleB}\footnote{https://cvc.yale.edu/projects/yalefacesB/yalefacesB.html} dataset includes $38$ individuals and around $64$ near frontal images under different illuminations for each individual. Similarly to the works \cite{LRR,LRTM}, we use a part of images which contains the first $10$ individuals, including $640$ frontal face images.

The {\it ORL}\footnote{http://www.uk.research.att.com/facedatabase.html} dataset consists of $40$ distinct subjects, each of which contains $10$ different images captured under different times, lighting, facial expressions, and facial details.

For all those datasets, similar to \cite{LRTM}, three types of features are extracted: intensity, LBP \cite{LBP} and Gabor \cite{Gabor}. The standard LBP features are extracted with the sampling size of $8$ pixels, and the blocking number of $7 \times 8$. The Gabor feature is extracted with one scale $\lambda = 4$ at four orientations $\theta = \{0^{o}, 45^{o}, 90^{o}, 135^{o}\}$. Therefore, the dimensionalities of LBP and Gabor are $3304$ and $6750$, respectively.

\begin{table*}[!ht]
\footnotesize
\renewcommand{\arraystretch}{1.3}
\begin{centering}
\begin{threeparttable}[]
\caption{\color{red}{Clustering results (mean $\pm$ standard deviation) on {\it Yale}. We set $\lambda = 1.1$ in proposed t-SVD-MSC.}}\label{result-yale}
\tabcolsep=4pt
\begin{minipage}{12cm}
\begin{tabular}{@{}l|cccccc}
\toprule[2pt] 
Method         &NMI &ACC &AR &F-score &Precision &Recall \\ \hline
$\text{SPC}_{\text{best}}$    &0.654 $\pm$ 0.009 &0.618 $\pm$ 0.030 &0.440 $\pm$ 0.011 &0.475 $\pm$ 0.011 &0.457 $\pm$ 0.011 &0.500 $\pm$ 0.010 \\
$\text{LRR}_{\text{best}}$    &0.709 $\pm$ 0.011 &0.697 $\pm$ 0.001 &0.512 $\pm$ 0.005 &0.547 $\pm$ 0.007 &0.529 $\pm$ 0.005 &0.567 $\pm$ 0.004\\ \hline
RMSC                          &0.684 $\pm$ 0.033 &0.642 $\pm$ 0.036 &0.485 $\pm$ 0.042 &0.517 $\pm$ 0.043 &0.500 $\pm$ 0.043 &0.535 $\pm$ 0.044\\
DiMSC                         &0.727 $\pm$ 0.010 &0.709 $\pm$ 0.003 &0.535 $\pm$ 0.003 &0.564 $\pm$ 0.010 &0.543 $\pm$ 0.012 &0.586 $\pm$ 0.009\\
LTMSC                         &0.765 $\pm$ 0.008 &0.741 $\pm$ 0.002 &0.570 $\pm$ 0.004 &0.598 $\pm$ 0.006 &0.569 $\pm$ 0.004 &0.629 $\pm$ 0.005\\
Ut-SVD-MSC                    &0.756 $\pm$ 0.012 &0.733 $\pm$ 0.005 &0.584 $\pm$ 0.003 &0.610 $\pm$ 0.006 &0.591 $\pm$ 0.005 &0.630 $\pm$ 0.006\\
t-SVD-MSC                     &\textbf{0.953 $\pm$ 0.008} &\textbf{0.963 $\pm$ 0.006} &\textbf{0.910 $\pm$ 0.010} &\textbf{0.915 $\pm$ 0.007} &\textbf{0.904 $\pm$ 0.005} &\textbf{0.927 $\pm$ 0.007}\\\hline \bottomrule[1pt]
\end{tabular}
\end{minipage}
\end{threeparttable}
\end{centering}
\end{table*}

\begin{table*}[!ht]
\footnotesize
\renewcommand{\arraystretch}{1.3}
\begin{centering}
\begin{threeparttable}[]
\caption{{\color{red}Clustering results (mean $\pm$ standard deviation) on {\it Extended YaleB}. We set $\lambda = 1.3$ in proposed t-SVD-MSC.}}\label{result-yaleB}
\tabcolsep=4pt
\begin{minipage}{12cm}
\begin{tabular}{@{}l|cccccc}
\toprule[2pt] 
Method         &NMI &ACC &AR &F-score &Precision &Recall \\ \hline
$\text{SPC}_{\text{best}}$    &0.360 $\pm$ 0.014  &0.366 $\pm$ 0.059 &0.225 $\pm$ 0.018 &0.308 $\pm$ 0.011 &0.296 $\pm$ 0.010 &0.310 $\pm$ 0.012 \\
$\text{LRR}_{\text{best}}$    &0.627 $\pm$ 0.040  &0.615 $\pm$ 0.013 &0.451 $\pm$ 0.002 &0.508 $\pm$ 0.004 &0.481 $\pm$ 0.002 &0.539 $\pm$ 0.001 \\ \hline
RMSC                          &0.157 $\pm$ 0.019  &0.210 $\pm$ 0.013 &0.060 $\pm$ 0.014 &0.155 $\pm$ 0.012 &0.151 $\pm$ 0.012 &0.159 $\pm$ 0.013 \\
DiMSC                         &0.636 $\pm$ 0.002  &0.615 $\pm$ 0.003 &0.453 $\pm$ 0.005 &0.504 $\pm$ 0.006 &0.481 $\pm$ 0.004 &0.534 $\pm$ 0.004 \\
LTMSC                         &0.637 $\pm$ 0.003  &0.626 $\pm$ 0.010 &0.459 $\pm$ 0.030 &0.521 $\pm$ 0.006 &0.485 $\pm$ 0.001 &0.539 $\pm$ 0.002 \\
Ut-SVD-MSC                    &0.479 $\pm$ 0.007  &0.470 $\pm$ 0.011 &0.274 $\pm$ 0.005 &0.350 $\pm$ 0.007 &0.327 $\pm$ 0.004 &0.375 $\pm$ 0.005\\
t-SVD-MSC                     &\textbf{0.667 $\pm$ 0.004} &\textbf{0.652 $\pm$ 0.000} &\textbf{0.500 $\pm$ 0.003} &\textbf{0.550 $\pm$ 0.002} &\textbf{0.514 $\pm$ 0.004} &\textbf{0.590 $\pm$ 0.004}\\\hline \bottomrule[1pt]
\end{tabular}
\end{minipage}
\end{threeparttable}
\end{centering}
\end{table*}

\begin{table*}[!ht]
\footnotesize
\renewcommand{\arraystretch}{1.3}
\begin{centering}
\begin{threeparttable}[]
\caption{{\color{red}Clustering results (mean $\pm$ standard deviation) on {\it ORL}. We set $\lambda = 0.2$ in proposed t-SVD-MSC.}}\label{result-ORL}
\tabcolsep=4pt
\begin{minipage}{12cm}
\begin{tabular}{@{}l|cccccc}
\toprule[2pt] 
Method         &NMI &ACC &AR &F-score &Precision &Recall \\ \hline
$\text{SPC}_{\text{best}}$    &0.884 $\pm$ 0.002 &0.725 $\pm$ 0.025 &0.655 $\pm$ 0.005 &0.664 $\pm$ 0.005 &0.610 $\pm$ 0.006 &0.728 $\pm$ 0.005 \\
$\text{LRR}_{\text{best}}$    &0.895 $\pm$ 0.006 &0.773 $\pm$ 0.003 &0.724 $\pm$ 0.020 &0.731 $\pm$ 0.004 &0.701 $\pm$ 0.001 &0.754 $\pm$ 0.002 \\ \hline
RMSC                          &0.872 $\pm$ 0.012 &0.723 $\pm$ 0.007 &0.645 $\pm$ 0.003 &0.654 $\pm$ 0.007 &0.607 $\pm$ 0.009 &0.709 $\pm$ 0.004 \\
DiMSC                         &0.940 $\pm$ 0.003 &0.838 $\pm$ 0.001 &0.802 $\pm$ 0.000 &0.807 $\pm$ 0.003 &0.764 $\pm$ 0.012 &0.856 $\pm$ 0.004 \\
LTMSC                         &0.930 $\pm$ 0.003 &0.795 $\pm$ 0.007 &0.750 $\pm$ 0.003 &0.768 $\pm$ 0.004 &0.766 $\pm$ 0.009 &0.837 $\pm$ 0.005 \\
Ut-SVD-MSC                    &0.874 $\pm$ 0.002 &0.765 $\pm$ 0.001 &0.666 $\pm$ 0.004 &0.675 $\pm$ 0.005 &0.643 $\pm$ 0.003 &0.708 $\pm$ 0.002 \\
t-SVD-MSC                     &\textbf{0.993 $\pm$ 0.002} &\textbf{0.970 $\pm$ 0.003} &\textbf{0.967 $\pm$ 0.002} &\textbf{0.968 $\pm$ 0.003} &\textbf{0.946 $\pm$ 0.004} &\textbf{0.991 $\pm$ 0.003}\\\hline \bottomrule[1pt]
\end{tabular}
\end{minipage}
\end{threeparttable}
\end{centering}
\end{table*}

\begin{table*}[!ht]
\footnotesize
\renewcommand{\arraystretch}{1.3}
\begin{centering}
\begin{threeparttable}[]
\caption{{\color{red}Clustering results (mean $\pm$ standard deviation) on {\it Notting-Hill}. We set $\lambda = 0.1$ in proposed t-SVD-MSC.}}\label{result-Notting-Hill}
\tabcolsep=4pt
\begin{minipage}{12cm}
\begin{tabular}{@{}l|cccccc}
\toprule[2pt] 
Method         &NMI &ACC &AR &F-score &Precision &Recall \\ \hline
$\text{SPC}_{\text{best}}$    &0.723 $\pm$ 0.008 &0.816 $\pm$ 0.000 &0.712 $\pm$ 0.020 &0.775 $\pm$ 0.015 &0.780 $\pm$ 0.018 &0.776 $\pm$ 0.013 \\
$\text{LRR}_{\text{best}}$    &0.579 $\pm$ 0.003 &0.794 $\pm$ 0.033 &0.558 $\pm$ 0.007 &0.653 $\pm$ 0.007 &0.672 $\pm$ 0.007 &0.636 $\pm$ 0.008 \\ \hline
RMSC                          &0.585 $\pm$ 0.002 &0.807 $\pm$ 0.013 &0.496 $\pm$ 0.004 &0.603 $\pm$ 0.005 &0.621 $\pm$ 0.002 &0.586 $\pm$ 0.011 \\
DiMSC                         &0.799 $\pm$ 0.001 &0.837 $\pm$ 0.021 &0.787 $\pm$ 0.001 &0.834 $\pm$ 0.001 &0.822 $\pm$ 0.005 &0.827 $\pm$ 0.009 \\
LTMSC                         &0.779 $\pm$ 0.003 &0.868 $\pm$ 0.000 &0.777 $\pm$ 0.002 &0.825 $\pm$ 0.002 &0.830 $\pm$ 0.002 &0.814 $\pm$ 0.004 \\
Ut-SVD-MSC                    &0.837 $\pm$ 0.005 &0.933 $\pm$ 0.015 &0.847 $\pm$ 0.001 &0.880 $\pm$ 0.005 &0.900 $\pm$ 0.004 &0.861 $\pm$ 0.009 \\
t-SVD-MSC                     &\textbf{0.900 $\pm$ 0.005} &\textbf{0.957 $\pm$ 0.010} &\textbf{0.900 $\pm$ 0.003} &\textbf{0.922 $\pm$ 0.003} &\textbf{0.937 $\pm$ 0.006} &\textbf{0.907 $\pm$ 0.005}\\\hline \bottomrule[1pt]
\end{tabular}
\end{minipage}
\end{threeparttable}
\end{centering}
\end{table*}

\begin{figure}[]
\renewcommand{\figurename}{Figure}
  \centering
    \includegraphics[width=0.5\textwidth]{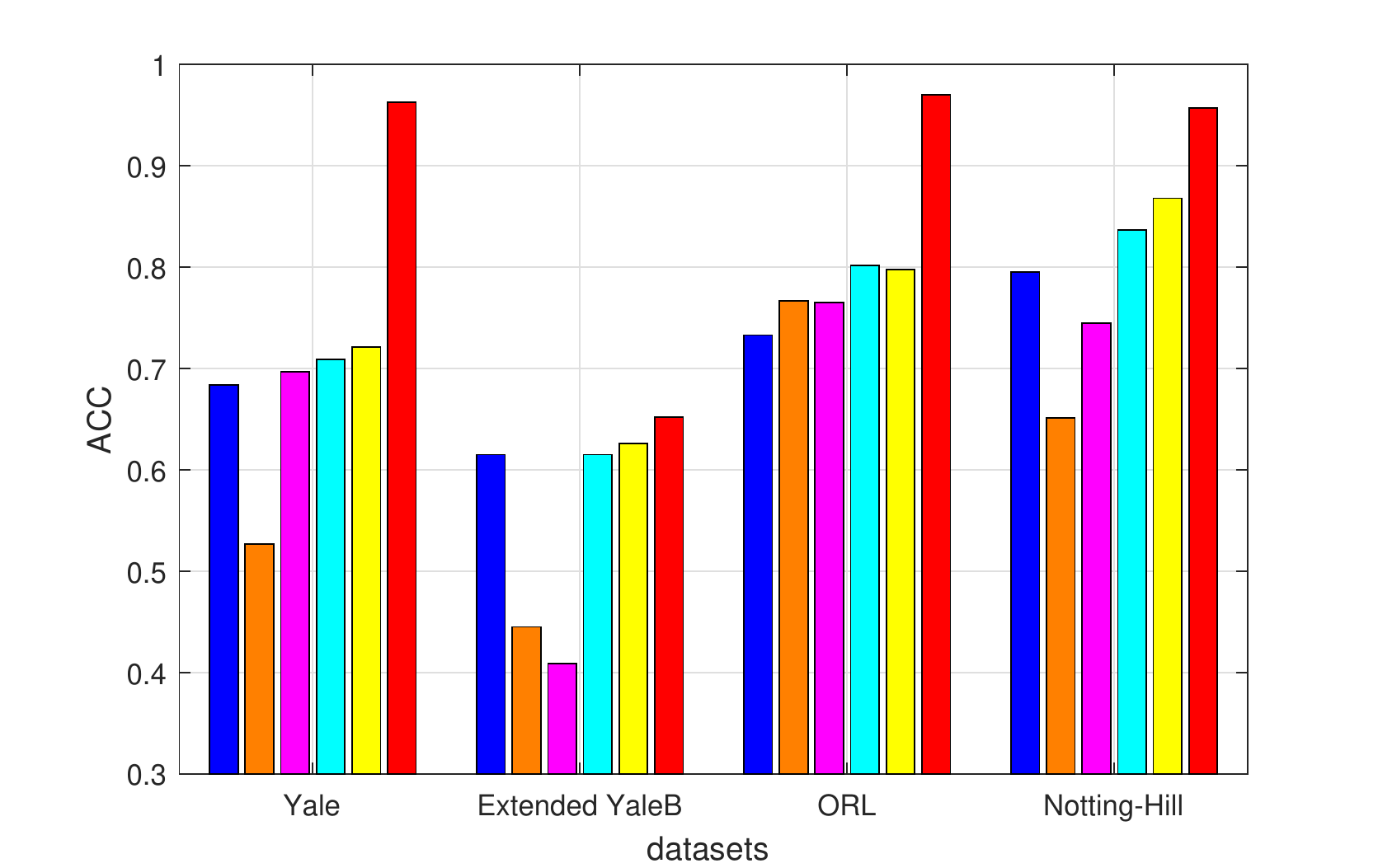}\\
  \vspace{-0.02in}
    \includegraphics[width=0.5\textwidth]{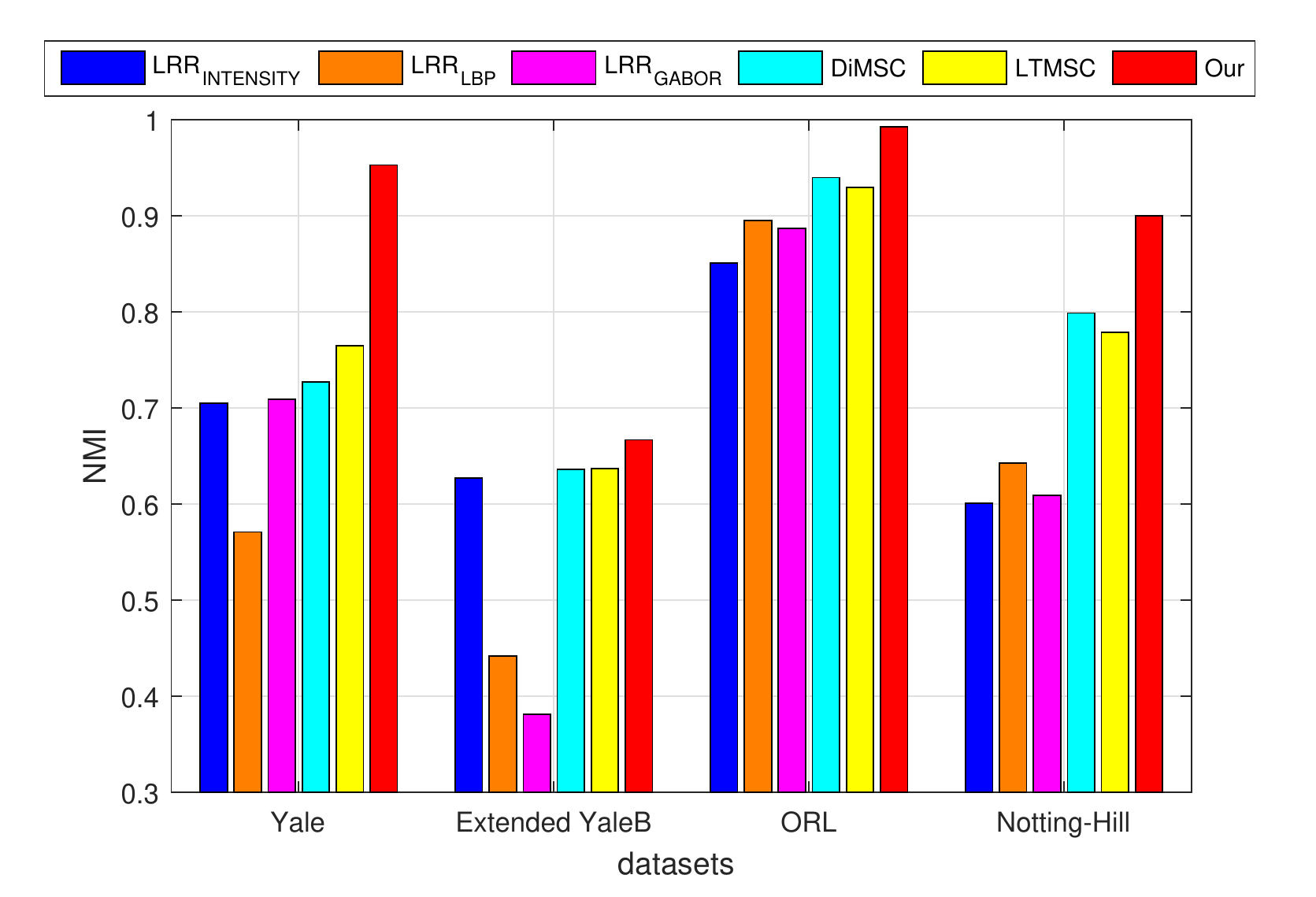}
  \caption{Comparison among LRR with all the view features, DiMSC, LTMSC and the proposed t-SVD-MSC in terms of accuracy and NMI on face clustering datasets.}
  \label{fig:LRR-All-View-compare} 
\end{figure}

As illustrated by the Table \ref{result-yale}, LTMSC performs the second best in terms of all metrics on Yale dataset, while the proposed approach presents a clear advance over it, {\it e.g.,} $0.953$ vs. $0.765$ in NMI, and $0.963$ vs. $0.741$ in ACC. Table \ref{result-ORL} gives the clustering results on the ORL dataset. It can be seen that quite a lot of methods achieve promising performance. However, our approach obtains a nearly perfect result in terms of all six metrics, {\it e.g.,} NMI $0.993$, ACC $0.970$, and Recall $0.991$, which means that our method still outperforms all the alternative approaches significantly. As shown in Table \ref{result-yaleB}, the improvement of t-SVD-MSC over other representative approaches (such as DiMSC and LTMSC) on Extended YaleB are not so much noticeable as on the above two datasets. We observe that, due to large variation of illumination, the LBP and Gabor features present significant lower capabilities of representation than intensity feature (see the second group bars in Fig. \ref{fig:LRR-All-View-compare}). {\color{red}Therefore, the basic assumption of the multi-view clustering might be violated so as to suffer the degradation of performance}. This observation coincides with the corresponding conclusions obtained in \cite{LRTM,DiMSC}.

The dataset {\it Notting-Hill} \cite{notting-hill} is constructed from the movie ``Notting-Hill'', where the faces of $5$ main casts are collected, including $4660$ faces in $76$ tracks. The original dataset consists of the facial images of the size of $120\times 150$, and we downsample each facial image to $40\times 50$. The features utilized in this dataset are the same with the features used in other face clustering datasets. Table \ref{result-Notting-Hill} shows the clustering result, where the proposed method also outperforms all other competitors in all metrics with clear large margins. Our result might even be comparable with the state-of-the-art result achieved by \cite{CMV-VFC} (with NMI 0.920 and ACC 0.934), where additional video-specific constraints are employed, {\it i.e.,} faces from the same track are likely to be from the same person, while faces do not belong to the same person if they appear together in the same video frame. Note that, the proposed method conduct video face clustering without any video-specific priori.

\begin{table*}[!ht]
\footnotesize
\renewcommand{\arraystretch}{1.3}
\begin{centering}
\begin{threeparttable}[]
\caption{{\color{red}Clustering results (mean $\pm$ standard deviation) on {\it Scene-15}. We set $\lambda = 1.5$ in proposed t-SVD-MSC.}}\label{result-Scene15}
\tabcolsep=4pt
\begin{minipage}{12cm}
\begin{tabular}{@{}l|cccccc}
\toprule[2pt] 
Method         &NMI &ACC &AR &F-score &Precision &Recall \\ \hline
$\text{SPC}_{\text{best}}$    &0.421 $\pm$ 0.010 &0.437 $\pm$ 0.015 &0.270 $\pm$ 0.010 &0.321 $\pm$ 0.022 &0.314 $\pm$ 0.016 &0.329 $\pm$ 0.020\\
$\text{LRR}_{\text{best}}$    &0.426 $\pm$ 0.018 &0.445 $\pm$ 0.013 &0.272 $\pm$ 0.015 &0.324 $\pm$ 0.010 &0.316 $\pm$ 0.015 &0.333 $\pm$ 0.015\\ \hline
RMSC                          &0.564 $\pm$ 0.023 &0.507 $\pm$ 0.017 &0.394 $\pm$ 0.025 &0.437 $\pm$ 0.019 &0.425 $\pm$ 0.021 &0.450 $\pm$ 0.024\\
DiMSC                         &0.269 $\pm$ 0.009 &0.300 $\pm$ 0.010 &0.117 $\pm$ 0.012 &0.181 $\pm$ 0.012 &0.173 $\pm$ 0.016 &0.190 $\pm$ 0.010\\
LTMSC                         &0.571 $\pm$ 0.011 &0.574 $\pm$ 0.009 &0.424 $\pm$ 0.010 &0.465 $\pm$ 0.007 &0.452 $\pm$ 0.003 &0.479 $\pm$ 0.008\\
Ut-SVD-MSC                    &0.555 $\pm$ 0.007 &0.510 $\pm$ 0.005 &0.375 $\pm$ 0.003 &0.422 $\pm$ 0.004 &0.389 $\pm$ 0.010 &0.460 $\pm$ 0.008\\
t-SVD-MSC                     &\textbf{0.858 $\pm$ 0.007} &\textbf{0.812 $\pm$ 0.007} &\textbf{0.771 $\pm$ 0.003} &\textbf{0.788 $\pm$ 0.001} &\textbf{0.743 $\pm$ 0.006} &\textbf{0.839 $\pm$ 0.003}\\\hline \bottomrule[1pt]
\end{tabular}
\end{minipage}
\end{threeparttable}
\end{centering}
\end{table*}

\begin{table*}[!ht]
\footnotesize
\renewcommand{\arraystretch}{1.3}
\begin{centering}
\begin{threeparttable}[]
\caption{{\color{red}Clustering results (mean $\pm$ standard deviation) on MITIndoor-67. We set $\lambda = 0.2$ in proposed t-SVD-MSC.}}\label{result-MITIndoor67}
\tabcolsep=4pt
\begin{minipage}{12cm}
\begin{tabular}{@{}l|cccccc}
\toprule[2pt] 
Method         &NMI &ACC &AR &F-score &Precision &Recall \\ \hline
$\text{SPC}_{\text{best}}^{\text{CNN}}$    &0.559 $\pm$ 0.009 &0.443 $\pm$ 0.011 &0.304 $\pm$ 0.011 &0.315 $\pm$ 0.013 &0.294 $\pm$ 0.010 &0.340 $\pm$ 0.014 \\
$\text{LRR}_{\text{best}}^{\text{CNN}}$    &0.226 $\pm$ 0.006 &0.120 $\pm$ 0.004 &0.031 $\pm$ 0.007 &0.045 $\pm$ 0.004 &0.044 $\pm$ 0.006 &0.047 $\pm$ 0.004\\ \hline
RMSC                                       &0.342 $\pm$ 0.004 &0.232 $\pm$ 0.009 &0.110 $\pm$ 0.003 &0.123 $\pm$ 0.002 &0.121 $\pm$ 0.003 &0.125 $\pm$ 0.003\\
DiMSC                                      &0.383 $\pm$ 0.003 &0.246 $\pm$ 0.000 &0.128 $\pm$ 0.005 &0.141 $\pm$ 0.004 &0.138 $\pm$ 0.001 &0.144 $\pm$ 0.002\\
LTMSC                                      &0.546 $\pm$ 0.004 &0.431 $\pm$ 0.002 &0.280 $\pm$ 0.008 &0.290 $\pm$ 0.002 &0.279 $\pm$ 0.006 &0.306 $\pm$ 0.005\\
GSNMF-CNN                                  &0.673 $\pm$ 0.003 &0.517 $\pm$ 0.003 &0.264 $\pm$ 0.005 &0.372 $\pm$ 0.002 &0.367 $\pm$ 0.004 &0.381 $\pm$ 0.001\\
Ut-SVD-MSC                                 &0.518 $\pm$ 0.010 &0.386 $\pm$ 0.007 &0.245 $\pm$ 0.013 &0.256 $\pm$ 0.007 &0.249 $\pm$ 0.006 &0.263 $\pm$ 0.006\\
t-SVD-MSC                                  &\textbf{0.750 $\pm$ 0.007} &\textbf{0.684 $\pm$ 0.005} &\textbf{0.555 $\pm$ 0.005} &\textbf{0.562 $\pm$ 0.008} &\textbf{0.543 $\pm$ 0.005} &\textbf{0.582 $\pm$ 0.004}\\\hline \bottomrule[1pt]
\end{tabular}
\end{minipage}
\end{threeparttable}
\end{centering}
\end{table*}

\subsection{Experiments on Scene Clustering}\label{scene-clustering}
{\it Scene-15}\footnote{http://www-cvr.ai.uiuc.edu/ponce\_grp/data/} dataset was gradually built by the works \cite{scene-15a,scene-15b,scene-15c} with 15 categories, including office, kitchen, living room, bedroom, etc. Images are about $250\times 300$ resolution, with $210$ to $410$ images per category. This dataset contains a wide range of outdoor and indoor scene environments. We extracted three kinds of handcrafted image features on this dataset: 1) Pyramid histograms of visual words (PHOW)\footnote{This feature was extracted by using vlfeat toolbox \cite{vlfeat}} feature \cite{phow} which was extracted with $8$ pixels' dense sampling step and $300$ visual words, resulting in a $1800$ dimensional feature. 2) Pairwise rotation invariant co-occurrence local binary pattern (PRI-CoLBP) feature, which was proven to be suitable for scene classification \cite{PRI-CoLBP}. Different from other LBP variants, PRI-CoLBP not only captured the spatial context co-occurrence information effectively, but also possessed rotation invariance. We use gray-scale PRI-CoLBP and choose the simplest template so that the final dimensionality is $590 \times 2 = 1180$. 3) CENsus TRansform hISTogram (CENTRIST) feature \cite{centrist}. It is a holistic representation which can capture structural properties such as rectangular shapes, flat surfaces and so on. By using the spatial pyramid technology, there are $1$, $5$, and $25$ blocks for levels $0$, $1$, and $2$, respectively. We use PCA to reduce the dimensionality of CENTRIST to $40$, then a level $2$ pyramid will result in a feature vector which has $40\times (1+5+25) = 1240$ dimensions.



\begin{figure*}
  \centering
  \renewcommand{\figurename}{Figure}
  \subfloat[t-SVD-MSC]{
    \includegraphics[width=0.49\textwidth]{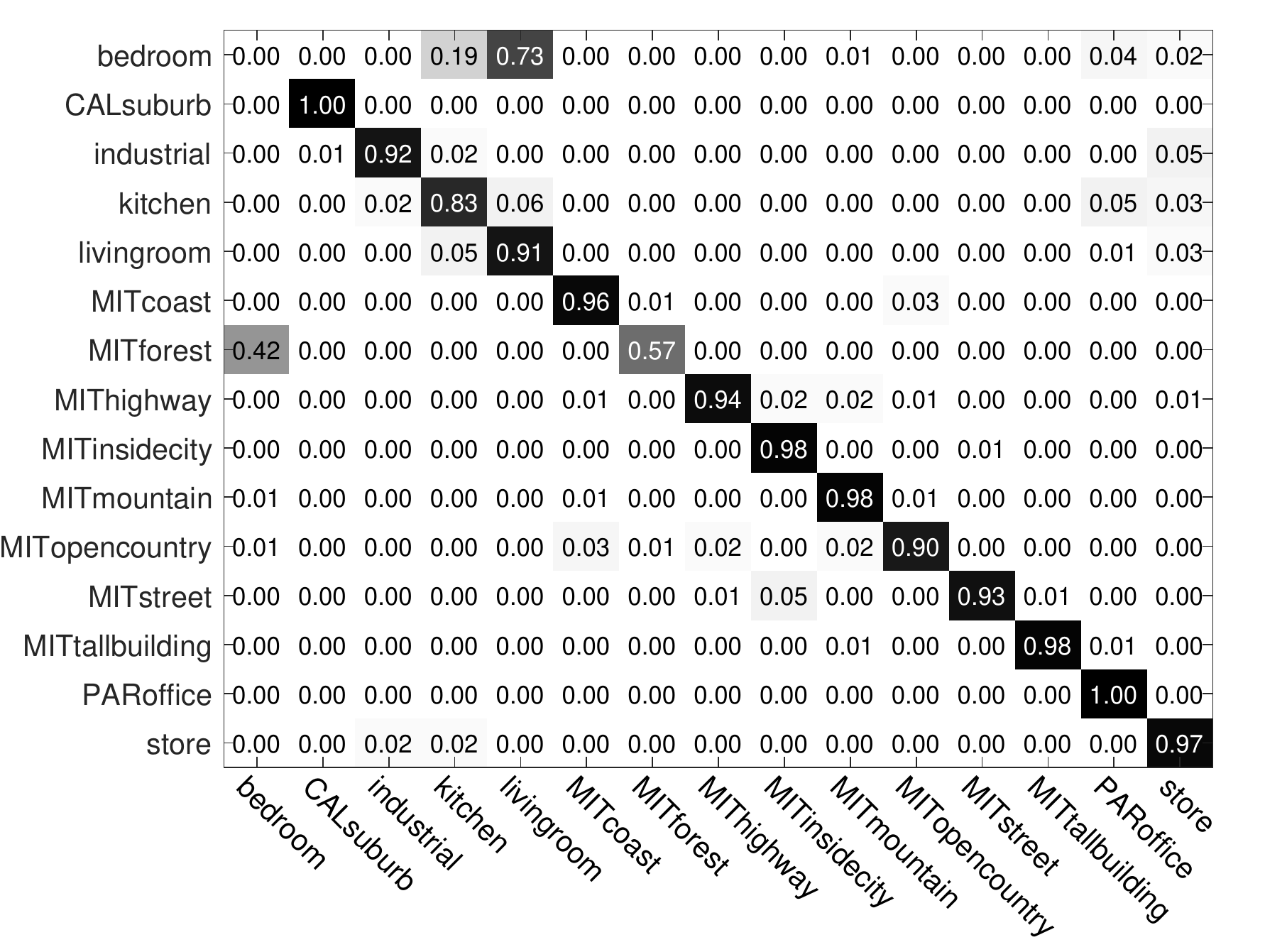}}
  \hspace{-0.2in}
  \subfloat[LTMSC]{
    \includegraphics[width=0.49\textwidth]{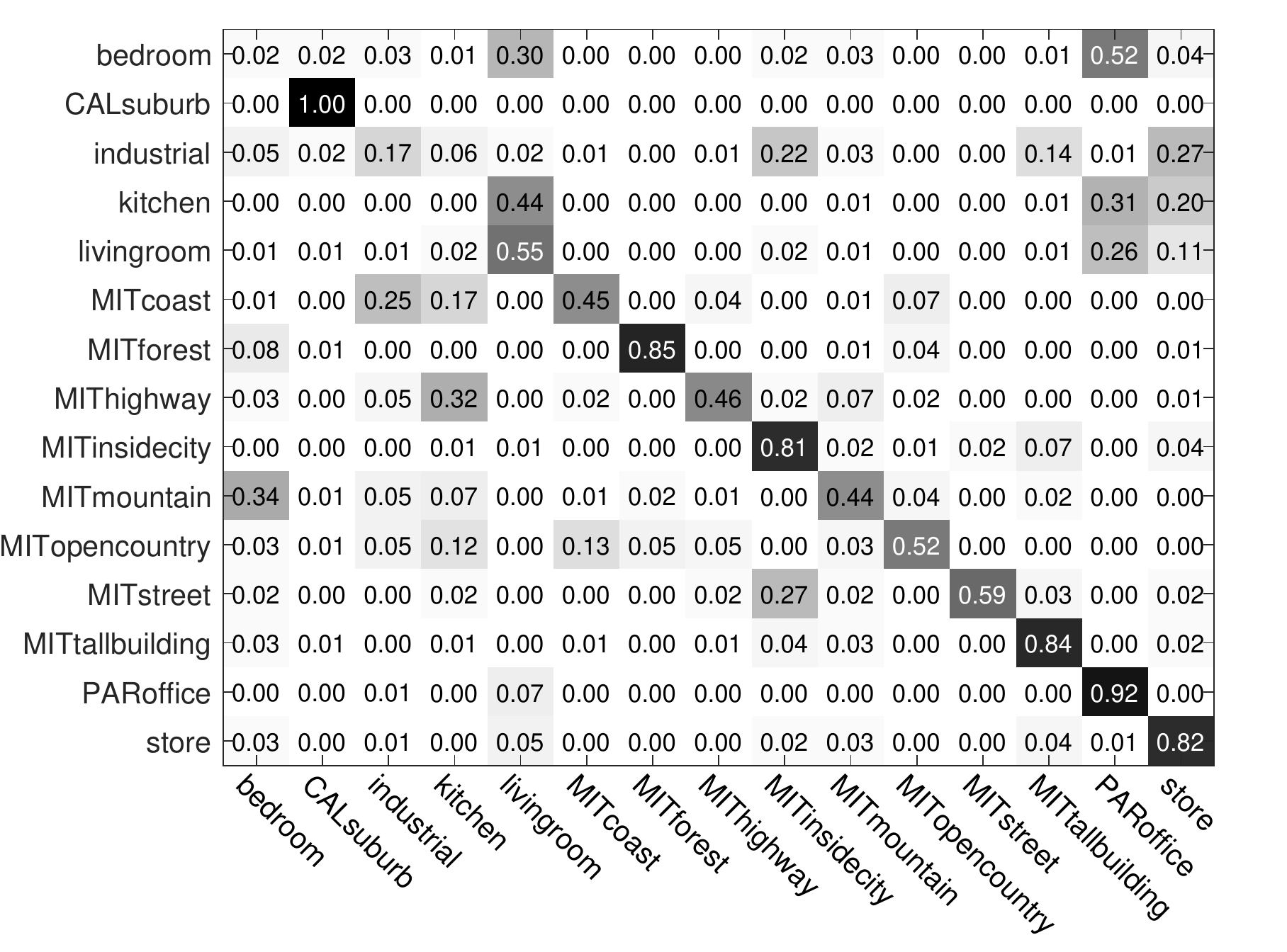}}
  \caption{{\color{red}Comparison the confusion matrices between LTMSC and the proposed t-SVD-MSC on {\it Scene-15} dataset.}}
  \label{fig:confusion-matrix} 
\end{figure*}

The clustering results are shown in Table \ref{result-Scene15}, where the noticeable performance gain can be concluded by comparing with the second best LTMSC algorithm. Moreover, confusion matrices of the LTMSC and the proposed method is shown in Fig. \ref{fig:confusion-matrix}, where row and column names are true and predicted labels respectively. Here, the cluster label is predicted by the best permutation mapping function used in the metric of ACC \cite{hungarian}. We can see that, compared with LTMSC, the proposed method wins in almost all categories in terms of clustering accuracy. {\color{red}The biggest confusion occurring between the indoor classes, such as bedroom and living room, coincides well with the the confusion distribution in \cite{scene-15c}}.

\begin{figure}[htb]
\setlength{\abovecaptionskip}{3pt}
\setlength{\belowcaptionskip}{0pt}
\renewcommand{\figurename}{Figure}
\centering
\includegraphics[width=0.5\textwidth]{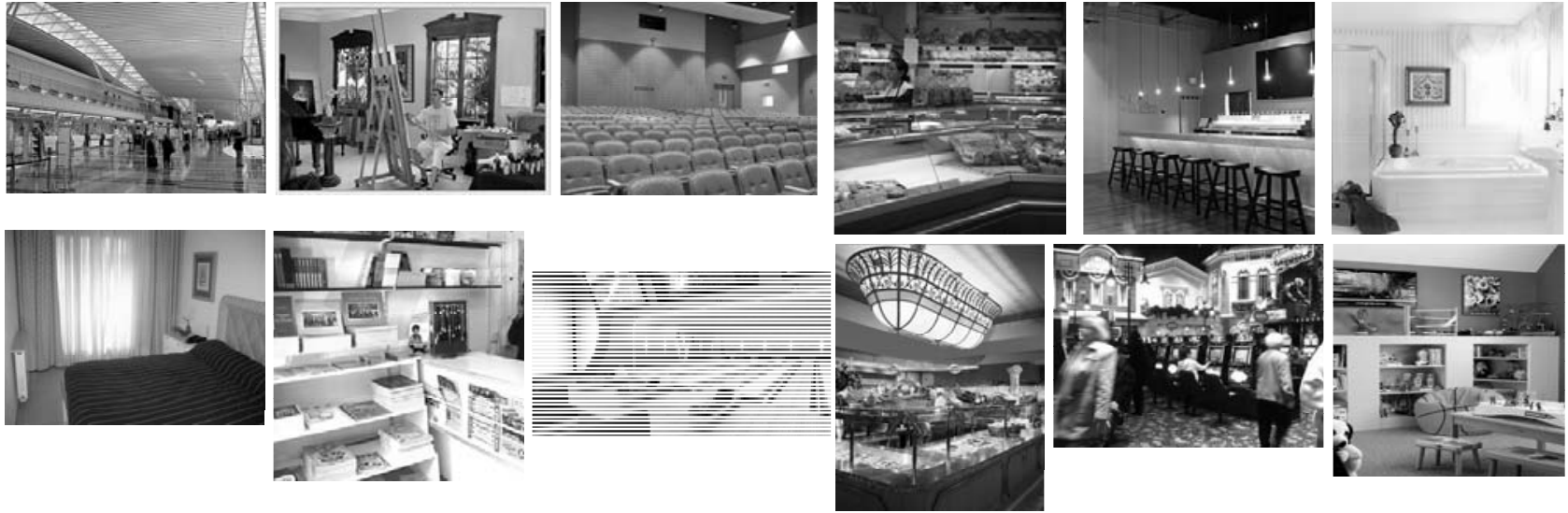}
\caption{Samples from MITIndoor-67 dataset.}
\label{fig:MITIndoor67-samples}
\end{figure}

{\it MITIndoor-67} dataset was firstly introduced by \cite{mitindoor}, which is a challenging dataset including $15$K indoor image spanning $67$ different categories. It provides a training subset ($5360$ images) for classification task, and we perform clustering on this subset. Some samples are shown in Fig. \ref{fig:MITIndoor67-samples}. To the best of our knowledge, hardly any traditional clustering methods can achieve good performance in such a challenging dataset. To pursuit better performance, besides the features used in Scene-15, we further import the VGG-VD \cite{VGG-VD}, which was pre-trained on ILSVRC12 \cite{imagenet}, as a new view to complement handcrafted features. We use the activations of the penultimate layer for feature extraction, and resize its smaller dimension of each image to $448$ for VGG19 while maintaining aspect ratio. The features are extracted from $5$ scales $\{2^{s}, s = -1, -0.5, 0, 0.5, 1\}$, and all local features are pooling together regardless of scales and locations. The MatConvNet toolbox \cite{matconvnet} is adopted to extract this feature. 

\begin{figure}
  \centering
  \renewcommand{\figurename}{Figure}
  \subfloat[]{
    \includegraphics[width=0.4\textwidth]{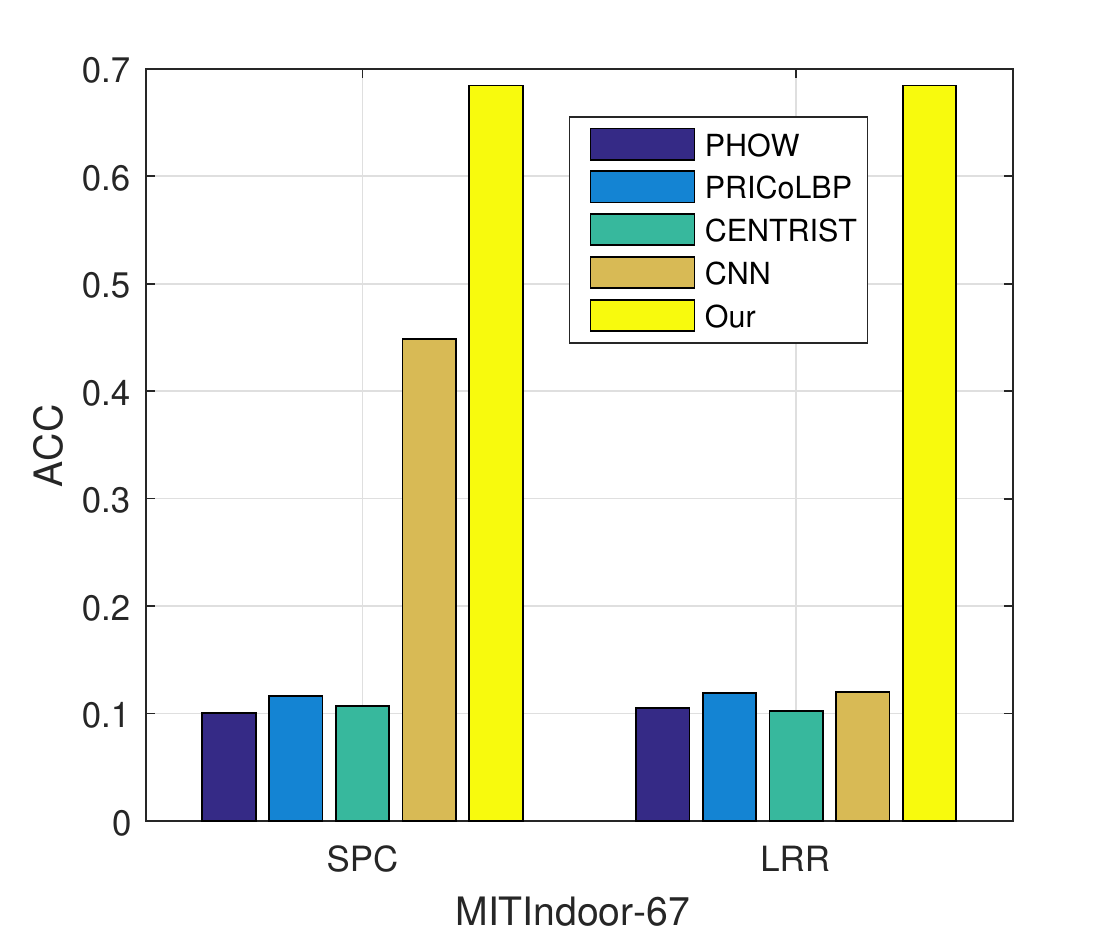}}
  \hspace{-0.2in}
  \subfloat[]{
    \includegraphics[width=0.4\textwidth]{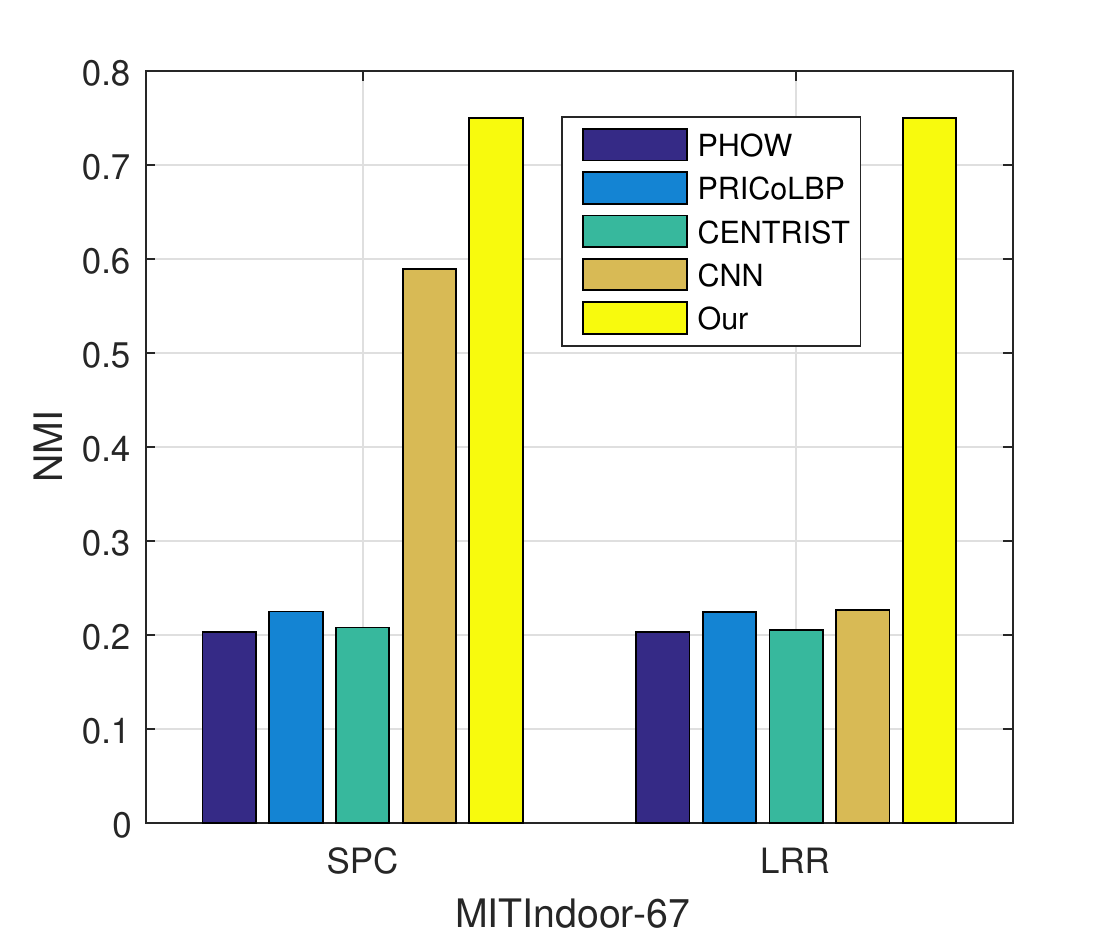}}
  \caption{Comparison between SPC/LRR with the single view feature and the proposed t-SVD-MSC in terms of accuracy and NMI on MITIndoor-67 dataset.}
  \label{fig:MITIndoor-SPC-LRR-View-compare} 
\end{figure}

Compared with GSNMF-CNN, our method gains significant improvement around $7.7\%$, $16.7\%$, $29.1\%$, $19.0\%$, $17.6\%$ and $20.1\%$ in terms of NMI, ACC, AR, F-score, Precision and Recall, respectively. Fig. \ref{fig:MITIndoor-SPC-LRR-View-compare} illustrates the comparison between SPC/LRR with different single view feature and the proposed t-SVD-MSC with multiview features. It can be observed that, the performance of SPC with raw VGG19 feature is much higher than that of SPC with traditional handcrafted features, so CNN feature is indeed an excellent representation even without any transfer. However, representing CNN feature in low-rank subspace will significantly degrade the performance, see the brown bar in LRR. The yellow bar in Fig. \ref{fig:MITIndoor-SPC-LRR-View-compare} indicates that, the proposed method could capture the complementarity between the handcrafted features and CNN feature, and boost the performance to a higher level.



\subsection{Experiments on Generic Clustering}

The {\it COIL-20}\footnote{http://www.cs.columbia.edu/CAVE/software/softlib/} dataset contains $1440$ images of $20$ object categories viewed from varying angels, with each category including $72$ images. Similar to \cite{LRTM,DiMSC}, all the images are normalized to $32\times 32$ with the same features used in Section \ref{face-clustering} being extracted. As shown in Table \ref{result-COIL20}, our method also outperforms three most recently published algorithms, {\it i.e.,} RMSC, DiMSC, and LTMSC, which further demonstrates the effectiveness of the proposed method.

The {\it Caltech-101} dataset \cite{Caltech101} contains $8677$ image of objects belonging to $101$ categories, with about $40$ to $800$ images per category. Currently, image clustering method are usually evaluated under small-scale experimental configuration, {\it e.g.,} using relatively simple datasets (such as Yale and ORL), or cropping a small portion of categories from a large dataset (such as using $5$, $7$ and $20$ sub-categories of the Caltech-101 dataset \cite{canyi-MVSSC,MVSC}). Here, we use the instances from all the categories to test whether the proposed method could handle relatively large and challenging dataset, and compare the clustering performance with the state-of-the-art unsupervised CNN-based clustering method, {\it i.e.,} GSNMF-CNN. We use the deep feature Inception V3 \cite{InceptionV3} in this dataset, since it achieves better results than VGG19. It is also extracted from the activations of the penultimate layer, leading to a $2048$-dimensional feature vector. The same feature is used in GSNMF-CNN.

The results are shown in Table \ref{result-Caltech101}. By using the more powerful deep feature, the baseline algorithms (SPC and LRR) perform better than some sophisticated methods such as RMSC and DiMSC. This is probably because RMSC and DiMSC suffer from the less representation capabilities of the handcrafted features on this dataset. Surprisingly, the LTMSC is not affected by some degenerate views and even does sightly better than GSNMF-CNN. Furthermore, in the testing of all the datasets, three observations need to be worth noting: (1) The method Ut-SVD-MSC sometimes shows comparable performance to other state-of-the-art approaches, but still has a significant gap with regard to the proposed method. Because Ut-SVD-MSC does not make full use of the structure of the self-representations coefficient tensor, while the proposed method preserves those coefficients in Fourier domain, as illustrated in Fig. \ref{fig:shift-tSVD}. (2) Due to noise or error in measurement, some of the available views may be misleading in revealing the true structure of the data, so that including them in the clustering process may have negative influence. The corresponding phenomena appear several times, for example, DiMSC on Scene-15, (RMSC, DiMSC, and LTMSC) on MITIndoor-67, and (RMSC and DiMSC) on Caltech-101, where their performances are worse than directly using spectral clustering with single best feature. On the contrary, the proposed method exhibits robustness to the existence of degenerate views. (3) CNN feature is usually better than handcrafted features in terms of representation capability. But this does not mean that it is enough for clustering only by using CNN feature. The performance gains of the proposed method on all these datasets confirm the ``complementary principle'' in the multi-view learning, which states that each view of the data may contain some knowledge that other views do not have. The complementary information between CNN feature and handcrafted features can help to improve the clustering performance.

\begin{table*}[!ht]
\footnotesize
\renewcommand{\arraystretch}{1.3}
\begin{centering}
\begin{threeparttable}[]
\caption{{\color{red}Clustering results (mean $\pm$ standard deviation) on {\it COIL-20}. We set $\lambda = 0.25$ in proposed t-SVD-MSC.}}\label{result-COIL20}
\tabcolsep=4pt
\begin{minipage}{12cm}
\begin{tabular}{@{}l|cccccc}
\toprule[2pt] 
Method                        &NMI &ACC &AR &F-score &Precision &Recall \\ \hline
$\text{SPC}_{\text{best}}$    &0.806 $\pm$ 0.008 &0.672 $\pm$ 0.063 &0.619 $\pm$ 0.018 &0.640 $\pm$ 0.017 &0.596 $\pm$ 0.021 &0.692 $\pm$ 0.013\\
$\text{LRR}_{\text{best}}$    &0.829 $\pm$ 0.006 &0.761 $\pm$ 0.003 &0.720 $\pm$ 0.020 &0.734 $\pm$ 0.006 &0.717 $\pm$ 0.003 &0.751 $\pm$ 0.002\\ \hline
RMSC                          &0.800 $\pm$ 0.017 &0.685 $\pm$ 0.045 &0.637 $\pm$ 0.044 &0.656 $\pm$ 0.042 &0.620 $\pm$ 0.057 &0.698 $\pm$ 0.026\\
DiMSC                         &0.846 $\pm$ 0.002 &0.778 $\pm$ 0.022 &0.732 $\pm$ 0.005 &0.745 $\pm$ 0.005 &0.739 $\pm$ 0.007 &0.751 $\pm$ 0.003\\
LTMSC                         &0.860 $\pm$ 0.002 &0.804 $\pm$ 0.011 &0.748 $\pm$ 0.004 &0.760 $\pm$ 0.007 &0.741 $\pm$ 0.009 &0.776 $\pm$ 0.006\\
Ut-SVD-MSC                    &0.841 $\pm$ 0.004 &0.788 $\pm$ 0.005 &0.732 $\pm$ 0.003 &0.746 $\pm$ 0.006 &0.731 $\pm$ 0.002 &0.760 $\pm$ 0.002\\
t-SVD-MSC                     &\textbf{0.884 $\pm$ 0.005} &\textbf{0.830 $\pm$ 0.000} &\textbf{0.786 $\pm$ 0.003} &\textbf{0.800 $\pm$ 0.004} &\textbf{0.785 $\pm$ 0.007} &\textbf{0.808 $\pm$ 0.001}\\\hline \bottomrule[1pt]
\end{tabular}
\end{minipage}
\end{threeparttable}
\end{centering}
\end{table*}

\begin{table*}[!ht]
\footnotesize
\renewcommand{\arraystretch}{1.3}
\begin{centering}
\begin{threeparttable}[]
\caption{{\color{red}Clustering results (mean $\pm$ standard deviation) on {\it Caltech-101}. We set $\lambda = 0.5$ in proposed t-SVD-MSC.}}\label{result-Caltech101}
\tabcolsep=4pt
\begin{minipage}{12cm}
\begin{tabular}{@{}l|cccccc}
\toprule[2pt] 
Method         &NMI &ACC &AR &F-score &Precision &Recall \\ \hline
$\text{SPC}_{\text{best}}^{\text{CNN}}$    &0.723 $\pm$ 0.032 &0.484 $\pm$ 0.019 &0.319 $\pm$ 0.014 &0.340 $\pm$ 0.025 &0.597 $\pm$ 0.018 &0.235 $\pm$ 0.020\\
$\text{LRR}_{\text{best}}^{\text{CNN}}$    &0.728 $\pm$ 0.014 &0.510 $\pm$ 0.009 &0.304 $\pm$ 0.017 &0.339 $\pm$ 0.008 &0.627 $\pm$ 0.012 &0.231 $\pm$ 0.010\\ \hline
RMSC                                       &0.573 $\pm$ 0.047 &0.346 $\pm$ 0.036 &0.246 $\pm$ 0.031 &0.258 $\pm$ 0.027 &0.457 $\pm$ 0.033 &0.182 $\pm$ 0.031\\
DiMSC                                      &0.589 $\pm$ 0.011 &0.351 $\pm$ 0.008 &0.226 $\pm$ 0.003 &0.253 $\pm$ 0.007 &0.362 $\pm$ 0.010 &0.191 $\pm$ 0.007\\
LTMSC                                      &0.788 $\pm$ 0.005 &0.559 $\pm$ 0.012 &0.393 $\pm$ 0.007 &0.403 $\pm$ 0.003 &0.670 $\pm$ 0.009 &0.288 $\pm$ 0.012\\
GSNMF-CNN                                  &0.775 $\pm$ 0.010 &0.534 $\pm$ 0.012 &0.246 $\pm$ 0.008 &0.275 $\pm$ 0.006 &0.230 $\pm$ 0.004 &\textbf{0.347 $\pm$ 0.006} \\
Ut-SVD-MSC                                 &0.742 $\pm$ 0.008 &0.483 $\pm$ 0.003 &0.334 $\pm$ 0.002 &0.344 $\pm$ 0.004 &0.612 $\pm$ 0.002 &0.239 $\pm$ 0.002\\
t-SVD-MSC                                  &\textbf{0.858 $\pm$ 0.003} &\textbf{0.607 $\pm$ 0.005} &\textbf{0.430 $\pm$ 0.005} &\textbf{0.440 $\pm$ 0.010} &\textbf{0.742 $\pm$ 0.007} &0.323 $\pm$ 0.009\\\hline \bottomrule[1pt]
\end{tabular}
\end{minipage}
\end{threeparttable}
\end{centering}
\end{table*}

\subsection{Model Analysis}\label{model-analysis}
\begin{figure*}
  \centering
  \renewcommand{\figurename}{Figure}
  \subfloat[ORL]{
    \includegraphics[width=0.46\textwidth]{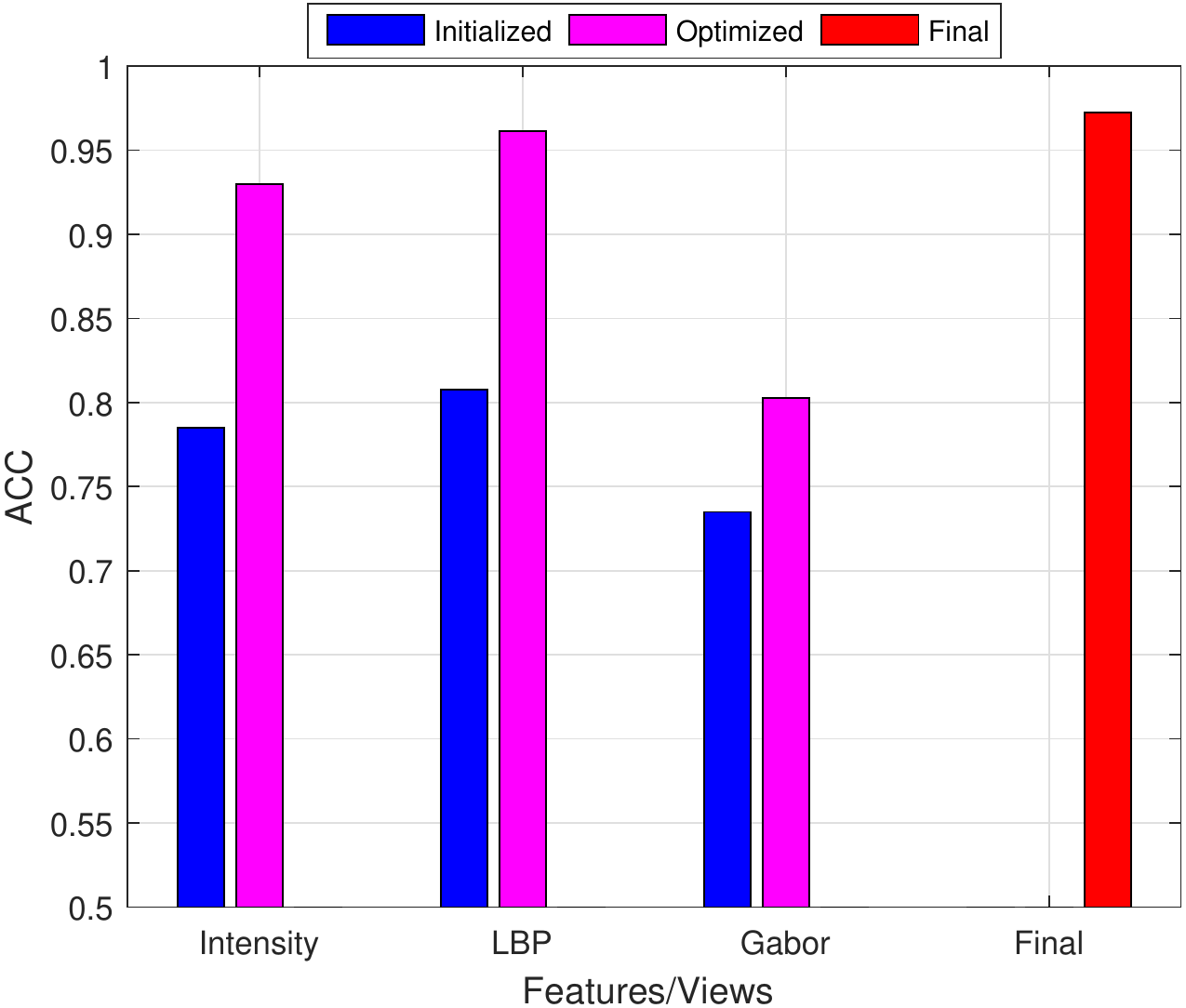}}
  \hspace{0.15in}
  \subfloat[MITIndoor-67]{
    \includegraphics[width=0.46\textwidth]{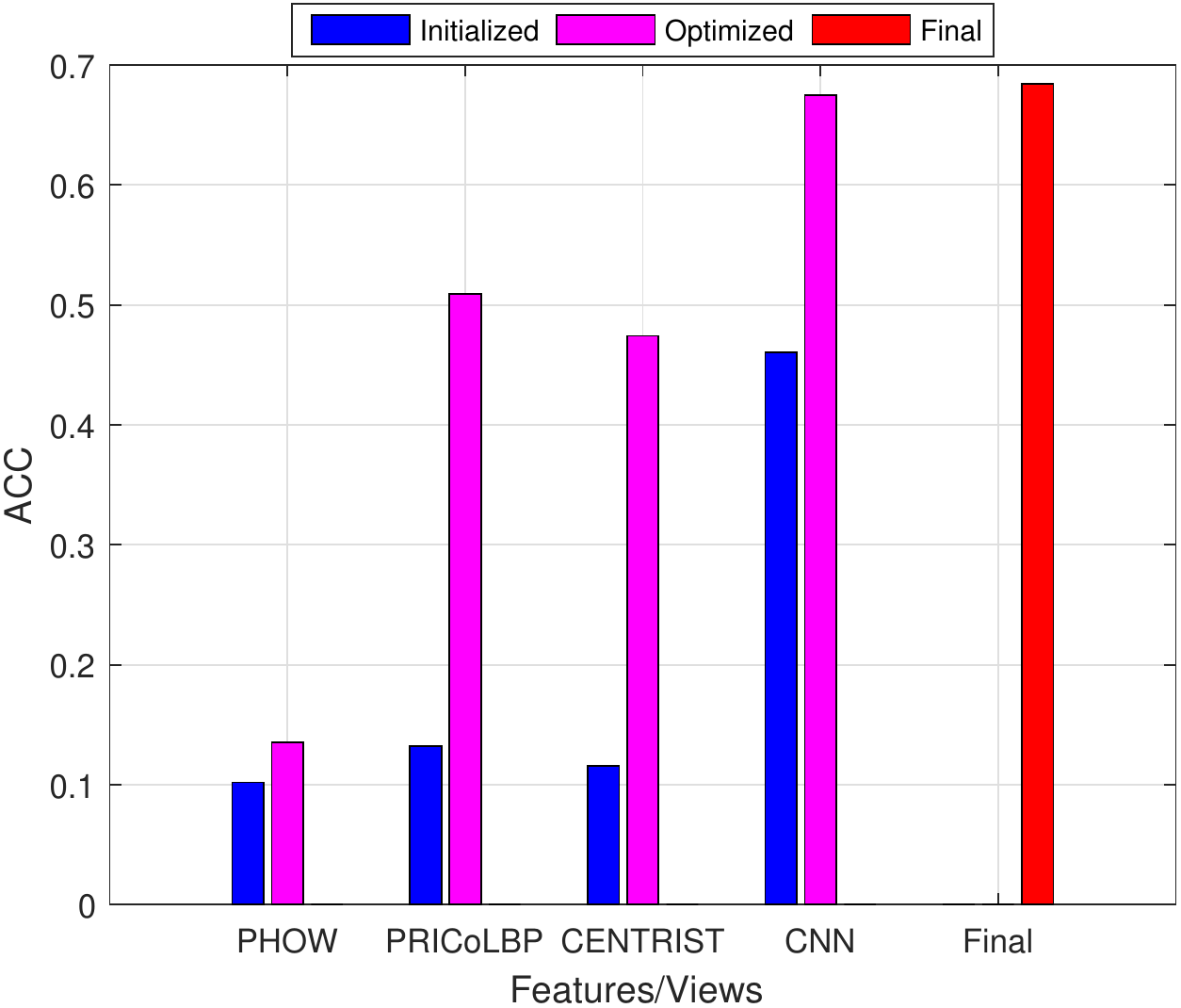}}
  \caption{{\color{red}The comparison of clustering accuracy by using coefficient matrices $\{\mathbf{Z}^{(v)}\}_{v=1}^{V}$ before (blue bar) and after (magenta bar) optimization on ORL and MITIndoor-67 datasets. }}
  \label{fig:before_and_after} 
\end{figure*}

{\color{red}
\subsubsection{Contributions of Multi-View Features}\label{view-contribution}

\begin{figure*}
  \centering
  \renewcommand{\figurename}{Figure}
  \vspace{0.1in}
  \subfloat[Intensity]{
    \includegraphics[width=0.4\textwidth]{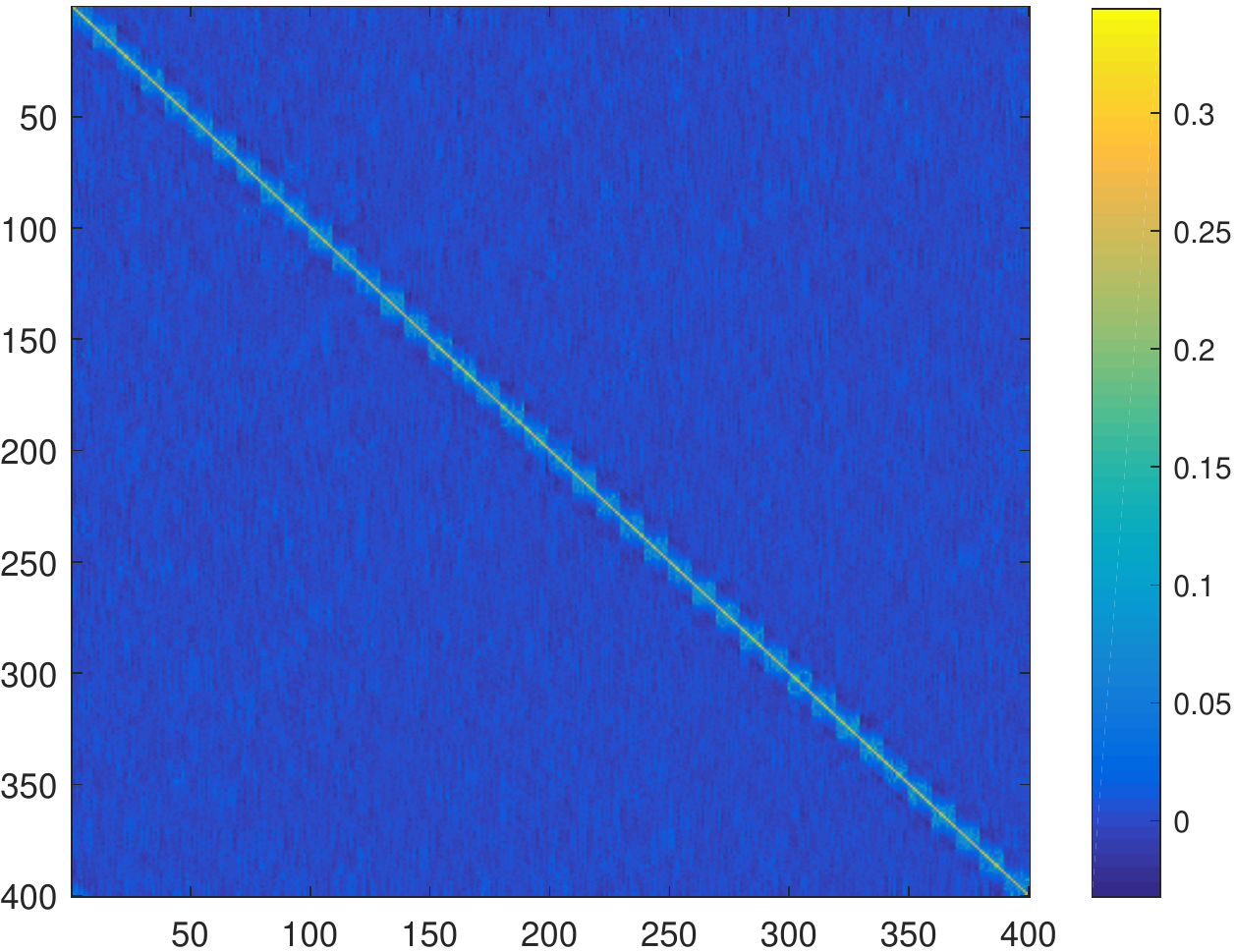}}
  \hspace{0.15in}
  \subfloat[LBP]{
    \includegraphics[width=0.4\textwidth]{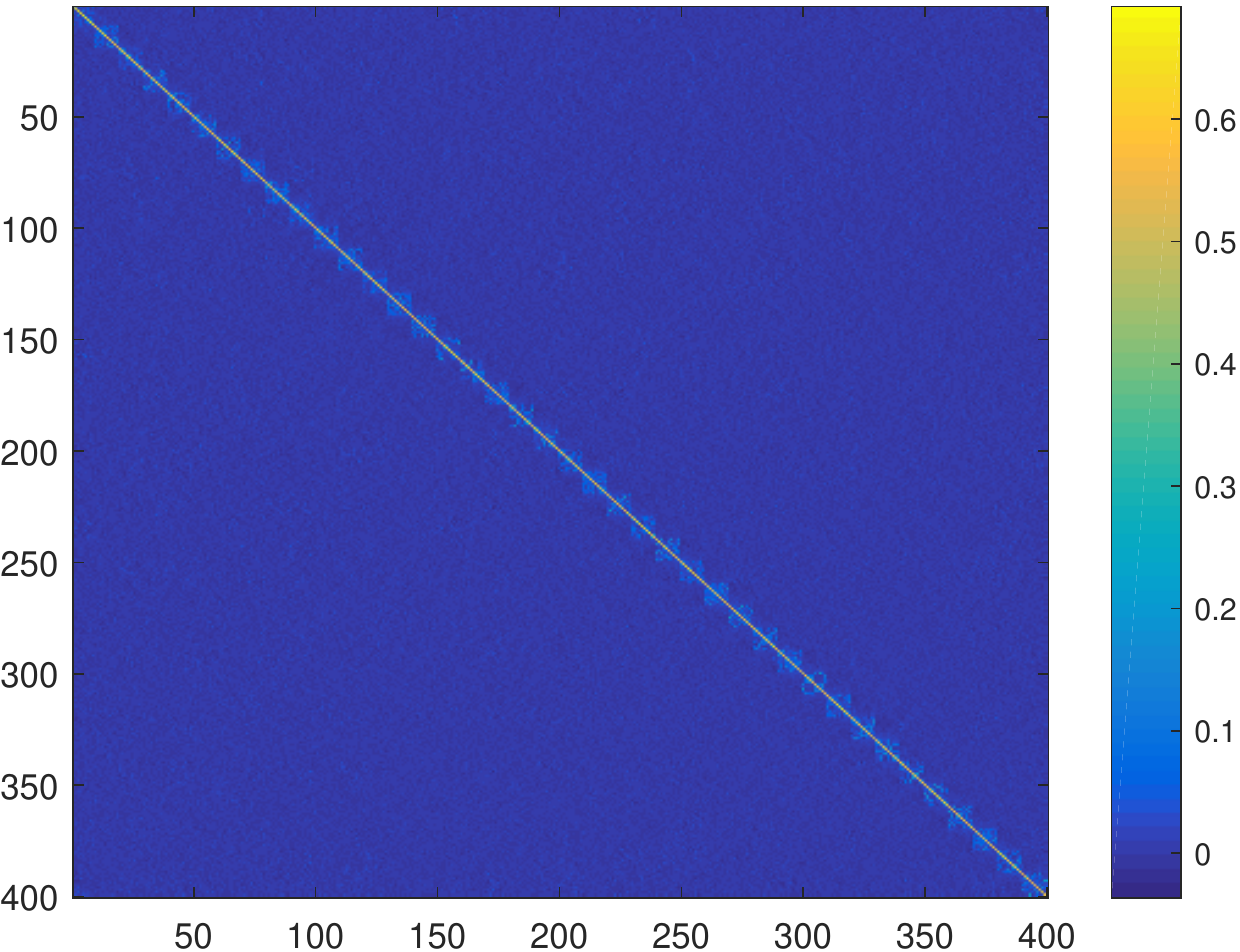}}\\
  \vspace{0.01in}
  \subfloat[Gabor]{
    \includegraphics[width=0.4\textwidth]{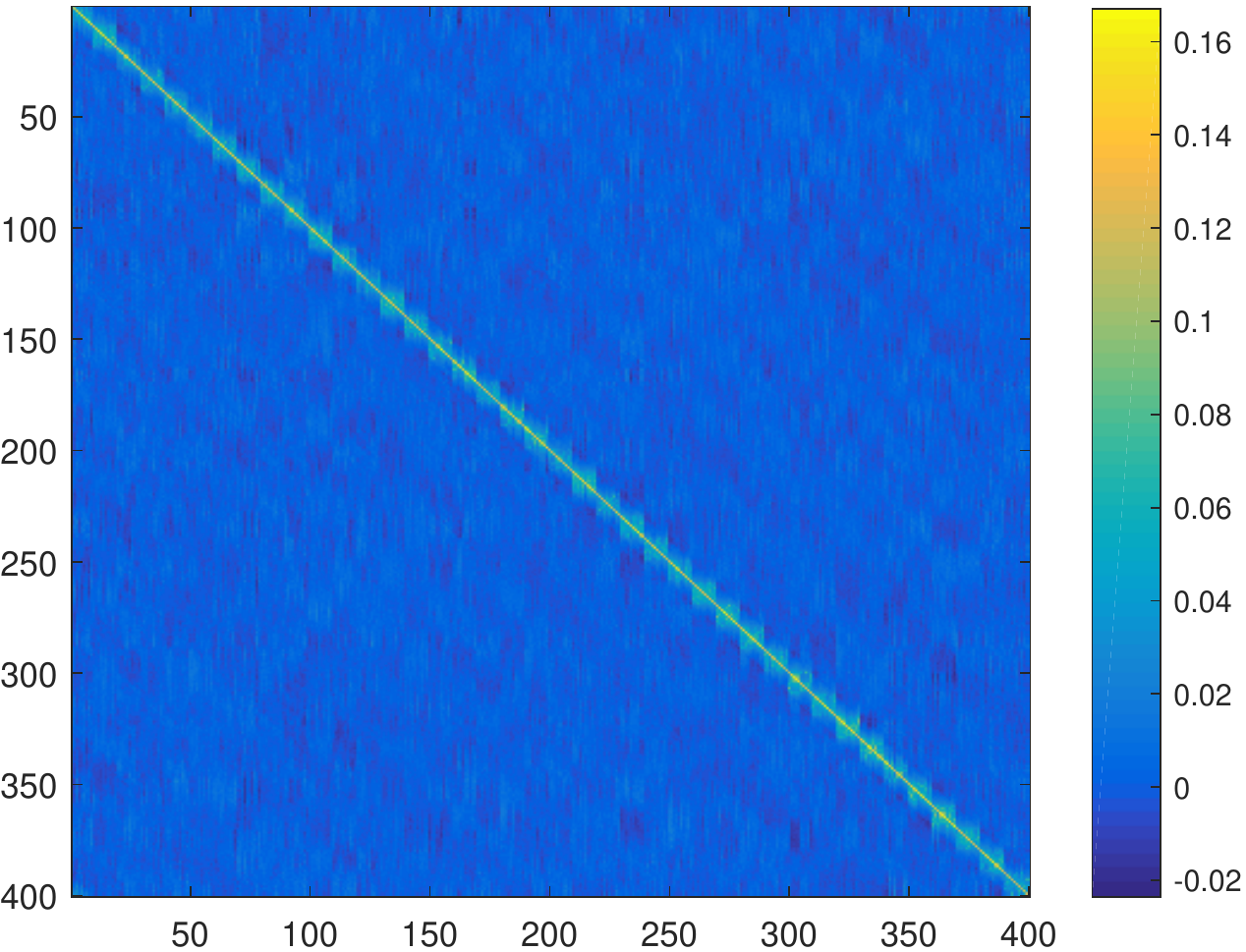}}
  \hspace{0.15in}
  \subfloat[Fused]{
    \includegraphics[width=0.4\textwidth]{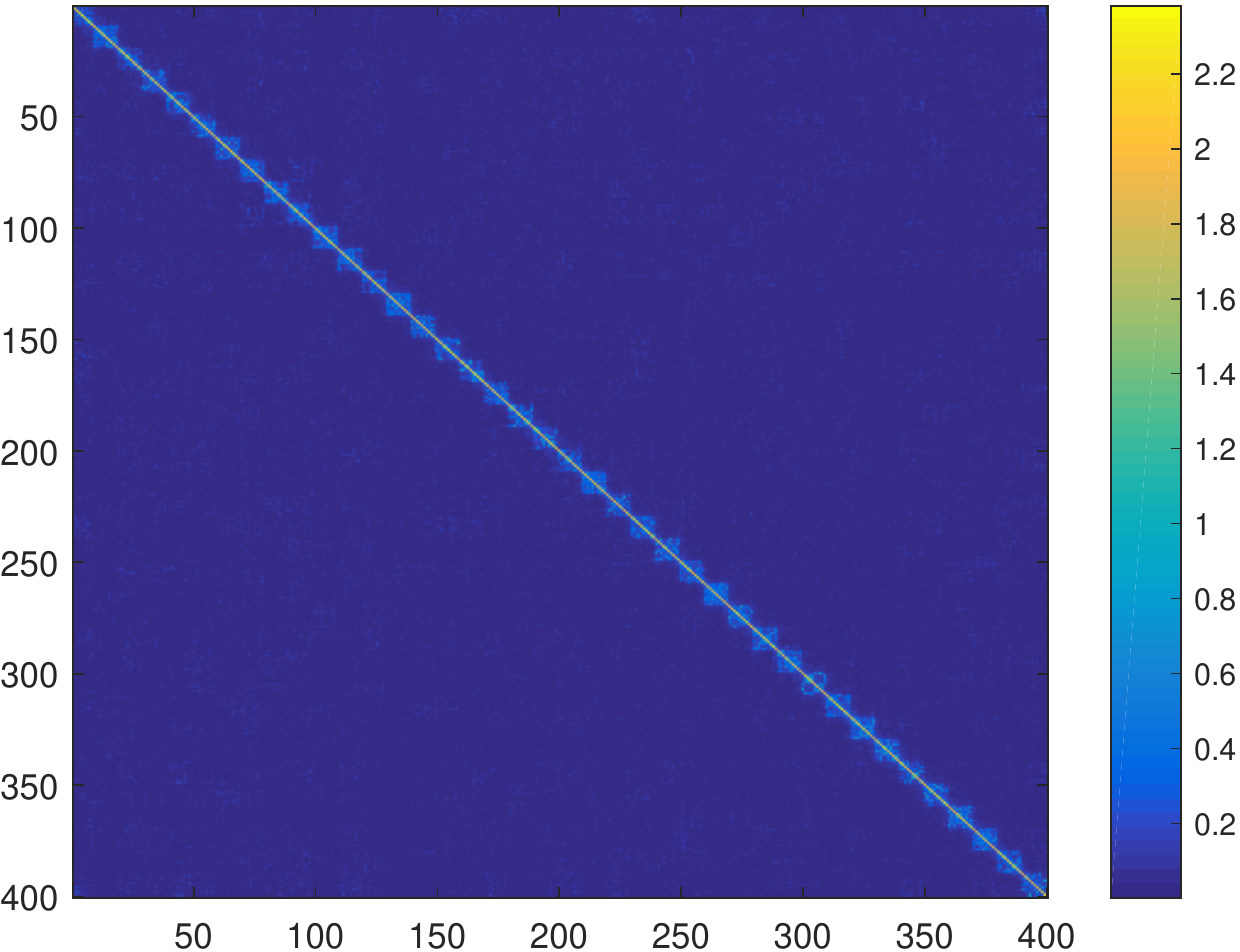}}
  \caption{{\color{red}(a)$\sim$(c) The illustration of affinity matrices $\mathbf{A}^{(v)} = \frac{1}{2}(|\mathbf{Z}^{(v)}| + |\mathbf{Z}^{(v)^{\mathbf{T}}}|), v=1,2,3$ for all the views/features in ORL datasets. (d) The final affinity matrix $\mathbf{Z} = \frac{1}{V}\sum_{v=1}^{V}(|\mathbf{Z}^{(v)}| + |\mathbf{Z}^{(v)^{\mathbf{T}}}|)/2$.}}
  \label{fig:orl_affinity} 
\end{figure*}

\begin{figure*}
  \centering
  \renewcommand{\figurename}{Figure}
  \vspace{0.1in}
  \subfloat[PHOW]{
    \includegraphics[width=0.4\textwidth]{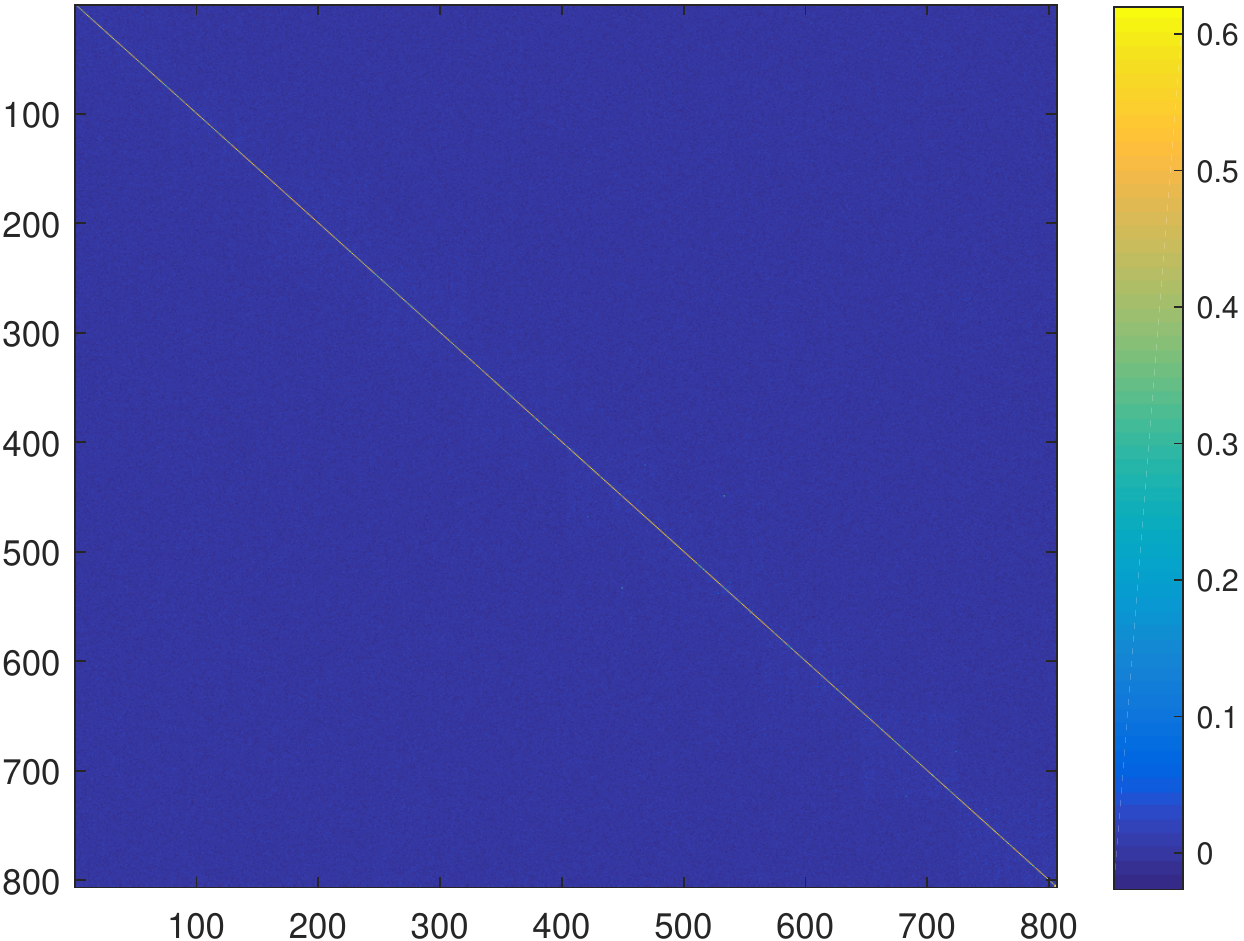}}
  \hspace{0.15in}
  \subfloat[PRICoLBP]{
    \includegraphics[width=0.4\textwidth]{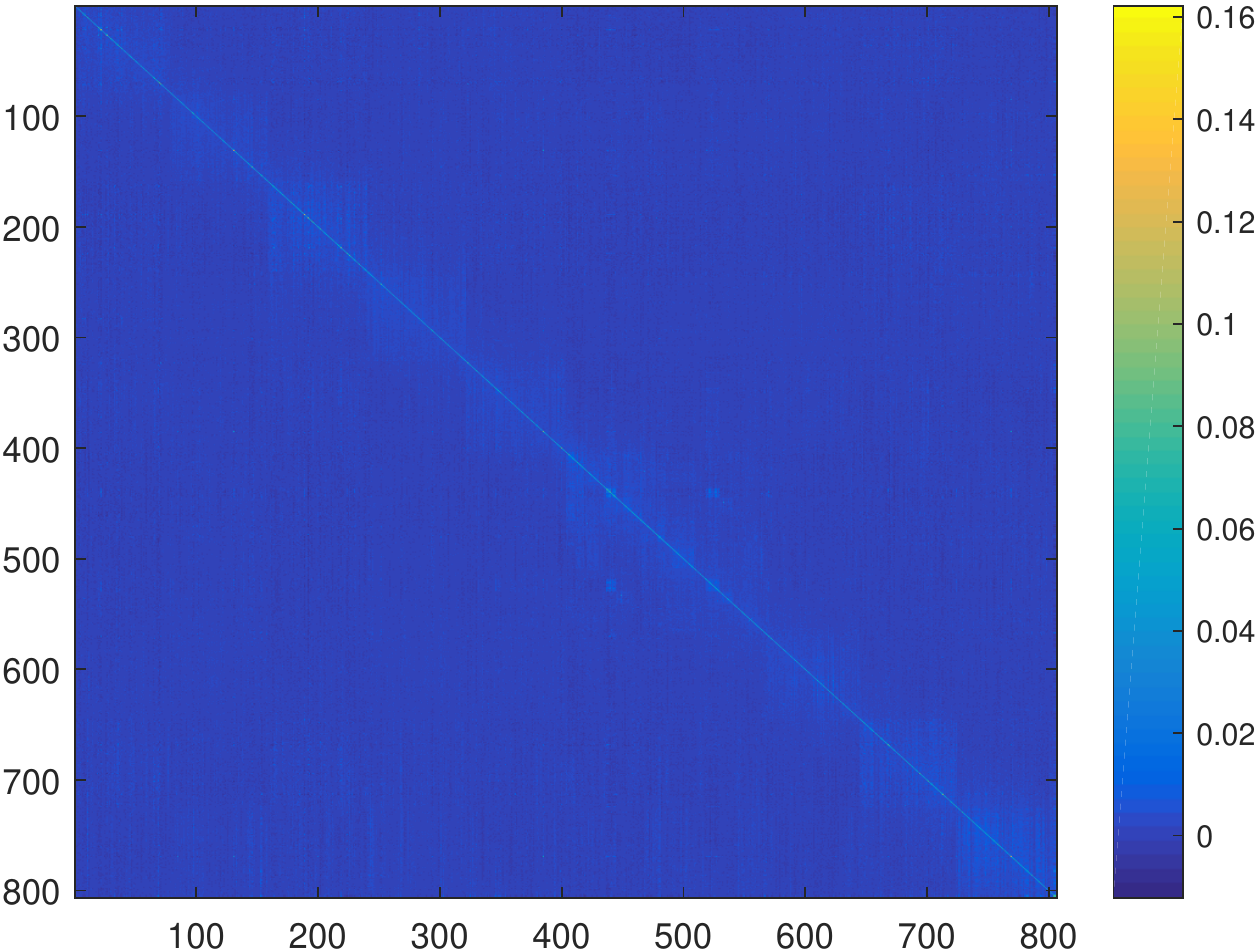}}\\
  \vspace{0.01in}
  \subfloat[CENTRIST]{
    \includegraphics[width=0.4\textwidth]{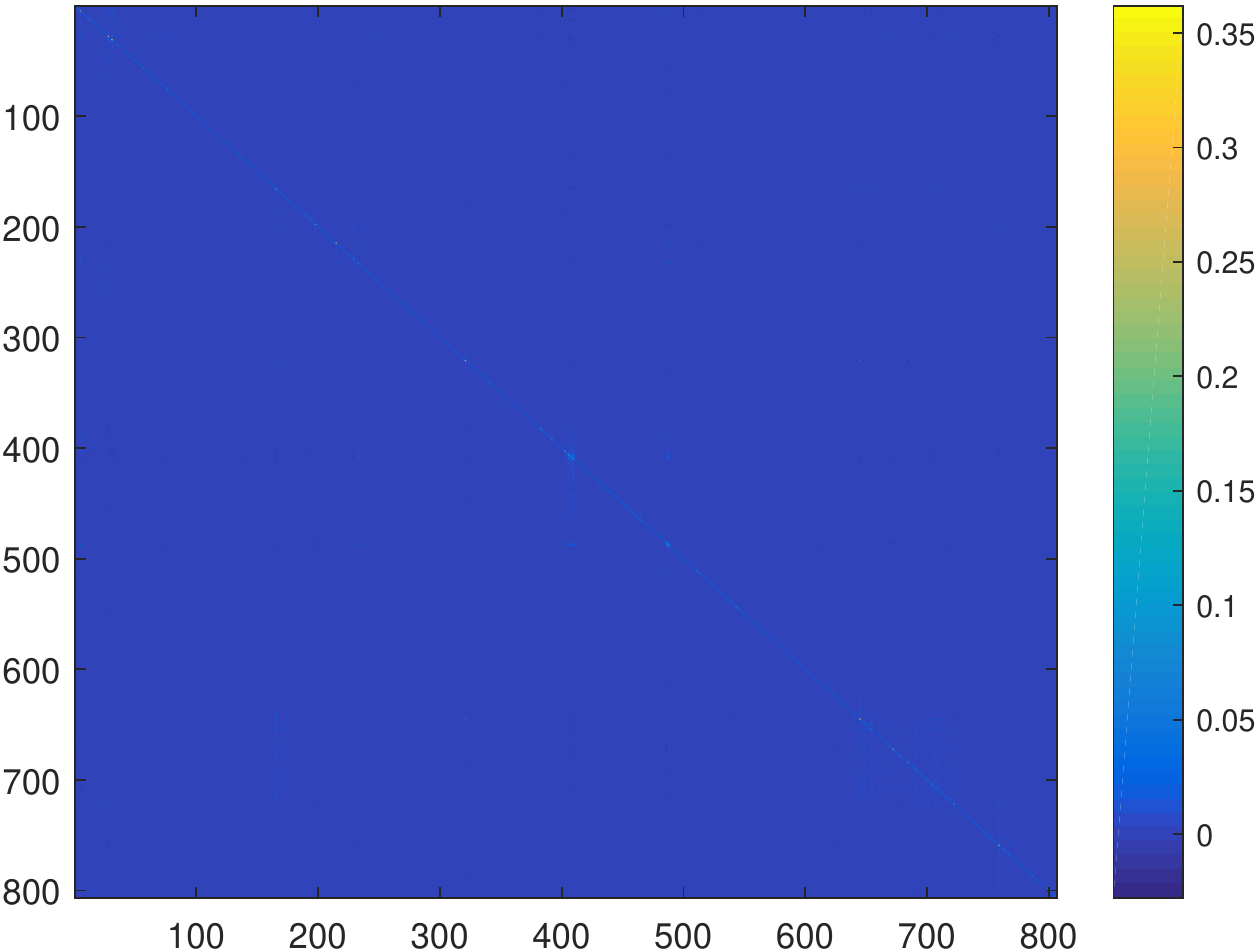}}
  \hspace{0.15in}
  \subfloat[CNN-VGG19]{
    \includegraphics[width=0.4\textwidth]{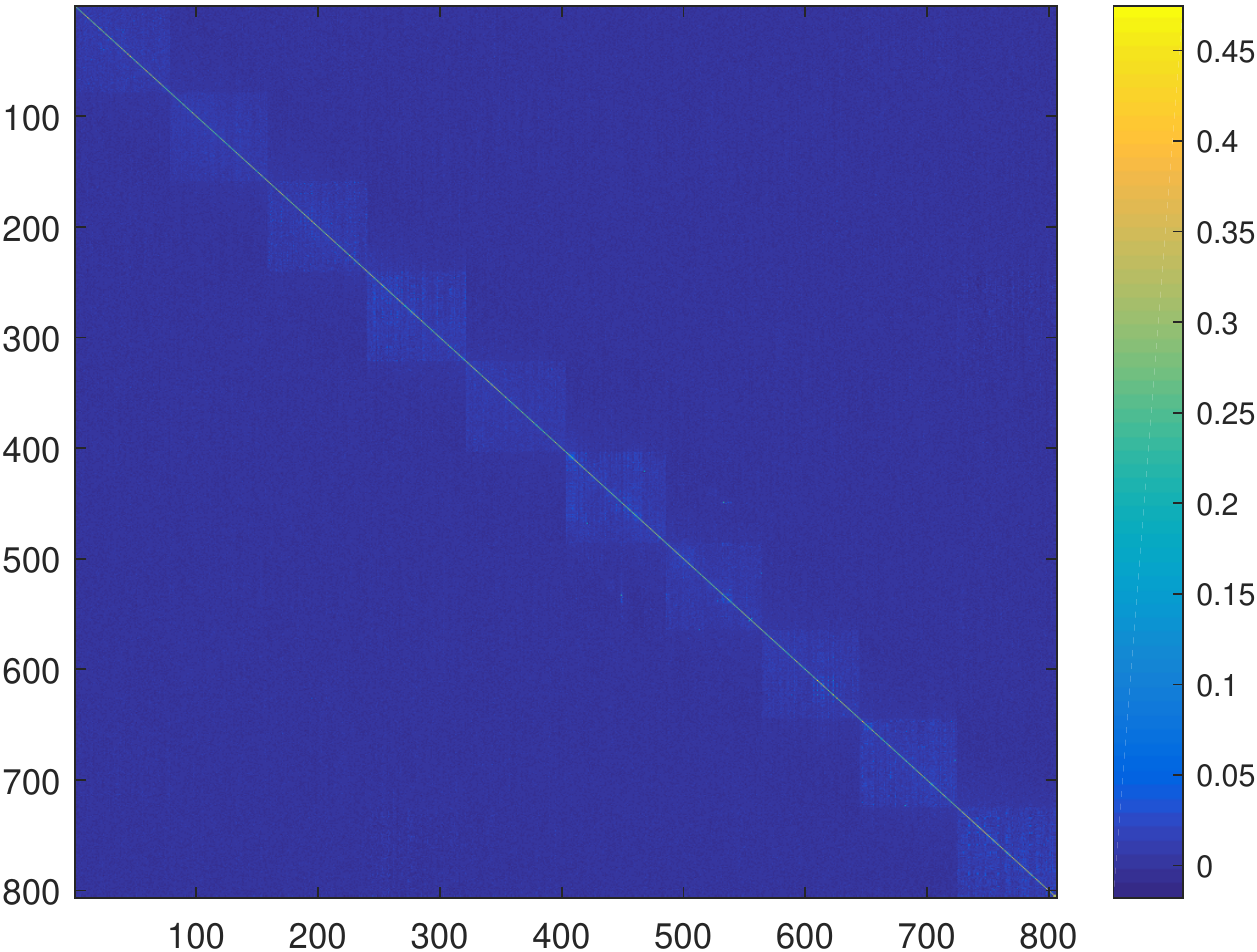}}\\
  \vspace{0.01in}
  \subfloat[Fused]{
    \includegraphics[width=0.4\textwidth]{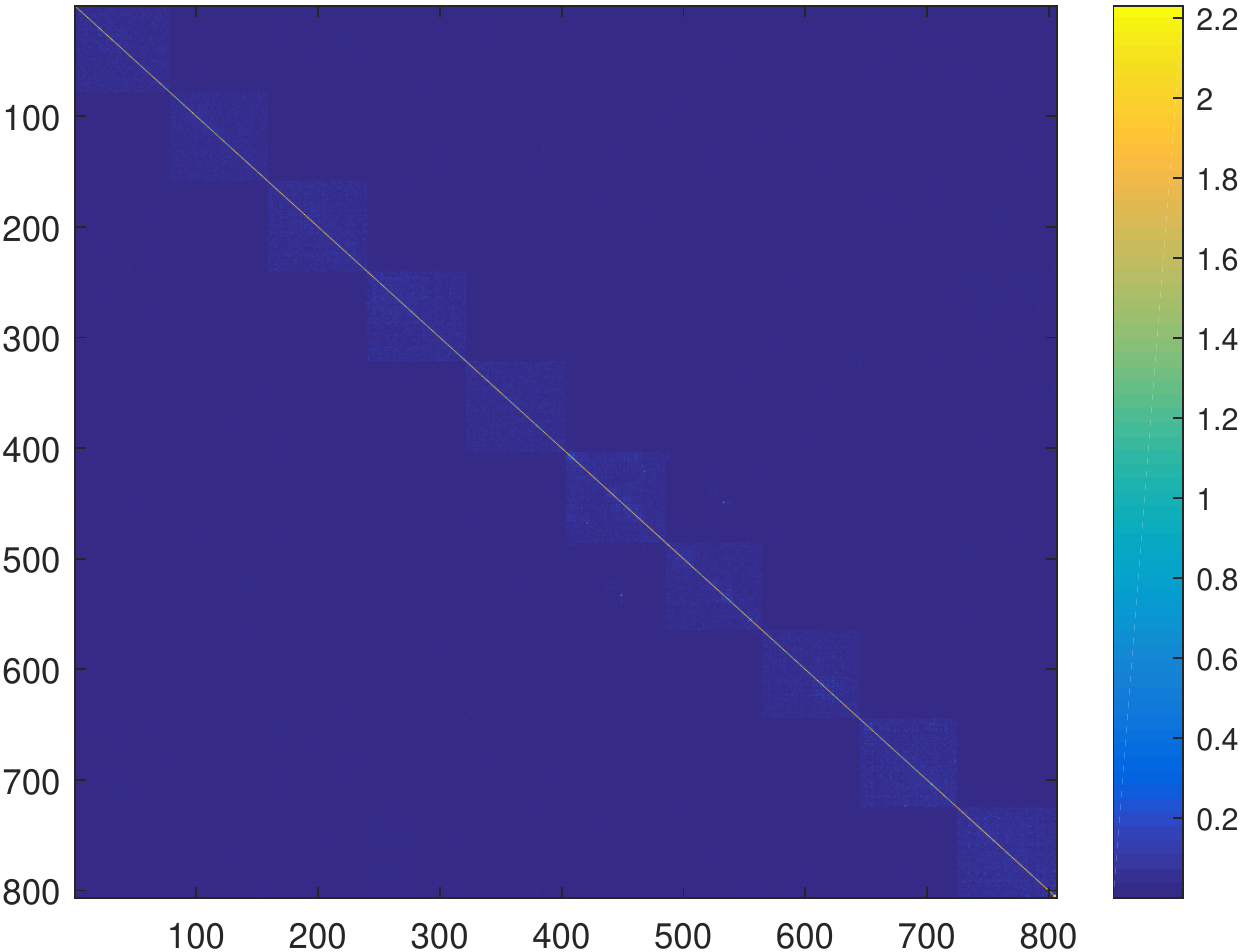}}
  \caption{{\color{red}(a)$\sim$(c) The illustration of affinity matrices $\mathbf{A}^{(v)} = \frac{1}{2}(|\mathbf{Z}^{(v)}| + |\mathbf{Z}^{(v)^{\mathbf{T}}}|), v=1,\ldots,4$ for all the views/features on MITIndoor-67 dataset. (d) The final affinity matrix $\mathbf{Z} = \frac{1}{V}\sum_{v=1}^{V}(|\mathbf{Z}^{(v)}| + |\mathbf{Z}^{(v)^{\mathbf{T}}}|)/2$. To see structure clearly, only the first ten clusters are displayed.}}
  \label{fig:mitindoor_affinity} 
\end{figure*}

In this subsection, we will analyze the contributions of multiple features to the final clustering result from the experimental perspective. In Fig. \ref{fig:orl_affinity}, we present the view-specific affinity matrices and the final affinity matrix for the ORL dataset. Since the LBP feature owns much more expressive capability than other two low-level feature in face description, the matrix corresponding to LBP (Fig. \ref{fig:orl_affinity} (b)) reveals the underlying clustering structures more clearly, which further validates our conclusion that discriminant feature contributes more to final result. Similar observation can be seen in Fig. \ref{fig:mitindoor_affinity}, where the block-diagonal structures are loomed and apparent for PRICoLBP and CNN-VGG19 features, respectively, while the affinity matrices for PHOW and CENTRIST features can hardly see the block-diagonal structures. This observation coincides with the conclusion that PRICoLBP is more suitable for scene classification than the other two handcrafted features [3], {\it i.e.,} PHOW and CENTRIST. Obviously, two discriminant features PRICoLBP and CNN-VGG19 contribute more to final clustering, which is demonstrated in the final affinity matrix in Fig. \ref{fig:mitindoor_affinity} (e).

Furthermore, we analyze the changes of affinity matrix for all the views before and after the proposed optimization procedure, so that the influence of the proposed model upon each view can be explored more thoroughly. To this end, the LRR solution of each feature (denotes by $\widetilde{\mathbf{Z}}^{(v)}$) is employed to initialize the $\mathbf{Z}^{(v)}$ (see the step $1$ in Algorithm 3) so as to obtain the optimized self-representation matrix (denotes by $\mathbf{Z}^{\ast(v)}$) in a unified tensor space. Fig. \ref{fig:before_and_after} shows the comparison of clustering accuracy by using $\{\mathbf{Z}^{(v)}\}_{v=1}^{V}$ before (blue bar) and after (magenta bar) optimization on the ORL and MITIndoor-67 datasets. Two key observations are listed as follows: 1) The feature type, which provides most contribution to final clustering, is kept the same before and after the optimization. 2) The performance of all the views are improved simultaneously, which is an evidence that the complementary information can be captured and propagated among all the views in high-order tensor space.}

\subsubsection{Parameter Tuning}

\begin{figure}
  \centering
  \renewcommand{\figurename}{Figure}
  \subfloat[ACC on Yale]{
    \includegraphics[width=0.35\textwidth]{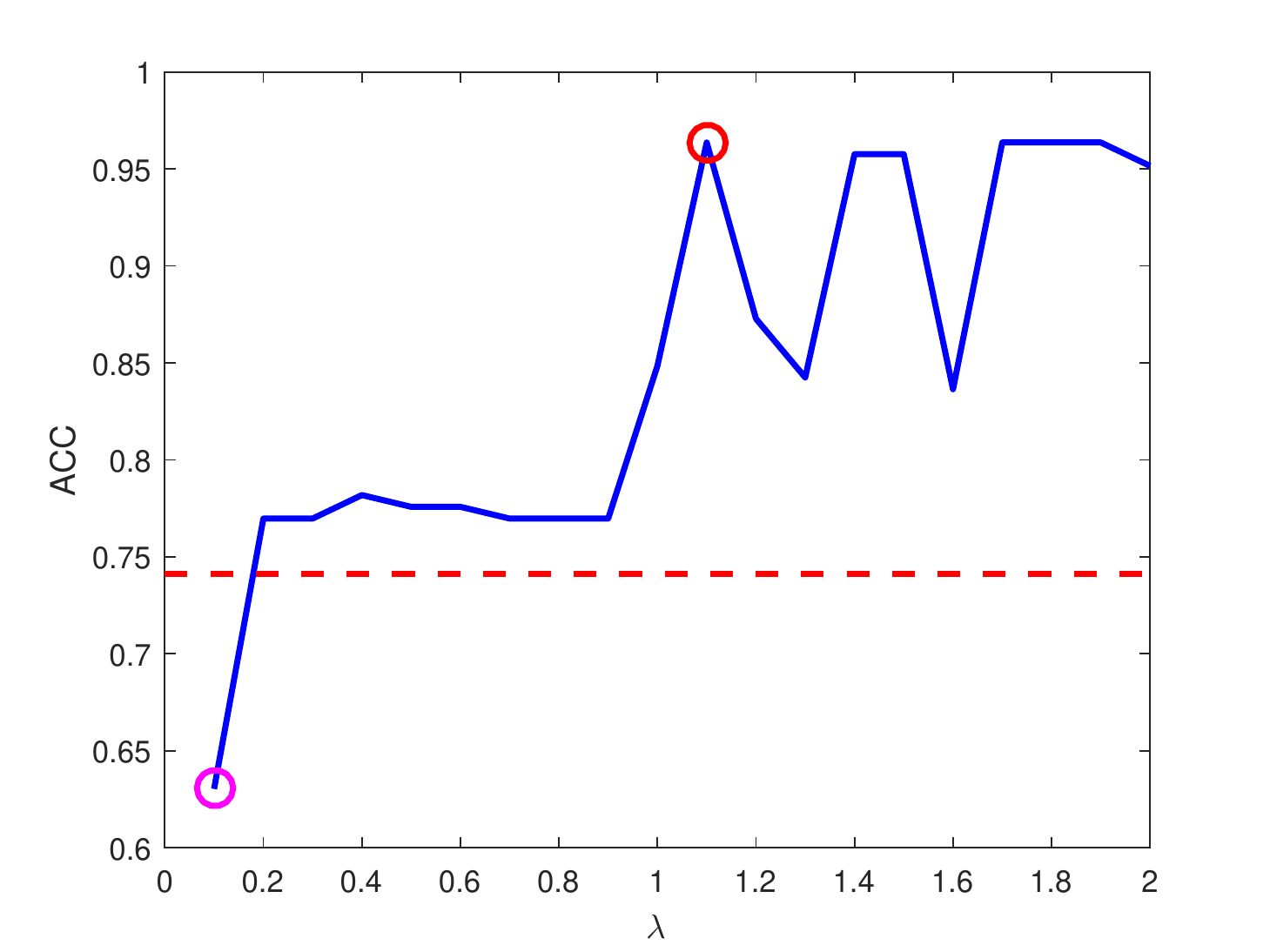}}
  \hspace{-0.2in}
  \subfloat[NMI on Yale]{
    \includegraphics[width=0.35\textwidth]{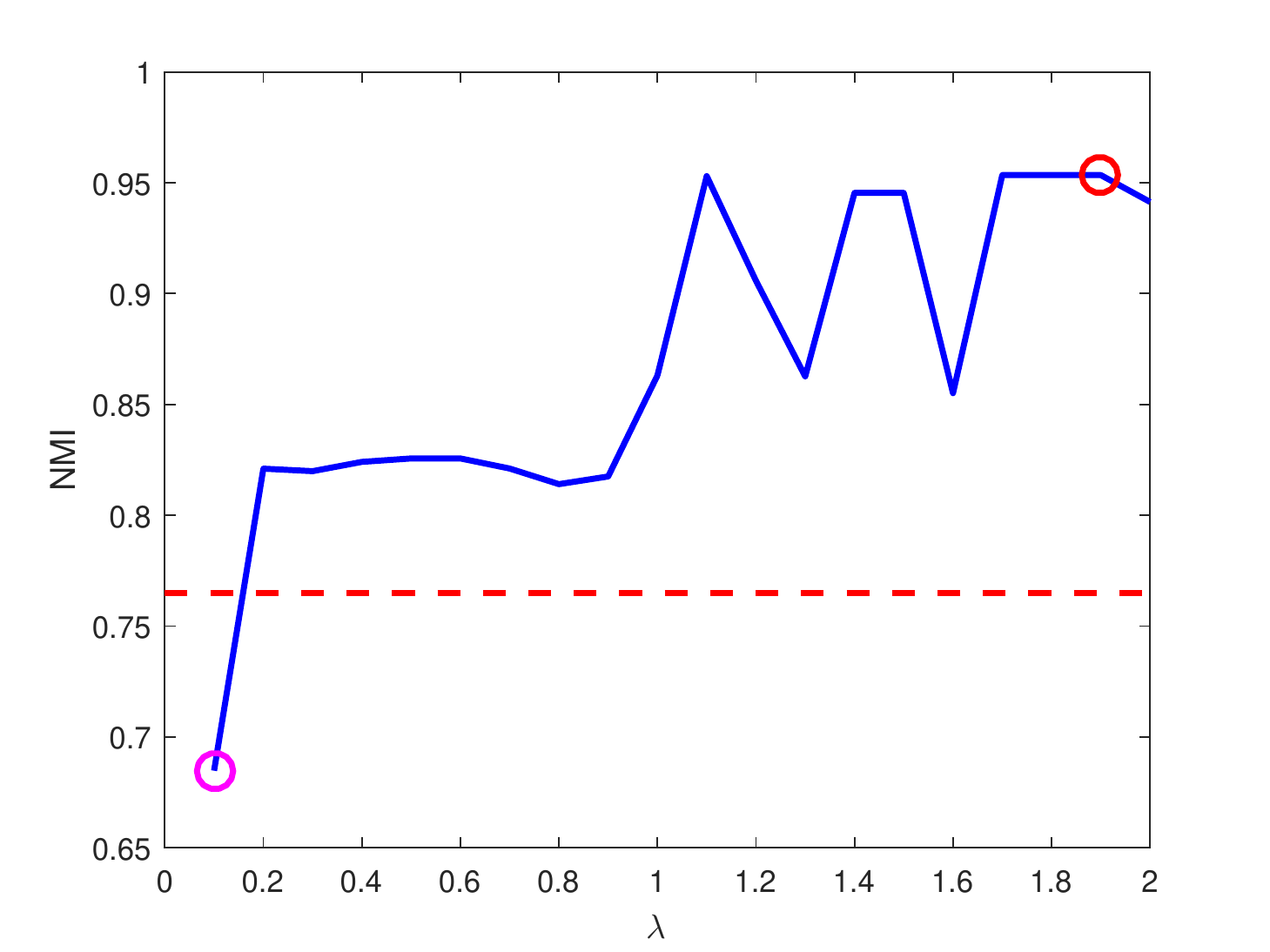}}
  \vspace{-0in}
  \subfloat[ACC on Scene-15]{
    \includegraphics[width=0.35\textwidth]{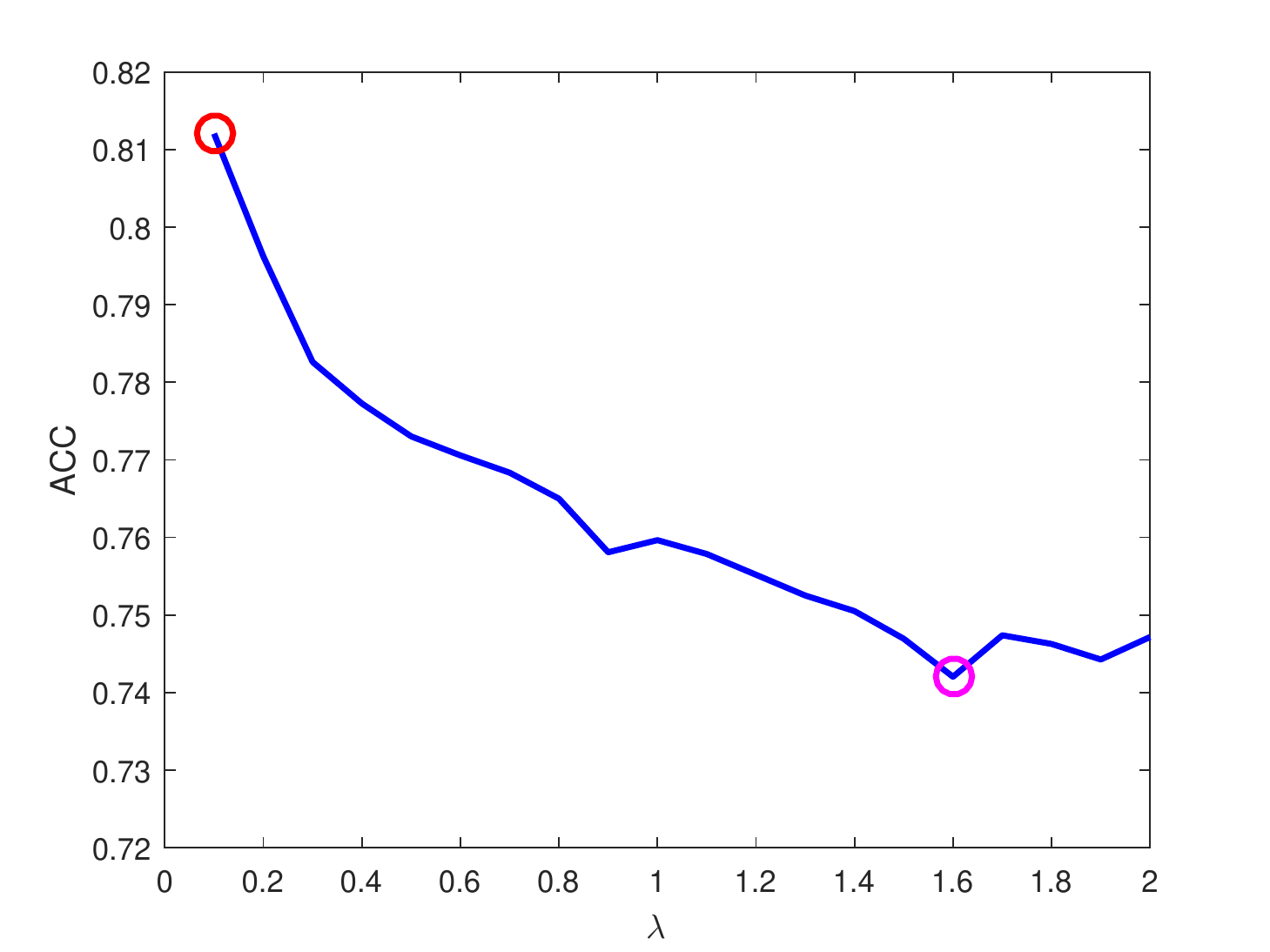}}
  \hspace{-0.2in}
  \subfloat[NMI on Scene-15]{
    \includegraphics[width=0.35\textwidth]{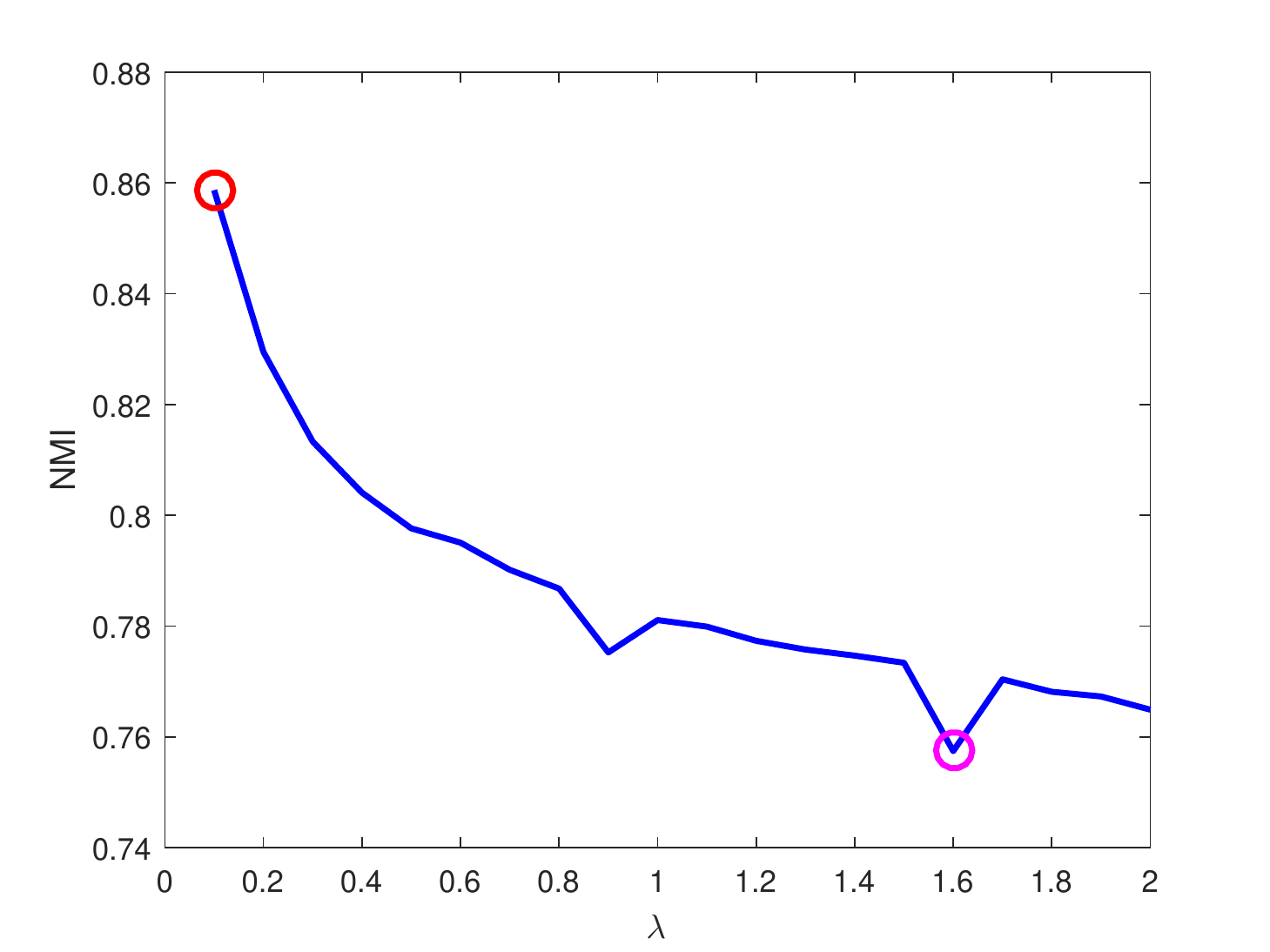}}
  \caption{Parameter ($\lambda$) tuning in terms of ACC and NMI on Yale and Scene-15 datasets.}
  \label{fig:plot-lambda} 
\end{figure}

We will discuss the parameter tuning in the proposed multi-view clustering model. Fortunately, the proposed model contains only one parameter $\lambda$ needed to be chosen. The parameter $\lambda > 0$ is used to balance the effects of the two parts in (\ref{original-problem}). Commonly, the choice of $\lambda$ depends on the prior knowledge of the error level of the data. Fig. \ref{fig:plot-lambda} shows the evaluation results on Yale and Scene-15 datasets by using different values of $\lambda$. Although the parameter $\lambda$ plays an important role on performance, most results are still better than other competitors, as can be seen from the red horizontal lines in Fig. \ref{fig:plot-lambda} (a) and (b), which denote the second best indexes. The same indexes are not presented in Fig. \ref{fig:plot-lambda} (c) and (d), as they are far below the minimal values of vertical ordinate. This implies the partial stability of the proposed model while $\lambda$ is varying.

{\color{red}
\subsubsection{Stability}
Overall, compared with all those competitors, the proposed method keeps relatively low standard deviation. Actually, the variance is mainly caused by the numerical calculation error of matrix inverse and matrix SVD, as well as the k-means algorithm in the final spectral clustering step. The matrix inverse operation is involved in optimization of $\mathbf{Z}^{(v)}$ in the step ($4$) in Algorithm \ref{proposed-MSC}, the SVD of complex matrix operation arises in updating $\boldsymbol{\mathcal{G}}$ in the step ($11$) (details in Algorithm \ref{TNN}). The final step ($18$), {\it i.e.,} the spectral clustering, also can incur the variance, since it contains the real matrix SVD and k-means algorithm. As we know, k-means algorithm is sensitive to initialization. However, relatively good affinity matrix provided by the proposed t-SVD-MSC can reduce the variance produced by k-means to some extent. This can be evidenced by observing the standard deviation of the proposed method in all the result tables (Table \ref{result-yale} $\sim$ Table \ref{result-Caltech101}), which indicates that the proposed t-SVD-MSC is a stable multi-view subspace clustering method.}

\subsubsection{Convergence and Computational Complexity}\label{convergence-and-complexity}

\begin{figure}[htb]
\setlength{\abovecaptionskip}{3pt}
\setlength{\belowcaptionskip}{0pt}
\renewcommand{\figurename}{Figure}
\centering
\includegraphics[width=0.4\textwidth]{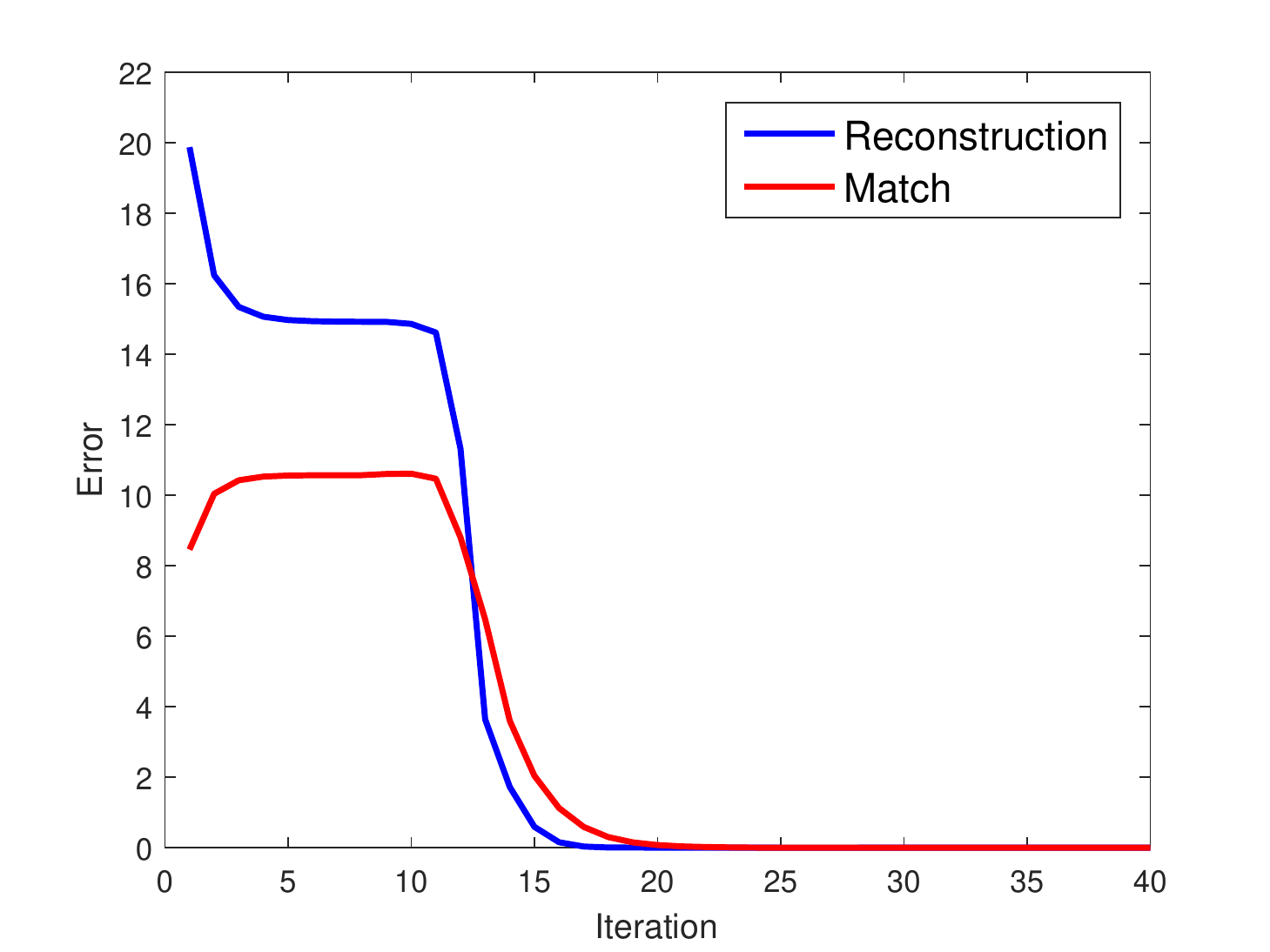}
\caption{Convergence curves on Scene-15 dataset.}
\label{fig:plot_convergence_Scene15}
\end{figure}

Thanks to the rotation of the coefficient tensor (see Fig. \ref{fig:shift-tSVD}), the computational complexity for SVD is reduced to $\mathcal{O}(N^{2}V^{2})$, compared with $\mathcal{O}(N^{3}V)$ for the unrotated tensor. In practice, the proposed optimization method for t-SVD-MSC converges fast, which is illustrated in Fig. \ref{fig:plot_convergence_Scene15}. The two curves record the reconstruction error (defined in Eq. (\ref{stop-criterion1})) and match error (Eq. (\ref{stop-criterion2})) in each iteration step. Additionally, the CPU times needed by the proposed method and its competitors are illustrated in Table \ref{cpu-compare}. Compared with other state-of-the-art algorithms, the proposed model is advanced not only in clustering performance, but also in saving time, especially when handling large-scale dataset (see the third row in Table \ref{cpu-compare}).
\begin{equation}\label{stop-criterion1}
    \text{Reconstruction Error} \doteq \frac{1}{V}\sum_{v=1}^{V}||\mathbf{X}^{(v)} - \mathbf{X}^{(v)}\mathbf{Z}^{(v)} - \mathbf{E}^{(v)}||_{\infty}
\end{equation}
\begin{equation}\label{stop-criterion2}
    \text{Match Error} \doteq \frac{1}{V}\sum_{v=1}^{V}||\mathbf{Z}^{(v)} - \mathbf{G}^{(v)}||_{\infty}
\end{equation}

\begin{table}[!htbp]
\caption{Comparison of CPU time of different methods, $s$, $m$ and $h$ denote second, minute and hour, respectively.}
\centering
\label{cpu-compare}
{
\begin{tabular}{|l||r|r|r|r|r|}
\hline
 & RMSC & DiMSC & LTMSC & Our \\
\hline\hline

\multirow{1}*{ORL}
      & 21.07s & 21.02s & 42.97s & $37.61s$\\\hline
\multirow{1}*{Scene-15}
      & $187.01m$ & $221.46m$ & $135.68m$ & $27.69m$ \\\hline
\multirow{1}*{Caltech-101}
      & $\sim 24 h$ & $> 27 h$ & $\sim 5h$ & $\sim 2h$ \\\hline
\end{tabular}
}%
\end{table}

\section{Conclusions}\label{sec:conclusion}
In this paper, a t-SVD based tensor low-rank subspace model is proposed to perform data clustering from multi-view features. To capture the complementary information from different views, the proposed method constrains the rotated subspace coefficient tensor through tensor multi-rank to explore the high order correlations. Then, the multi-view clustering problem have been formulated in a unified optimization framework, and an efficient algorithm is proposed to find the optimal solution. The proposed t-SVD-MSC is then applied to three kinds of image clustering datasets: face clustering, scene clustering, and generic object clustering. Extensive evaluation of our method is conducted on several challenge datasets, where a clear advance over contemporary MSC approaches is achieved. Meanwhile, the proposed model presents strongly robust to degenerate views. By utilizing CNN feature as a new view, the results show that t-SVD-MSC is very competitive with the recent proposed CNN based clustering approach on challenge datasets.



\section{Appendix}
Proof of the Theorem \ref{theory:tsvc}:
\begin{proof}
In Fourier domain, the optimization problem of Eq. (\ref{fml:ntsvd}) can be reformulated as
\begin{align}
\label{fml:obj_ntsvd_tc_y_2}
\boldsymbol{\mathcal{G}}_{f} =  &~\argmin_{\boldsymbol{\mathcal{G}}_{f}} ~ \tau||\mathrm{bdiag}(\boldsymbol{\mathcal{G}}_{f})||_{*} + \frac{1}{2n_{3}}||\boldsymbol{\mathcal{G}}_{f} - \boldsymbol{\mathcal{F}}_{f}||_{F}^{2}\\
 =  &~\argmin_{\boldsymbol{\mathcal{G}}_{f}} ~ \sum_{j=1}^{n_{3}}\tau'||\boldsymbol{\mathcal{G}}_{f}^{(j)}||_{*} + \frac{1}{2}||\boldsymbol{\mathcal{G}}_{f}^{(j)} - \boldsymbol{\mathcal{F}}_{f}^{(j)}||_{F}^{2}\label{fml:obj_ntsvd_tc_y_3},
\end{align}
where $\tau'=n_{3}\tau$. Then Eq. (\ref{fml:obj_ntsvd_tc_y_3}) can be separated into $n_{3}$ independent subproblems,
\begin{equation}
\label{fml:obj_ntsvd_tc_y_subproblem}
\boldsymbol{\mathcal{G}}_{f}^{(j)} = \argmin_{\boldsymbol{\mathcal{G}}_{f}^{(j)}} ~ \tau'||\boldsymbol{\mathcal{G}}_{f}^{(j)}||_{*} + \frac{1}{2}||\boldsymbol{\mathcal{G}}_{f}^{(j)} - \boldsymbol{\mathcal{F}}_{f}^{(j)}||_{F}^{2},
\end{equation}
where $j = 1,2,\ldots,n_{3}$. Note that Eq. (\ref{fml:obj_ntsvd_tc_y_subproblem}) is the \emph{F}-norm based nuclear norm low rank matrix approximation problem represented in Fourier domain. According to the result on gradients of unitarily invariant norms, Eq. (\ref{fml:obj_ntsvd_tc_y_subproblem}) can also be solved by a soft-thresholding operation \cite{svt},
\begin{equation}\label{fml:solution_yfj}
\boldsymbol{\mathcal{G}}_{f}^{(j)} = D_{\tau'}(\boldsymbol{\mathcal{F}}_{f}^{(j)})= \boldsymbol{\mathcal{U}}_{f}^{(j)}\boldsymbol{\mathcal{S}}_{f,\tau'}^{(j)}\boldsymbol{\mathcal{V}}_{f}^{(j)^{\mathrm{T}}},
\end{equation}
here, $\boldsymbol{\mathcal{G}}_{f}^{(j)} = \boldsymbol{\mathcal{U}}_{f}^{(j)}\boldsymbol{\mathcal{S}}_{f}^{(j)}\boldsymbol{\mathcal{V}}_{f}^{(j)^{\mathrm{T}}}$, $\mathcal{D}_{\tau'}(\cdot)$ is the SVT operation with with threshold $\tau'$ (see Section \ref{sec:notations}), and $\mathcal{S}_{f,\tau'}^{(j)}= \mathrm{diag}\{({\cal \mathcal{S}}_{f}^{(j)}(i,i)-\tau' )_{+}\}$. Then, we can get
\begin{equation}
\label{fml:solution_yf}
\boldsymbol{\mathcal{G}}_{f} = \mathrm{bdfold}\left\{\mathrm{bdiag}(\boldsymbol{{\cal U}}_{f})\mathrm{bdiag}(\boldsymbol{{\cal S}}_{f,\tau'}) \mathrm{bdiag}(\boldsymbol{{\cal V}}_{f})^{\mathrm{T}}\right\}
\end{equation}
and
\begin{equation}
\label{fml:solution_y}
\boldsymbol{\mathcal{G}} = \boldsymbol{\mathcal{U}}*\boldsymbol{\mathcal{\tilde{S}}}*\boldsymbol{\mathcal{V}^{\mathrm{T}}}
\end{equation}
where $\boldsymbol{\mathcal{\tilde{S}}} = \mathrm{ifft}(\boldsymbol{{\cal S}}_{f,\tau'},[~], 3)$. Suppose that $\boldsymbol{{\cal J}}$ is an $n_{1} \times n_{2} \times n_{3}$ f-diagonal tensor whose diagonal element in the Fourier domain is $\boldsymbol{{\cal J}}_{f}(i,i,j) = (1 - \frac{\tau'}{\boldsymbol{{\cal S}}_{f}^{(j)}(i,i)})_{+}$, then we have that $\boldsymbol{{\cal S}}_{f,\tau'}(i,i,:) = \boldsymbol{{\cal S}}_{f}(i,i,:)\boldsymbol{{\cal J}}_{f}(i,i,:)$ in the Fourier domain, as well as $\boldsymbol{{\cal \tilde{S}}}(i,i,:) = \boldsymbol{{\cal S}}(i,i,:) \circ \boldsymbol{{\cal J}}(i,i,:)$ in the original domain. Because both $\boldsymbol{{\cal S}}$ and $\boldsymbol{{\cal J}}$ are f-diagonal, $\boldsymbol{{\cal \tilde{S}}}$ can be formulated as $\boldsymbol{{\cal \tilde{S}}} = \boldsymbol{{\cal S}} * \boldsymbol{{\cal J}}$. Therefore, a convolution based tubal-shrinkage operator in the original domain is equivalent to the tensor SVT in the Fourier domain.
\end{proof}

\end{document}